\documentclass[11pt]{article}

\usepackage[final]{acl}

\usepackage{times}
\usepackage{latexsym}

\usepackage[T1]{fontenc}

\usepackage[utf8]{inputenc}

\usepackage{microtype}

\usepackage{inconsolata}

\usepackage{graphicx}
\usepackage{kotex}
\usepackage{amsmath,amssymb}
\usepackage{mathtools} 
\usepackage{wrapfig}        
\usepackage{booktabs}       
\usepackage{multirow}       
\usepackage{array}          
\usepackage{threeparttable} 
\usepackage{graphicx}       
\usepackage{caption}        
\usepackage{makecell}
\usepackage{subcaption}
\usepackage[most]{tcolorbox}
\usepackage{xcolor}
\usepackage{listings}
\usepackage{inconsolata} 
\usepackage{caption}
\usepackage{cleveref}
\usepackage{algorithm}
\usepackage{algpseudocode}
\usepackage{etoc}
\etocdepthtag.toc{mtchapter}
\etocsettagdepth{mtchapter}{subsubsection}
\etocsettagdepth{mtappendix}{none}

\crefname{figure}{Fig.}{Figs.}
\Crefname{figure}{Fig.}{Figs.}

\title{LIBERO-Para: A Diagnostic Benchmark and Metrics \\ for Paraphrase Robustness in VLA Models}

\author{
  Chanyoung Kim$^{1*}$, Minwoo Kim$^{2*}$, Minseok Kang$^{2}$, Hyunwoo Kim$^{2}$, Dahuin Jung$^{2\dagger}$ \\
  \\
  $^{1}$School of Computer Science and Engineering, Soongsil University, Seoul, Korea \\
  $^{2}$Department of Artificial Intelligence, Chung-Ang University, Seoul, Korea \\
  \texttt{dahuinjung@cau.ac.kr} \\
}
  \usepackage{fancyhdr}

\usepackage{xcolor}

\newif\ifrevision
\revisionfalse  

\ifrevision
  \newcommand{\rev}[1]{\textcolor{blue}{#1}}
\else
  \newcommand{\rev}[1]{#1}
\fi

\usepackage{eso-pic}
\begin{document}
\maketitle
\AddToShipoutPictureBG*{%
  \AtPageLowerLeft{%
    \raisebox{2.2\baselineskip}{%
      \makebox[\paperwidth][c]{%
        \footnotesize Accepted to EMNLP 2026 (Main Conference)}}}}
\let\thefootnote\relax
\footnotetext{$^{*}$Equal contribution. $^{\dagger}$Corresponding author.}
\addtocounter{footnote}{0}
\begin{abstract}
   Vision–Language–Action (VLA) models achieve strong performance in robotic manipulation by leveraging pre-trained vision–language backbones. However, in downstream robotic settings, they are typically fine-tuned with limited data, leading to overfitting to specific instruction formulations and leaving robustness to paraphrased instructions underexplored. To study this gap, we introduce LIBERO-Para, a controlled benchmark that independently varies action expressions and object references for fine-grained analysis of linguistic generalization. Across seven VLA configurations (0.6B–7.5B), we observe consistent performance degradation of 22–52\,pp under paraphrasing. This degradation is primarily driven by object-level lexical variation: even simple synonym substitutions cause large drops, indicating reliance on surface-level matching rather than semantic grounding. Moreover, 80–96\% of failures arise from planning-level trajectory divergence rather than execution errors, showing that paraphrasing disrupts task identification. Binary success rate treats all paraphrases equally, obscuring whether models perform consistently across difficulty levels or rely on easier cases. To address this, we propose PRIDE, a metric that quantifies paraphrase difficulty using semantic and syntactic factors. Our benchmark and code are available at \url{https://github.com/cau-hai-lab/LIBERO-Para}.
   %
\end{abstract}
\section{Introduction}

\begin{figure}[t]
  \centering
  \includegraphics[width=.9\linewidth]{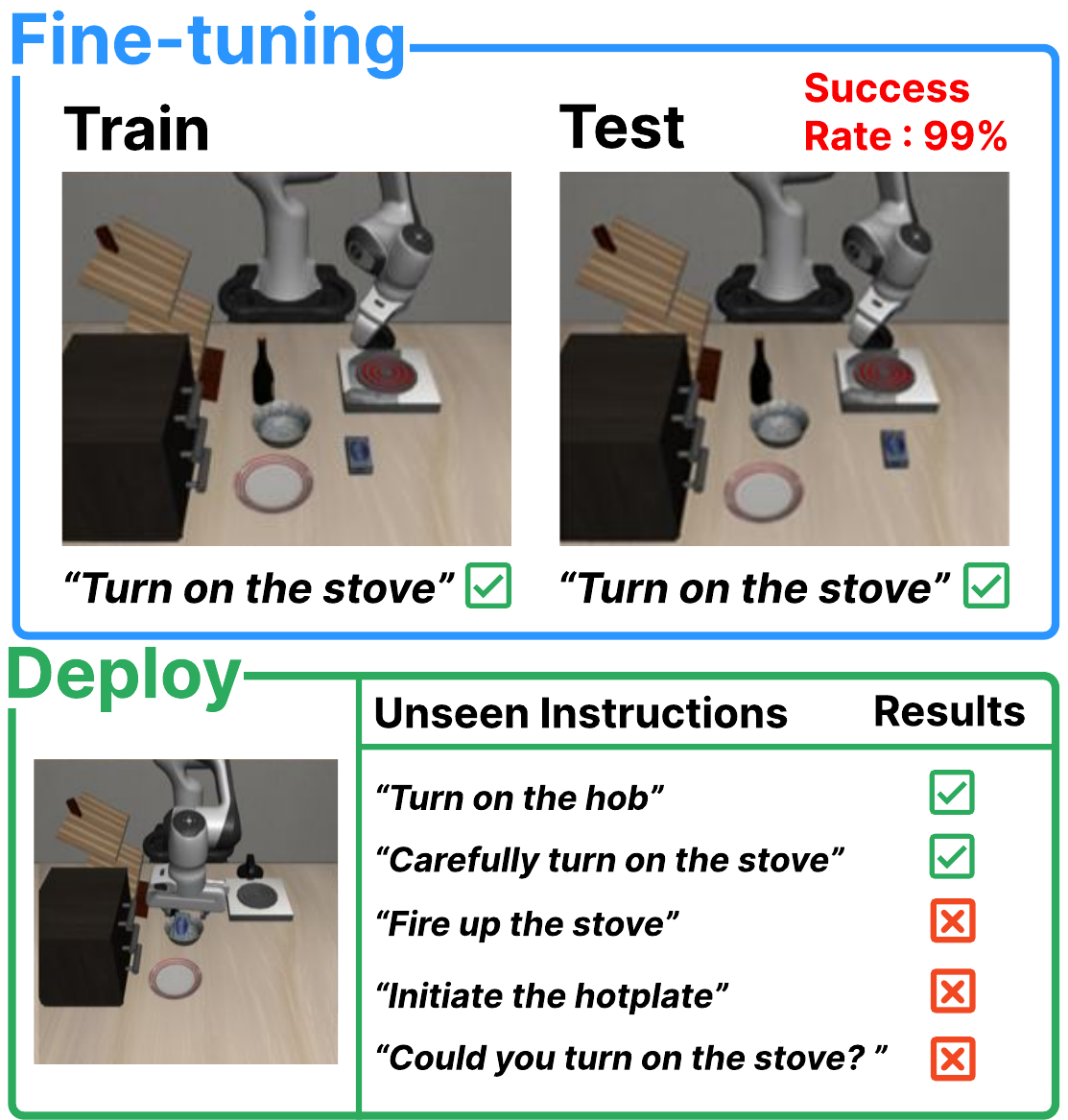}
  \caption{Illustration of paraphrase robustness gap under data-scarce fine-tuning: VLA models can overfit to seen instruction phrasings during fine-tuning and fail to generalize to paraphrased variants at deployment.}
  \label{fig:paraphrase-gap}
\end{figure}

Vision-Language-Action (VLA) models have emerged as a promising paradigm for robotic manipulation~\citep{rt2,pi0}. By leveraging large pre-trained vision-language models (VLMs) as backbones, VLA models acquire instruction-following capabilities through large-scale multimodal perception~\citep{rt2,pi0}. To deploy such models in specific environments (e.g., kitchens, homes, offices, or laundry rooms)~\citep{pi0, pi05, rt2, mobilealoha, tidybot}, existing approaches typically perform fine-tuning using environment-specific demonstration data~\citep{mobilealoha}. However, acquiring such data entails considerable cost and labor overhead. Consequently, real-world deployment often necessitates data-scarce fine-tuning, which can induce overfitting and degrade the general knowledge embedded in pre-trained VLA models~\citep{retain,libero_pro}. Such overfitting raises a practical concern that models may perform well on seen instruction phrasings yet fail to generalize to unseen paraphrased instructions at deployment. Under these circumstances, as shown in Fig.~\ref{fig:paraphrase-gap}, benchmarks that rigorously assess robustness after fine-tuning become important~\citep{libero_pro}.

However, the LIBERO benchmark~\citep{libero}, which has become widely adopted in current VLA research, evaluates models under identical instructions during both training and evaluation. It primarily measures visual generalization to novel object configurations or scene layouts, while leaving robustness to linguistic variation largely unexamined. Consequently, the linguistic robustness of VLA models remains insufficiently validated~\citep{libero_plus,ladev,libero_pro}.

Several benchmarks have examined linguistic variation in VLA evaluation~\citep{calvin, ladev}. However, as summarized in Tab.~\ref{tab:benchmark}, these approaches exhibit key limitations for assessing paraphrase robustness. Paraphrasing is often treated as one axis among broader multimodal perturbations~\citep{libero_pro, libero_plus, wang2026libero-x}, or conflated with task-level semantic changes that alter the intended behavior~\citep{hou2026langgap}, rather than isolating meaning-preserving variation. Furthermore, linguistic properties specific to robotic manipulation instructions are not explicitly modeled, and the distance between paraphrases is not formally quantified, limiting the ability to analyze which types of variation most severely impact performance.

To address these limitations, we introduce LIBERO-Para, a controlled benchmark for evaluating paraphrase robustness in VLA models, along with PRIDE (Paraphrase Robustness Index in Robotic Instructional DEviation), a metric that combines keyword similarity (lexical shift) and structural similarity (syntactic variation) with task success for fine-grained robustness analysis. LIBERO-Para is grounded in the linguistic structure of robotic manipulation instructions—where actions and objects serve as core semantic elements—and adopts a two-axis design that independently varies action expressions and object references. Our analysis reveals three key findings:
\begin{itemize}
\setlength{\itemsep}{2pt}
\setlength{\parskip}{2pt}
\setlength{\parsep}{2pt}
    \item \textbf{Paraphrase Fragility Persists:} Performance consistently degrades under paraphrased instructions across architectures, scales, and fine-tuning strategies.
    \item \textbf{Object-Level Bottleneck:} Object-level lexical variation is the dominant source of degradation, indicating reliance on surface-level matching rather than semantic grounding.
    \item \textbf{Planning-Level Failures:} 80–96\% of failures arise from trajectory divergence, suggesting errors in task identification rather than action execution.
\end{itemize}
This work contributes to advancing VLA systems beyond high performance toward robustness to linguistic variation and reliable task interpretation.
\section{Related Work}

\begin{table}[t]
\centering
\resizebox{\columnwidth}{!}{%
\begin{tabular}{lcccc}
\hline
\textbf{Benchmark}
& \textbf{Scope}
& \makecell{\textbf{Paraphrase} \\ \textbf{Control}}
& \makecell{\textbf{Variation} \\ \textbf{Axis}}
& \makecell{\textbf{Para.} \\ \textbf{Types}} \\
\hline
CALVIN      & Instruction & $\times$     & Sentence               & 1 \\
LADEV       & Paraphrase  & $\times$     & Sentence               & 1 \\
LIBERO-PRO  & Multimodal  & $\Delta$     & Sentence               & 2 \\
LIBERO-Plus & Multimodal  & $\Delta$     & Sentence               & 5 \\
LIBERO-X    & Multimodal  & $\Delta$     & Sentence               & 5 \\
LangGap     & Task Semantics    & $\times$     & 4 Semantic dims         & 4 \\
\hline
LIBERO-Para (Ours) & Paraphrase  & $\checkmark$ & Action $\times$ Object & 43 \\
\hline
\end{tabular}%
}
\caption{Comparison with existing benchmarks for linguistic robustness. LIBERO-Para provides full paraphrase control with fine-grained action $\times$ object variation axes and 43 linguistically grounded types. $\times$: not supported, $\Delta$: partially supported, $\checkmark$: fully supported.}
\label{tab:benchmark}
\end{table}
\subsection{Vision-Language-Action Models}

Vision-Language-Action (VLA) models map visual and linguistic input to robot actions. Early approaches extend LLM backbones to autoregressively decode discrete action tokens~\citep{rt2, openvla}. Recent work has diversified along several architectural axes: parallel decoding with action chunking, which predicts all actions in a single forward pass~\citep{openvla-oft}; VLMs with a flow-matching action expert, a separate decoder paired with a billion-scale VLM~\citep{pi0, pi05, cai2026xiaomirobotics}; lightweight bridge-based adaptation, which routes VLM representations to a compact policy head via cross-attention~\citep{wang2025vlaadapter}; and soft-prompted cross-embodiment designs, which encode embodiment-specific knowledge through learnable tokens~\citep{zheng2025xvla}. The latter two operate at 0.6--0.9B scale, unlike earlier multi-billion-parameter designs. Despite this diversity, all models require environment-specific fine-tuning with limited demonstration data. We evaluate representatives from each family to assess whether paraphrase robustness is architecture-specific or a shared vulnerability.
\begin{figure*}[t]
    \centering
    \includegraphics[width=\textwidth]{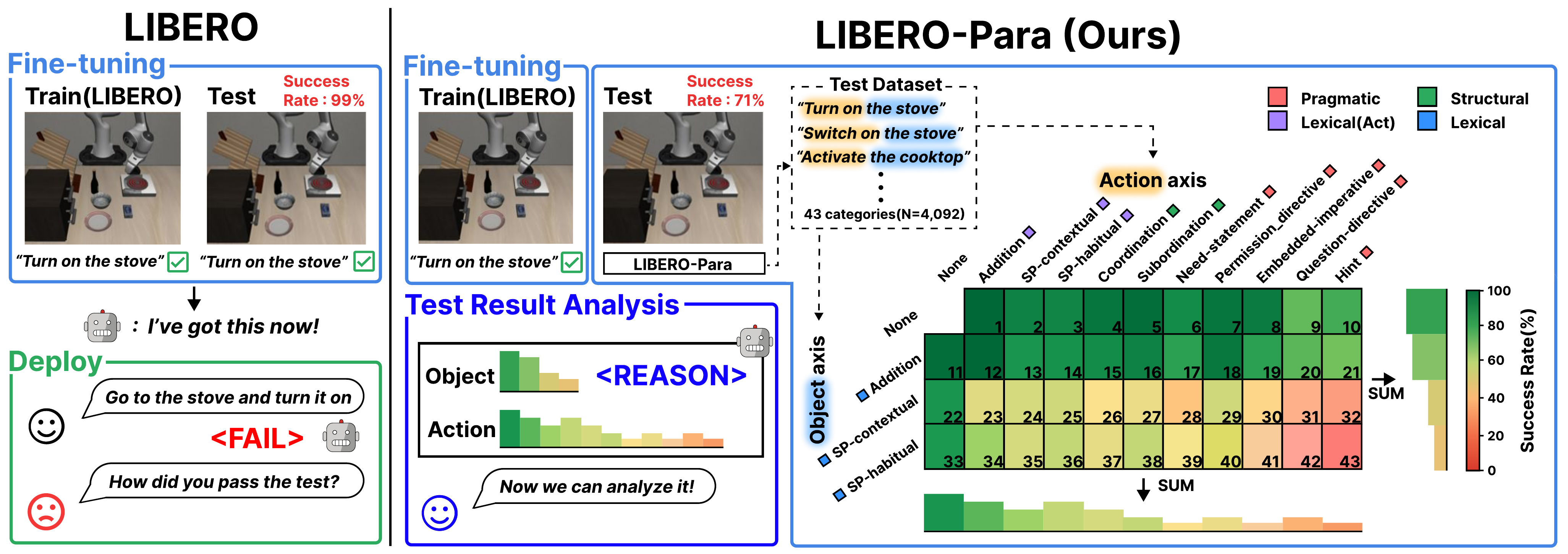}
    \caption{Overview of LIBERO-Para. Compared to LIBERO, LIBERO-Para evaluates paraphrase robustness under data-scarce fine-tuning via a controlled two-axis design (action vs. object), enabling interpretable analysis.}
    \label{fig:libero-para-overview}
    \vspace{-0.05in}
\end{figure*}
\subsection{Benchmarks for VLA models}
\label{sec:related_vla_benchmarks}
A range of benchmarks have been proposed to evaluate linguistic conditioning in VLA models; Tab.~\ref{tab:benchmark} compares their design choices. CALVIN~\citep{calvin} and LADEV~\citep{ladev} assess generalization to rephrased instructions, but treat paraphrasing as unstructured sentence-level variation without linguistic categorization. LIBERO-PRO, LIBERO-Plus, and LIBERO-X~\citep{libero_pro, libero_plus, wang2026libero-x} introduce multimodal perturbation-based evaluations that include linguistic variation as one of several axes, revealing limited dependence on genuine linguistic understanding; however, paraphrasing remains a secondary concern within their broader evaluation scope. LangGap~\citep{hou2026langgap} targets language conditioning more directly, but its perturbations alter the intended behavior (e.g., changing which object to grasp), conflating task-level semantic changes with linguistic variation. \rev{In contrast, our LIBERO-Para differs in two key aspects. First, it isolates
meaning-preserving linguistic variation from task-level semantic changes. Second, instead of sentence-level perturbations with ad-hoc categories, it identifies the essential components of manipulation instructions, action verbs and object references, and decomposes paraphrases along these axes following established linguistic taxonomies~\citep{kovatchev-etal-2018-etpc, ervin-tripp1976}. This factorization yields 43 fine-grained variation types (Tab.~\ref{tab:benchmark}) and exposes a super-additive interaction that single-axis or sentence-level perturbation cannot capture.}

\section{LIBERO-Para: A Controlled VLA Benchmark for Paraphrase Robustness}
\label{sec:liberopara}

We introduce LIBERO-Para, a controlled benchmark for evaluating the paraphrase robustness of VLA models. As shown in Tab.~\ref{tab:benchmark}, existing benchmarks offer limited control over paraphrase variation; our design addresses this through a two-axis scheme that independently varies action expressions and object references—the two core linguistic components of robotic manipulation instructions. This separation enables controlled analysis of how different linguistic factors affect VLA performance.
LIBERO-Para is constructed on top of LIBERO-Goal, a setting in which linguistic understanding is essential: all tasks start from an identical initial state, making the instruction the sole cue for task identification. We generate the benchmark by paraphrasing only the instructions while keeping all other factors fixed. All paraphrases are held out for evaluation, allowing assessment of generalization to unseen linguistic variations under data-scarce fine-tuning scenarios, as illustrated in Fig.~\ref{fig:libero-para-overview}.


\subsection{Action Variation}
The action axis captures variation in how actions are linguistically expressed. We define three types of action variation grounded in established paraphrase taxonomies. (1) Lexical variation modifies the action verb at the word level, including synonym substitution and adverb insertion. (2) Structural variation alters the sentence-level realization of the action expression, such as coordination and subordination. (3) Pragmatic variation expresses actions indirectly, covering indirect speech acts. Lexical and structural variations are instantiated based on the Extended Paraphrase Typology \citep{kovatchev-etal-2018-etpc}. Pragmatic variations are defined in accordance with \citet{ervin-tripp1976}. Fig.~\ref{fig:taxonomy-examples} (bottom) presents representative examples for each type.
\subsection{Object Variation} 
\begin{figure}[t]
    \centering
    \includegraphics[width=\linewidth]{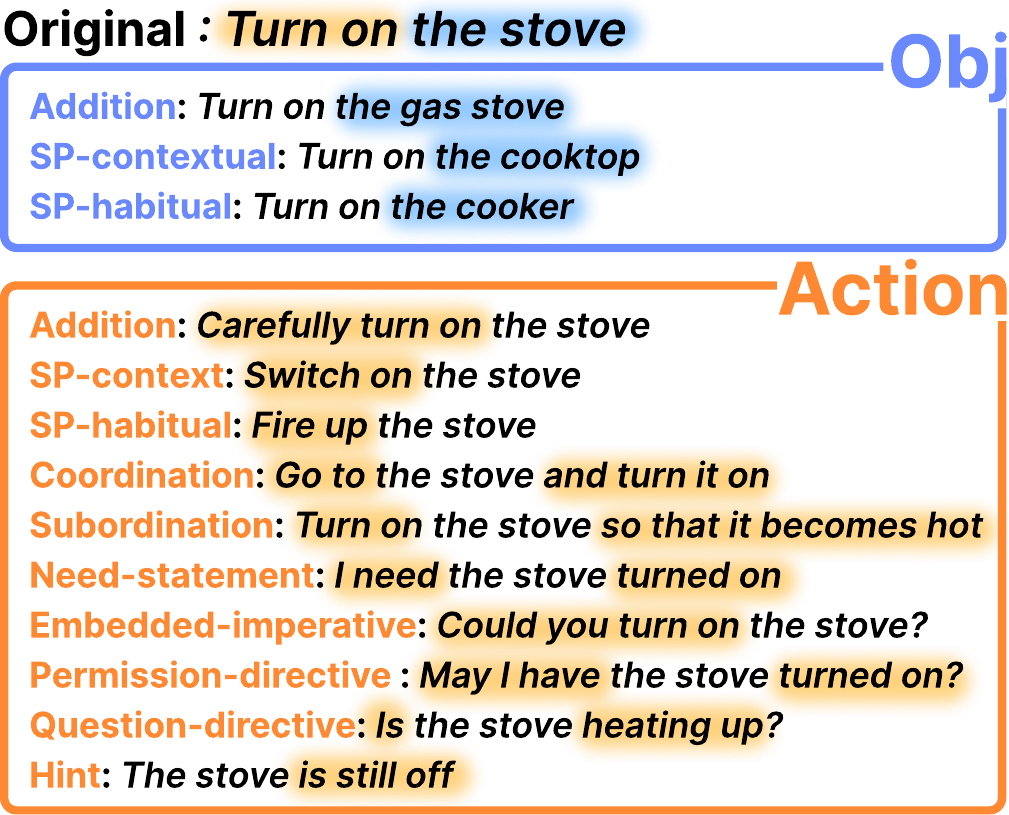}
    \caption{Examples of axis-specific paraphrases. Object variations modify target object references (e.g., same-polarity substitution, addition), while action variations cover lexical, structural, and pragmatic realizations grounded in established taxonomies.}
    \label{fig:taxonomy-examples}
\end{figure}

The object axis captures variation in how objects are referenced in instructions. In robotic manipulation, object references are typically realized as noun phrases with limited complexity (e.g., ``\textit{the stove}'' $\rightarrow$ ``\textit{the cooktop}''). We focus on lexical-level variation. Following the Extended Paraphrase Typology ~\citep{kovatchev-etal-2018-etpc}, we define three subtypes: addition and same-polarity substitution (contextual and habitual variants). Fig.~\ref{fig:taxonomy-examples} (top) illustrates representative examples.

\subsection{Compositional Variation}
Beyond individual axes, we evaluate compositional paraphrases varying both action and object expressions. This setting analyzes whether the two
axes have independent or interaction effects on VLA performance, \rev{which cannot be determined from single-axis perturbations by construction.} Fig.~\ref{fig:libero-para-overview} presents an example with a success-rate grid over variation combinations and per-axis success rates.

To ensure balanced evaluation, the benchmark includes approximately 100 samples per variation type, resulting in a total of 4,092 paraphrased instructions. Additional details on the taxonomy, paraphrase generation process, and excluded variation types—including justifications for why certain types are inapplicable to robotic manipulation instructions—are provided in Appendix~\ref{app:taxonomy}.

\section{PRIDE: Paraphrase Robustness Index in Robotic Instructional DEviation}
\label{sec:metric}

Existing VLA benchmarks rely on a binary success metric, assigning $1$ if the robot completes the instructed task and $0$ otherwise. However, this metric does not distinguish between easy and difficult paraphrases, obscuring whether performance reflects robust linguistic understanding or reliance on simpler instruction variants.

To enable interpretable robustness evaluation, we propose PRIDE, a metric that quantifies the linguistic deviation between an original instruction and its paraphrase. Unlike general-purpose metrics, PRIDE is tailored to robotic instructions and decomposes paraphrase variation along two axes: (1) keyword variation and (2) structural variation.

\subsection{Keyword Similarity $S_K$}
\label{sec:metric:semantic}

Keyword similarity measures how much core keywords expressing actions and target objects are preserved between an original instruction and its paraphrase. Robotic manipulation instructions are typically structured around explicit actions and their corresponding objects, often following a canonical form such as ``\texttt{[ACT] the [OBJ]}'' (e.g., \textit{"pick up the bowl"}). As a result, the intended behavior is often determined by a small set of task-critical tokens rather than by the sentence as a whole.

This property limits the usefulness of form-based NLP metrics for paraphrased robotic instructions. For example, n-gram metrics such as BLEU~\cite{bleu} emphasize lexical overlap and may fail to distinguish paraphrases that preserve actions and objects but differ in expression, through synonym substitution or word-order variation. Function words can also cause superficial grammatical changes to overly influence similarity scores relative to action- or object-level semantics.

In our setting, the two sentences are given as a paraphrase pair. Thus, rather than reassessing overall semantic equivalence, it is more useful to analyze how task-critical components change. Accordingly, we define a keyword-level similarity that focuses on content words expressing actions and objects excluding function words.

\begin{figure}[t]
    \centering
    \includegraphics[width=1\linewidth]{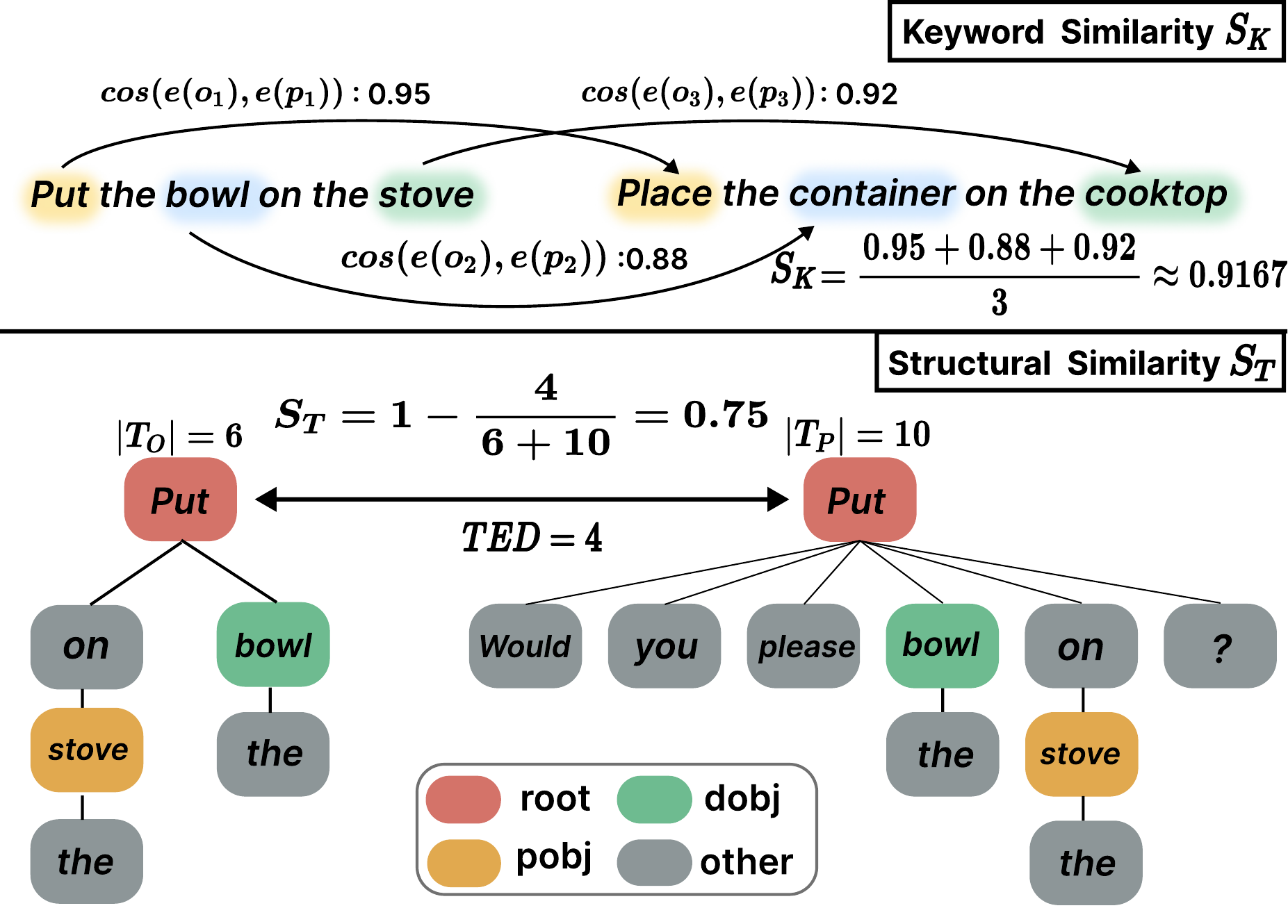}
    \caption{$S_K$ (top) and $S_T$ (bottom) computation. $S_K$ is based on semantic matching between task-critical content words, while $S_T$ uses dependency-tree edit distance. Node colors indicate dependency relations: root (sentence root), dobj (direct object), pobj (object of preposition), and others (remaining types, simplified for visualization; all included in computation).}
    \label{fig:pd-overview}
\end{figure}

Let $O=\{o_1,\dots,o_n\}$ and $P=\{p_1,\dots,p_m\}$ denote the sets of content words extracted from the original and paraphrased instructions, respectively. Each word is represented by an embedding $e(\cdot)$ obtained from Sentence-BERT~\cite{sentence}. The keyword similarity $S_K(O,P)$ is computed by matching each content word $o_i$ in the original instruction to the most similar word in the paraphrase, measured by cosine similarity, and averaging over all $o_i$:
\begin{equation}
S_K(O,P)
=\frac{1}{n}\sum_{i=1}^{n}
\max_{j\in\{1,\dots,m\}}
\cos\big(e(o_i), e(p_j)\big),
\label{eq:1}
\end{equation}
where $\cos(\cdot,\cdot)$ denotes cosine similarity. Fig.~\ref{fig:pd-overview} (top) illustrates the computation of $S_K$. A higher value of $S_K(O,P)$ indicates better preservation of the original instruction’s key content words in the paraphrase. \rev{We further discuss a dependency-head-aware extension of $S_K$ in Appendix~\ref{sec:appen_metric}.}

\subsection{Structural Similarity $S_T$}
\label{sec:metric:structural}

While keyword similarity $S_K$ captures the preservation of core lexical items, it does not account for syntactic changes. Transformations such as active-passive alternation or clause reordering can substantially alter the form of an instruction while preserving the same keywords. To capture such variation, we introduce structural similarity $S_T$.

We measure syntactic change using the tree edit distance (TED)~\citep{augstenboehlen2013similarityjoins} between the dependency trees of the original and paraphrased instructions, denoted by $T_O$ and $T_P$. TED is defined as the minimum number of edit operations—node and edge insertions, deletions, and substitutions—required to transform one tree into the other. To focus on structural differences, we compute TED on dependency trees whose node labels are part-of-speech (POS) tags and whose edge labels are dependency relations rather than surface words, reducing sensitivity to lexical substitutions.

To mitigate sentence-length effects, we normalize TED by the combined size of two trees and define structural similarity $S_T(T_O,T_P)$ as follows:
\begin{equation}
S_T(T_O,T_P)
=1-\frac{\mathrm{TED}(T_O,T_P)}{|T_O|+|T_P|}.
\label{eq:2}
\end{equation}
where $\lvert \cdot \rvert$ denotes the number of nodes in tree $T$. Fig.~\ref{fig:pd-overview} (bottom) illustrates the computation of $S_T$. Lower values of $S_T(T_O,T_P)$ indicate greater structural divergence, such as word-order changes or reorganization of modification relations.
\rev{Additional validation of dependency-parser stability and the choice of TED normalization is provided in Appendix~\ref{sec:appen_metric}.}

\subsection{PRIDE Score}
\label{sec:metric:pride}

Robustly assessing paraphrased robotic manipulation instructions requires considering both (i) whether task-critical action/object keywords are preserved ($S_K$) and (ii) how much the imperative structure is altered ($S_T$). \rev{We verify that the two components indeed measure distinct
dimensions, each correlating consistently with independent lexical and
syntactic measures of its own axis (Appendix~\ref{sec:appen_metric}).} Accordingly, we define Paraphrase Distance (PD) by combining $S_K$ and $S_T$ to quantify the overall deviation between an original robotic instruction and its paraphrase:
\begin{equation}
    \mathrm{PD}
    = 1 - \Big(
    \alpha S_K(O,P)
    + (1-\alpha) S_T(T_O,T_P)
    \Big),
\label{eq:3}
\end{equation}
\rev{where $\alpha \in [0,1]$ is a diagnostic weighting factor that controls the relative emphasis on keyword and structural similarity, rather than a hyperparameter to be tuned for performance. We use $\alpha=0.5$ as a neutral default that weights the two components equally; varying $\alpha$ exposes whether a model is more sensitive to lexical or structural variation. Higher PD indicates greater semantic and structural deviation. Combining PD with task success yields the PRIDE score:}
\begin{equation}
\mathrm{PRIDE}=
\begin{cases}
\mathrm{PD}, & \text{success,}\\
0, & \text{failure.}
\end{cases}
\label{eq:4}
\end{equation}
\rev{PRIDE is a success-conditioned robustness score. A failed sample receives zero robustness credit, while PD, defined for every instruction regardless of task success, continues to represent its linguistic difficulty. Assigning zero to failures therefore does not discard difficulty information, but reflects that the model failed to handle the corresponding paraphrase. PRIDE thus complements binary success metrics by distinguishing whether a VLA model can succeed under paraphrased instructions that exhibit larger semantic and structural deviations.}

\section{Experiment}
\begin{table}[t]
\centering
\small
\begin{tabular}{lcc}
\toprule
\multirow{2}{*}{Method} & LIBERO-Goal & LIBERO-Para \\
 & SR & SR (Drop) \\
\midrule
OpenVLA-OFT\textsubscript{goal}  & 97.9 & 64.7 (\textcolor{red}{-33.2}) \\
OpenVLA-OFT\textsubscript{mixed} & 96.1 & 63.7 (\textcolor{red}{-32.4}) \\
$\pi$\textsubscript{0.5}             & 97.6 & 71.4 (\textcolor{red}{-26.2}) \\
$\pi$\textsubscript{0.5} (expert-only) & 78.6 & 39.1 (\textcolor{red}{-39.5}) \\
X-VLA              & 97.8 & 62.1 (\textcolor{red}{-35.7}) \\
VLA-Adapter        & 98.2 & 46.3 (\textcolor{red}{-51.9}) \\
Xiaomi-Robotics-0    & 98.8 & 76.0 (\textcolor{red}{-22.8}) \\
\bottomrule
\end{tabular}
\vspace{-0.05in}
\caption{Success rate (SR) comparison between LIBERO-Goal and LIBERO-Para. Drop denotes the absolute decrease in success rate.}
\label{tab:sr_drop}
\end{table}
\subsection{Setup}
We evaluate seven model configurations (0.6B--7.5B) spanning four architecture families: parallel decoding with action chunking (OpenVLA-OFT), VLMs with a flow-matching action expert ($\pi$\textsubscript{0.5}, Xiaomi-Robotics-0), soft-prompted cross-embodiment (X-VLA), and bridge-based adaptation (VLA-Adapter). Within the same architecture, we include controlled comparisons on fine-tuning data scope (OFT\textsubscript{goal} vs.\ OFT\textsubscript{mixed}) and VLM training strategy ($\pi$\textsubscript{0.5} full vs.\ expert-only). Full specifications are in Appendix~\ref{appen:Experiment_Setup}.
\subsection{Results}


\paragraph{Success Rate Comparison.}
Tab.~\ref{tab:sr_drop} compares success rates between LIBERO-Goal and LIBERO-Para. All models exhibit consistent performance degradation ranging from 22.8\,pp to 51.9\,pp, indicating that the effect is not architecture-specific but pervasive across models. Even the top-performing models on LIBERO-Goal (Xiaomi-Robotics-0: 98.8\%, VLA-Adapter: 98.2\%) suffer drops of 22.8\,pp and 51.9\,pp under paraphrasing, respectively, with VLA-Adapter losing nearly half of its performance.


\paragraph{PRIDE Reveals Hidden Severity.}
Uniform SR treats all paraphrases equally, assigning the same reward to easy and difficult variations, and thus cannot distinguish success limited to easy paraphrases from success that generalizes to harder ones. PRIDE mitigates this limitation by weighting rewards by difficulty. Tab.~\ref{tab:pride_overestimation} re-evaluates the same results under PRIDE. VLA-Adapter (22.0\%) and $\pi$\textsubscript{0.5} expert-only (18.2\%) show large drops relative to SR, indicating success mainly on easy paraphrases and systematic failures on harder variations. In contrast, $\pi$\textsubscript{0.5} (8.4\%) and Xiaomi-Robotics-0 (8.9\%) exhibit lower overestimation, showing more uniform robustness. \rev{PRIDE also predicts success rate more faithfully than general-purpose
NLP metrics such as METEOR~\cite{meteor}, BERTScore, and BLEU, achieving the strongest
correlation with SR (Pearson $|r|=0.802$) and ranking first on 6 of 7 VLAs:
the sole exception is the model whose low base-task performance confounds
the correlation (Tab.~\ref{tab:pride_vs_other},
Appendix~\ref{sec:appen_metric}).}

\begin{table}[t]
\centering
\resizebox{\columnwidth}{!}{
\begin{tabular}{lccc}
\toprule
Method & SR & PRIDE & \textbf{Overestimation (\%)} \\
\midrule
VLA-Adapter                          & 46.3 & 36.1 & \textbf{22.0} \\
$\pi$\textsubscript{0.5} (expert-only) & 39.1 & 32.0 & \textbf{18.2} \\
X-VLA                                & 62.1 & 52.7 & \textbf{15.1} \\
OpenVLA-OFT\textsubscript{mixed}     & 63.7 & 56.3 & \textbf{11.6} \\
OpenVLA-OFT\textsubscript{goal}      & 64.7 & 58.8 & \textbf{9.1}  \\
Xiaomi-Robotics-0                      & 76.0 & 69.2 & \textbf{8.9}  \\
$\pi$\textsubscript{0.5}             & 71.4 & 65.4 & \textbf{8.4}  \\
\bottomrule
\end{tabular}
}
\caption{SR and PRIDE scores on LIBERO-Para, sorted by overestimation. Overestimation is computed as (SR\,--\,PRIDE)\,/\,SR, indicating how much uniform success rate overstates a model's paraphrase robustness.}
\label{tab:pride_overestimation}
\end{table}

\begin{figure}[t]
\centering
\includegraphics[width=\columnwidth]{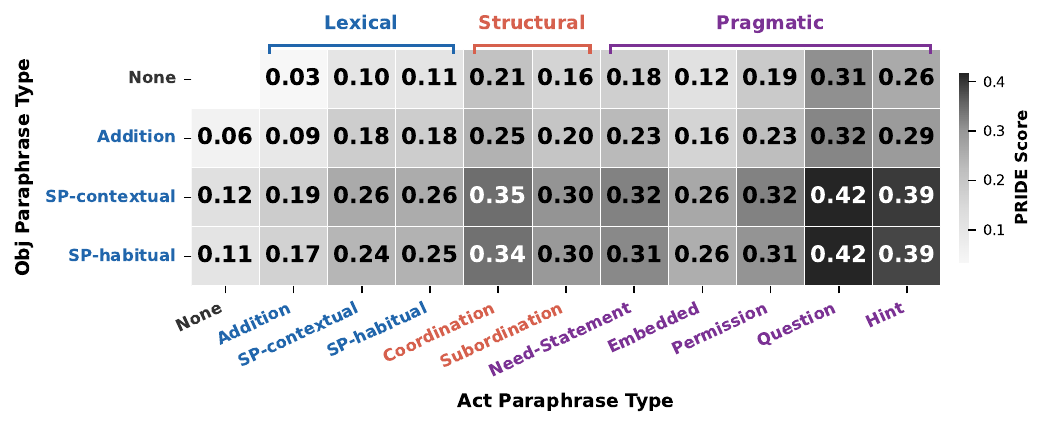}
\caption{Average PRIDE score per Object × Action cell in LIBERO-Para (darker = harder). Scores increase along both axes, with the most indirect action types (Question, Hint) combined with object paraphrasing reaching the highest (SP-habitual × Question: 0.42).}
\label{fig:pride_score_table}
\end{figure}
\begin{figure*}[t]
\centering
\includegraphics[width=\textwidth]{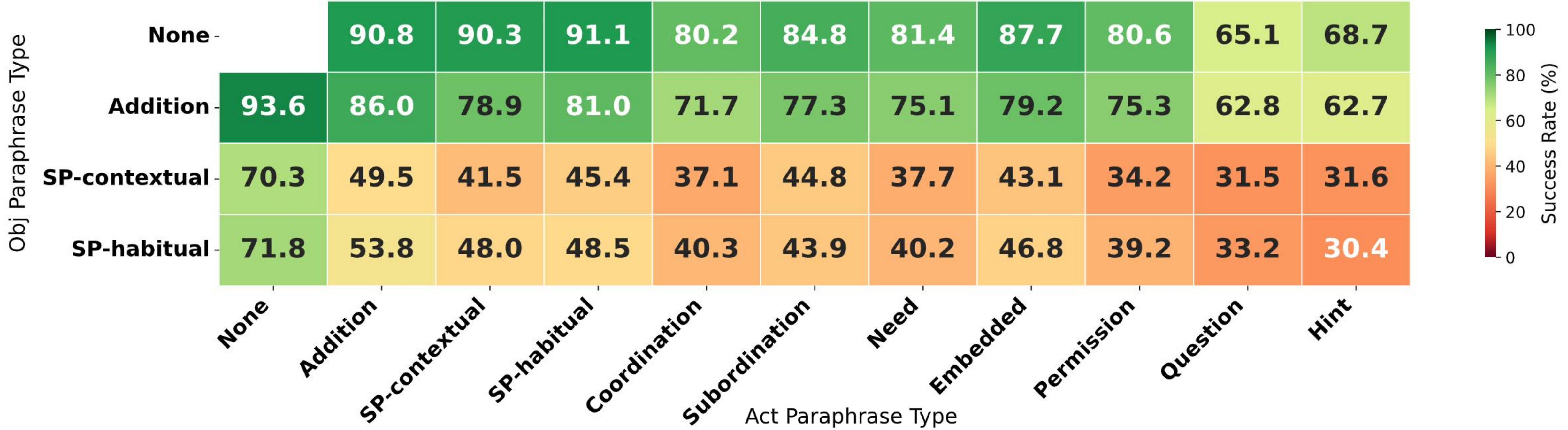}
\caption{Model-average success rate per Object × Action cell. Object-paraphrased rows drop sharply compared to object-preserved rows, reaching 30.4\% at SP-habitual × Hint.}
\label{fig:model_average_sr_table}
\end{figure*}
\begin{table}[t]
\centering
\small
\begin{tabular}{lccc}
\toprule
Cell group & $n$ & Mean SR (\%) & Drop \\
\midrule
Original (None $\times$ None)& 1 & 95.0 & \phantom{$-$}-- \\
Action-only (Obj: None)& 10 & 82.1 & \textcolor{red}{$-12.9$} \\
Object-only (Act: None)& 3 & 78.6 & \textcolor{red}{$-16.4$} \\
Compositional & 30 & 52.4 & \textcolor{red}{$-42.6$} \\
\bottomrule
\end{tabular}
\caption{\rev{Success rate (SR) by cell group ($n$: number of cells), averaged over 7 VLAs. Original from Tab.~\ref{tab:sr_drop}, others aggregated from Fig.~\ref{fig:model_average_sr_table}. Drop is relative to Original.}}
\label{tab:super_additive}
\end{table}
Fig.~\ref{fig:pride_score_table} and Fig.~\ref{fig:model_average_sr_table} confirm these trends at cell level: degradation intensifies along both axes, with the sharpest drops when object paraphrasing combines with indirect actions. Notably, the gap between object-preserved rows (None, Addition) and object-paraphrased rows (SP-contextual, SP-habitual) is substantially larger than the action-type gap within the same object condition, suggesting that object-level variation is a stronger driver of failure than action indirectness. \rev{Beyond this asymmetry, the two axes do not act independently. Tab.~\ref{tab:super_additive} decomposes success rates by cell partition: varying both axes jointly degrades performance by 42.6\,pp, exceeding the sum of the individual axis effects (12.9 + 16.4 = 29.3\,pp). This super-additive interaction is inaccessible to single-axis paraphrasing.} We investigate the object--action asymmetry and its underlying causes in Sec.~\ref{finding:finding2}--\ref{finding:finding3}, after first examining whether architecture or training choices mitigate the overall degradation (Sec.~\ref{finding:finding1}).


\section{Analysis}
\label{sec:analysis}
\subsection{Finding 1: Paraphrase Fragility Persists Across Architectures, Data Scales, and Fine-tuning Strategies}
\label{finding:finding1}
Before analyzing where and how failures occur in Sec.  ~\ref{finding:finding2} and~\ref{finding:finding3}, we examine whether paraphrase fragility can be attributed to specific factors by varying architecture family, training data scope, and VLM fine-tuning strategy.
\paragraph{Architecture Diversity.}
Across seven configurations spanning four architecture families (OpenVLA-OFT, $\pi$\textsubscript{0.5}/Xiaomi-Robotics-0, X-VLA, VLA-Adapter), all models show substantial success rate drops under paraphrasing, ranging from 22.8\,pp to 51.9\,pp (Tab.~\ref{tab:sr_drop}). The 7.5B OpenVLA-OFT shows PRIDE comparable to the 0.9B X-VLA. All models exhibit PRIDE overestimation of 8.4–22.0\% (Tab.~\ref{tab:pride_overestimation}). Overall, VLAs experience significant degradation under paraphrasing, regardless of architecture or scale.

\paragraph{Data Scope.}
OpenVLA-OFT\textsubscript{mixed} expands task-level data diversity by 4× compared to OpenVLA-OFT\textsubscript{goal} within the same architecture and simulator. However, both models exhibit similar success rate drops on LIBERO-Para (32.4\,pp vs.\ 33.2\,pp), suggesting that increasing task diversity through additional training samples does not improve robustness to linguistic variation in learned tasks.
\paragraph{VLM Training Strategy.}
We compare the standard $\pi$\textsubscript{0.5} model (jointly fine-tuning the VLM and Action Expert) with a variant that freezes the VLM of the base VLA and fine-tunes only the Action Expert. The frozen-VLM variant shows substantially lower performance on LIBERO-Goal (97.6 $\rightarrow$ 78.6; Tab.~\ref{tab:sr_drop}) and does not exhibit improved robustness on LIBERO-Para (SR: 39.1, PRIDE: 32.0; Tab.~\ref{tab:pride_overestimation}). Although the jointly fine-tuned model achieves higher success rates on LIBERO-Para (71.4 vs.\ 39.1), both variants still show substantial drops under paraphrasing. This suggests that joint adaptation of the VLM and Action Expert is essential for downstream performance, while fine-tuning on limited demonstrations may degrade pretrained semantics, causing paraphrase vulnerability.



Taken together, paraphrase fragility persists across all three factors. This indicates that the robustness gap cannot be explained solely by architecture, data scope, or fine-tuning strategy, but points to a deeper challenge. \textit{Which linguistic variations, then, are most responsible for these failures?}

\subsection{Finding 2: Object Grounding Emerges as a Primary Bottleneck}
\label{finding:finding2}
\begin{figure}[t]
\centering
\includegraphics[width=\columnwidth]{figures/Fig7.pdf}
\caption{Success rate comparison between object-preserved (None, Addition) and object-paraphrased (SP-contextual, SP-habitual) instructions. All models show substantial drops, from 19.8\,pp ($\pi_{0.5}$ expert-only) to 51.0\,pp (OpenVLA-OFT\textsubscript{mixed}). $\Delta$ annotated per pair.}
\label{fig:object_preserved_vs_paraphrased}
\end{figure}
Sec.~\ref{finding:finding1} shows that paraphrase fragility persists regardless of architecture, data scope, or fine-tuning strategy. We next examine where this degradation concentrates. Our analysis reveals an asymmetry: object-level variation emerges as the dominant source of failure, while action indirectness introduces additional degradation.
Fig.~\ref{fig:object_preserved_vs_paraphrased} compares success rates between object-preserved and object-paraphrased instructions across models. When the object is paraphrased—even through common synonyms such as replacing \textit{stove} with \textit{range}—performance drops by 19.8\,pp ($\pi$\textsubscript{0.5} expert-only) to 51.0\,pp (OpenVLA-OFT\textsubscript{mixed}) across models. This gap appears consistently across architectures, suggesting that current VLAs rely more on surface-level keyword matching than on semantic understanding of objects. Notably, OpenVLA-OFT\textsubscript{mixed}, trained with four times more tasks, exhibits nearly the same gap as OpenVLA-OFT\textsubscript{goal} (51.0\,pp vs.\ 48.3\,pp; Fig.~\ref{fig:object_preserved_vs_paraphrased}), indicating that task diversity and object-paraphrase robustness are decoupled. PRIDE $\alpha$ sweep experiments further confirm that keyword-level lexical variation around object references drives most of the degradation, compared to syntactic restructuring (Appendix~\ref{appen:Analysis_finding2}).
In addition, action indirectness introduces a stepwise performance decline: as instructions become less explicit, success rates drop from 82.7\% (None) to around 48\% (Question, Hint) (see Appendix~\ref{appen:Analysis_finding2} for the full action-axis breakdown).

This asymmetry reflects structural properties of tabletop manipulation. The action space is restricted to a small set of motor primitives (e.g., pick, place, push and open), and each object typically supports only a few feasible actions (e.g., \textit{stove} $\rightarrow$ turn on), allowing models to converge to the correct primitive even under varied phrasing. In contrast, the object space is much larger and lexically open-ended, concentrating combinatorial complexity on object references. This vulnerability may be amplified by current VLA training data, where objects are often referred to by a single canonical name (Fig.~\ref{fig:canonical_instructions}), making grounding sensitive even to simple synonym substitutions. These observations suggest that the primary bottleneck in paraphrase robustness lies in object grounding, with action indirectness introducing additional degradation.
Having identified the dominant factor behind these failures, we next ask: \textit{do these failures arise during execution of the correct task, or do models generate different trajectories from the outset?}

\begin{figure}[t]
\centering
\includegraphics[width=\columnwidth]{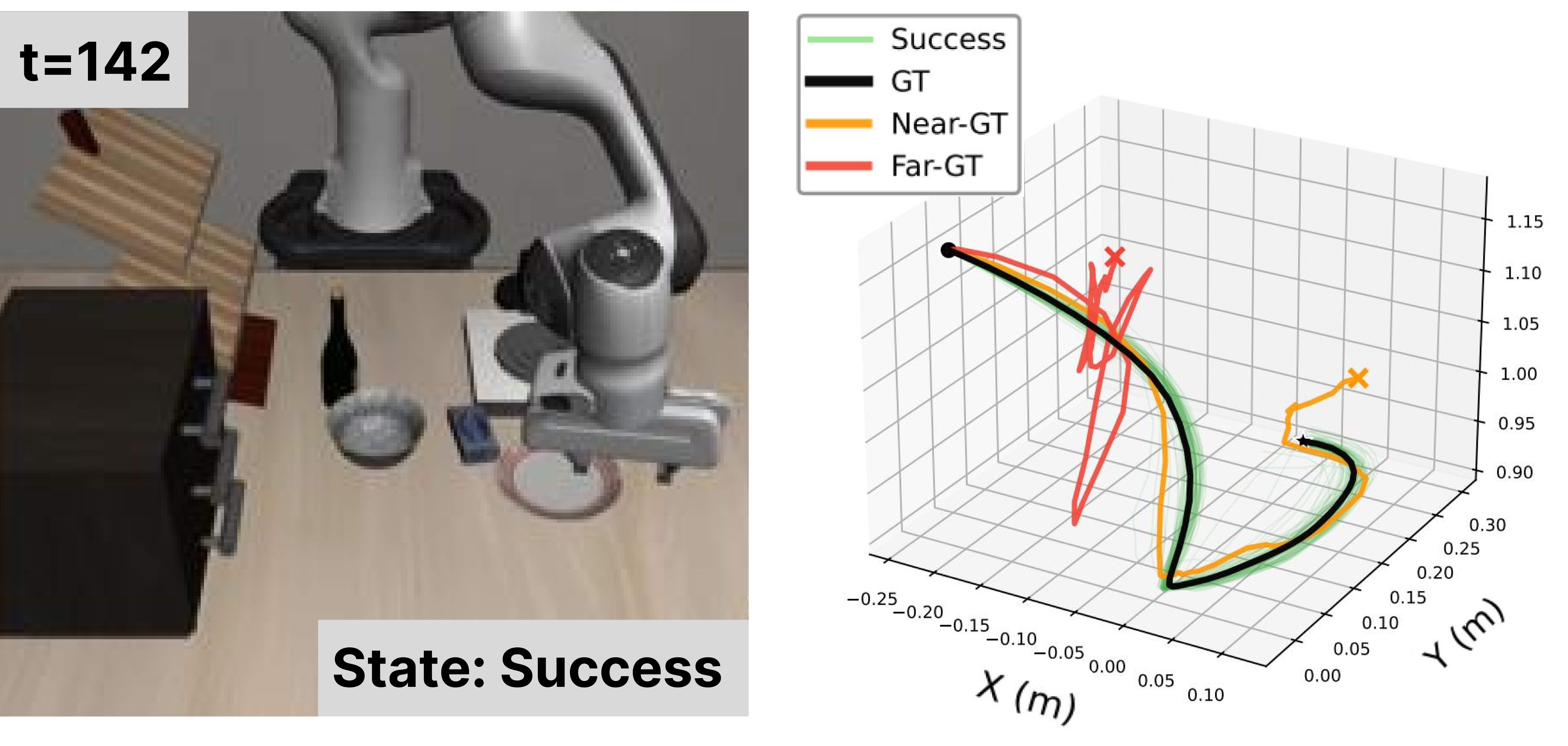}
\caption{(Left) LIBERO scene for Task 2: Push the plate to the front of the stove. (Right) 3D end-effector trajectories under a paraphrased instruction ($\pi$\textsubscript{0.5}). Green: successful episodes; black: their mean (GT); orange: Near-GT failure (tracks GT but fails); red: Far-GT failure (diverges early).}
\label{fig:nearGT_farGT_visualization}
\end{figure}
\subsection{Finding 3: Failures Are Predominantly Planning-Level, Not Execution-Level}
\label{finding:finding3}
\begin{table}[t]
\centering
\resizebox{\columnwidth}{!}{
\begin{tabular}{lcccc}
\toprule
\multirow{2}{*}{Model} & Success & \multicolumn{3}{c}{Failure} \\
\cmidrule(l){3-5}
 & Rate & Near-GT & Far-GT & \textbf{Far-GT(\%)} \\
\midrule
OpenVLA-OFT\textsubscript{goal}        & 64.7 &  1.6 & 33.7 & \textbf{95.5} \\
Xiaomi-Robotics-0                         & 76.0 &  1.8 & 22.2 & \textbf{92.5} \\
VLA-Adapter                             & 46.3 &  4.2 & 49.5 & \textbf{92.2} \\
$\pi$\textsubscript{0.5}               & 71.4 &  2.4 & 26.2 & \textbf{91.6} \\
OpenVLA-OFT\textsubscript{mixed}       & 63.7 &  3.3 & 33.0 & \textbf{90.9} \\
X-VLA                                   & 62.1 &  5.2 & 32.7 & \textbf{86.3} \\
$\pi$\textsubscript{0.5} (expert-only) & 39.1 & 12.5 & 48.4 & \textbf{79.5} \\
\bottomrule
\end{tabular}
}
\caption{Failure classification on LIBERO-Para (by Far-GT (\%)). Near-GT: execution-level failure near GT trajectory. Far-GT: planning-level failure from GT. Across models, 79.5--95.5\% of failures are planning-level.}
\label{tab:failure_classification}
\end{table}
Sec.~\ref{finding:finding2} identified object grounding as the primary bottleneck, with action indirectness introducing additional degradation. To determine whether failures arise during execution of the correct task or from generating entirely different trajectories from the outset, we classify failures based on trajectory similarity to successful executions. For each task, we define the mean successful trajectory as pseudo ground-truth (hereafter GT) and compute the Dynamic Time Warping (DTW) distance~\citep{sakoe1978dynamic} between each failure trajectory and the GT. Because LIBERO-Goal training data follows a fixed trajectory with minimal path variation, the mean success trajectory serves as a reliable pseudo GT. Failures within a threshold $\tau$ (the maximum DTW distance among successful episodes) are categorized as \textbf{Near-GT} (execution-level), indicating correct task execution but failure due to minor motor control errors. Failures exceeding $\tau$ are categorized as \textbf{Far-GT} (planning-level), indicating fundamentally different trajectories and thus failure in task identification. As shown in Fig.~\ref{fig:nearGT_farGT_visualization}, Near-GT trajectories remain close to successful ones, whereas Far-GT trajectories diverge substantially. Additional methodological details are provided in Appendix~\ref{appen:Analysis_finding3}.

Tab.~\ref{tab:failure_classification} summarizes the classification results. Across models, 79.5\%–95.5\% of failures are Far-GT, while Near-GT cases account for less than 5\% in most models. This indicates that under paraphrased instructions, models rarely fail along the correct trajectory but instead generate different trajectories from the outset. The only exception is $\pi$\textsubscript{0.5} expert-only (Near-GT: 12.5\%), where the frozen VLM may identify the task correctly but the non-adapted Action Expert fails to execute it precisely, consistent with the VLM training strategy analysis in Sec.~\ref{finding:finding1}. These findings align with Sec.~\ref{finding:finding2}: when object grounding fails, the model plans actions toward incorrect targets, producing trajectories that diverge from the GT. The dominant failure mode lies in task identification rather than motor control, suggesting that improving paraphrase robustness should focus on instruction semantic-to-task mapping rather than action-execution control.

\paragraph{\rev{Cross-Benchmark Validation.}}
\rev{Finally, to assess whether our findings generalize beyond LIBERO, we construct
CALVIN-Para by applying the same two-axis design to CALVIN~\citep{calvin}, a
long-horizon language-conditioned benchmark with a distinct simulator, task
suite, and instruction style. We evaluate two CALVIN-native VLAs,
RoboFlamingo~\citep{roboflamingo} and FLOWER~\citep{flower}, whose
architectures differ from all seven LIBERO configurations. We confirm that all
three findings consistently hold for these models. Despite near-saturated
canonical performance, both models collapse under paraphrasing
(92.3\%$\to$53.1\% and 99.7\%$\to$58.3\%), extending Finding 1 to two
additional architectures. Object-axis paraphrasing degrades SR by 32.6 and
37.7\,pp, reproducing the object bottleneck of Finding 2. Likewise, 93.4\% and
97.1\% of failures are planning-level (Far-GT), matching Finding 3. SR also
overstates robustness by 18.1\% and 17.2\% relative to PRIDE, confirming that
PRIDE captures hidden severity. Setup and results are in
Appendix~\ref{sec:calvinpara}.}

\section{Conclusion}


This work investigates paraphrase robustness in modern VLA models using LIBERO-Para, a controlled benchmark that independently varies action and object expressions, and PRIDE, a fine-grained robustness metric. We find that paraphrase fragility persists across architectures, scales, and fine-tuning strategies, with object-level lexical variation as the dominant source of degradation and 80–96\% of failures arising from planning-level trajectory divergence rather than execution errors. \rev{All three findings reproduce on CALVIN-Para, indicating they are not artifacts of a single benchmark.} These results reveal a fundamental limitation: current VLA models map instructions to task identification by surface-level matching rather than robust object grounding. Improving paraphrase robustness therefore requires prioritizing instruction-to-task identification over control refinement, with object grounding as a key direction.

\section*{Limitations}


We evaluate VLA models only in the LIBERO simulator, which differs
from physical platforms in rendering, physics, and sensor noise, so whether
these vulnerabilities persist in the real world is untested. Second, we vary
one type per axis at a time, leaving compound variations to future work.
Third, we do not study paraphrase-based augmentation, since LLM-generated
paraphrases could overlap distributionally with the benchmark. \rev{Finally,
we do not probe attention or intermediate representations: the evaluated
models span four architecture families differing in VLM--action-expert
coupling, attention scheme, backbone, and scale
(Tab.~\ref{tab:appen_model_spec}), so single-model probing characterizes
that model rather than VLAs broadly. Testing the structural hypothesis of
Sec.~\ref{finding:finding2} is left to future work.}

\section*{Acknowledgments} 

This work was supported by the National Research Foundation of Korea (NRF) grant (RS-2025-00555943); by the Institute of Information \& Communications Technology Planning \& Evaluation (IITP) grants (No. RS-2021-II211341; Artificial Intelligence Graduate School Program (Chung-Ang University), RS-2026-25513331; Digital Columbus Project, and IITP-2026-RS-2026-25546026; Leading Generative AI Human Resources Development) funded by the Korea government (MSIT).


\clearpage










\bibliography{custom}

\begin{thebibliography}{51}
\providecommand{\natexlab}[1]{#1}

\bibitem[{Alayrac et~al.(2022)Alayrac, Donahue, Luc, Miech, Barr, Hasson, Lenc,
  Mensch, Millican, Reynolds et~al.}]{flamingo}
Jean-Baptiste Alayrac, Jeff Donahue, Pauline Luc, Antoine Miech, Iain Barr,
  Yana Hasson, Karel Lenc, Arthur Mensch, Katherine Millican, Malcolm Reynolds,
  and 1 others. 2022.
\newblock Flamingo: a visual language model for few-shot learning.
\newblock \emph{Advances in neural information processing systems},
  35:23716--23736.

\bibitem[{Anthropic(2025)}]{anthropic2025claude4}
Anthropic. 2025.
\newblock System card: Claude opus 4 \& claude sonnet 4.
\newblock Anthropic system card.

\bibitem[{Augsten and B{\"o}hlen(2013)}]{augstenboehlen2013similarityjoins}
Nikolaus Augsten and Michael~H. B{\"o}hlen. 2013.
\newblock \href {https://doi.org/10.2200/S00586ED1V01Y201311DTM040}
  {\emph{Similarity Joins in Relational Database Systems}}.
\newblock Synthesis Lectures on Data Management. Morgan \& Claypool Publishers.

\bibitem[{Bai et~al.(2025)Bai, Team et~al.}]{qwen3vl}
Shuai Bai, Qwen Team, and 1 others. 2025.
\newblock \href {https://arxiv.org/abs/2511.21631} {Qwen3-vl technical report}.
\newblock \emph{arXiv preprint arXiv:2511.21631}.

\bibitem[{Banerjee and Lavie(2005)}]{meteor}
Satanjeev Banerjee and Alon Lavie. 2005.
\newblock Meteor: An automatic metric for mt evaluation with improved
  correlation with human judgments.
\newblock In \emph{Proceedings of the acl workshop on intrinsic and extrinsic
  evaluation measures for machine translation and/or summarization}, pages
  65--72.

\bibitem[{Black et~al.(2024)Black, Brown, Driess, Esmail, Equi, Finn, Fusai,
  Groom, Hausman, Ichter, Jakubczak, Jones, Ke, Levine, Li-Bell, Mothukuri,
  Nair, Pertsch, Shi, Tanner, Vuong, Walling, Wang, Zhilinsky et~al.}]{pi0}
Kevin Black, Noah Brown, Danny Driess, Adnan Esmail, Michael Equi, Chelsea
  Finn, Niccolo Fusai, Lachy Groom, Karol Hausman, Brian Ichter, Szymon
  Jakubczak, Tim Jones, Liyiming Ke, Sergey Levine, Adrian Li-Bell, Mohith
  Mothukuri, Suraj Nair, Karl Pertsch, Lucy~Xiaoyang Shi, and 6 others. 2024.
\newblock \href {https://arxiv.org/abs/2410.24164} {{$\pi_0$}: A
  vision-language-action flow model for general robot control}.
\newblock \emph{arXiv preprint arXiv:2410.24164}.

\bibitem[{Bu et~al.(2025)Bu, Cai, Chen, Cui, Ding, Feng, Gao, He, Hu, Huang
  et~al.}]{agibot}
Qingwen Bu, Jisong Cai, Li~Chen, Xiuqi Cui, Yan Ding, Siyuan Feng, Shenyuan
  Gao, Xindong He, Xuan Hu, Xu~Huang, and 1 others. 2025.
\newblock Agibot world colosseo: A large-scale manipulation platform for
  scalable and intelligent embodied systems.
\newblock \emph{arXiv preprint arXiv:2503.06669}.

\bibitem[{Cai et~al.(2026)Cai, Guo, He, Jin, Li, Lin, Liu, Liu, Ma, Ma
  et~al.}]{cai2026xiaomirobotics}
Rui Cai, Jun Guo, Xinze He, Piaopiao Jin, Jie Li, Bingxuan Lin, Futeng Liu, Wei
  Liu, Fei Ma, Kun Ma, and 1 others. 2026.
\newblock Xiaomi-robotics-0: An open-sourced vision-language-action model with
  real-time execution.
\newblock \emph{arXiv preprint arXiv:2602.12684}.

\bibitem[{Comanici et~al.(2025)Comanici, Bieber, Schaekermann, Pasupat,
  Sachdeva, Dhillon, Blistein, Ram, Zhang, Rosen et~al.}]{gemini}
Gheorghe Comanici, Eric Bieber, Mike Schaekermann, Ice Pasupat, Noveen
  Sachdeva, Inderjit Dhillon, Marcel Blistein, Ori Ram, Dan Zhang, Evan Rosen,
  and 1 others. 2025.
\newblock Gemini 2.5: Pushing the frontier with advanced reasoning,
  multimodality, long context, and next generation agentic capabilities.
\newblock \emph{arXiv preprint arXiv:2507.06261}.

\bibitem[{Ervin-Tripp(1976)}]{ervin-tripp1976}
Susan Ervin-Tripp. 1976.
\newblock \href {https://doi.org/10.1017/S0047404500006849} {Is {S}ybil there?
  {T}he structure of some {A}merican {E}nglish directives}.
\newblock \emph{Language in Society}, 5(1):25--66.

\bibitem[{Fei et~al.(2025)Fei, Wang, Shi, Dai, Cai, Qian, Ji, He, Zhang, Fei,
  Fu, Gong, and Qiu}]{libero_plus}
Senyu Fei, Siyin Wang, Junhao Shi, Zihao Dai, Jikun Cai, Pengfang Qian, Li~Ji,
  Xinzhe He, Shiduo Zhang, Zhaoye Fei, Jinlan Fu, Jingjing Gong, and Xipeng
  Qiu. 2025.
\newblock \href {https://arxiv.org/abs/2510.13626} {Libero-plus: In-depth
  robustness analysis of vision-language-action models}.
\newblock \emph{arXiv preprint arXiv:2510.13626}.

\bibitem[{Feinstein and Cicchetti(1990)}]{feinstein1990high}
Alvan~R Feinstein and Domenic~V Cicchetti. 1990.
\newblock High agreement but low kappa: I. the problems of two paradoxes.
\newblock \emph{Journal of clinical epidemiology}, 43(6):543--549.

\bibitem[{Fu et~al.(2024)Fu, Zhao, and Finn}]{mobilealoha}
Zipeng Fu, Tony~Z. Zhao, and Chelsea Finn. 2024.
\newblock \href {https://arxiv.org/abs/2401.02117} {Mobile {ALOHA}: Learning
  bimanual mobile manipulation with low-cost whole-body teleoperation}.
\newblock \emph{arXiv preprint arXiv:2401.02117}.

\bibitem[{Gwet(2008)}]{gwet2008computing}
Kilem~Li Gwet. 2008.
\newblock Computing inter-rater reliability and its variance in the presence of
  high agreement.
\newblock \emph{British Journal of Mathematical and Statistical Psychology},
  61(1):29--48.

\bibitem[{Hochreiter and Schmidhuber(1997)}]{lstm}
Sepp Hochreiter and J{\"u}rgen Schmidhuber. 1997.
\newblock Long short-term memory.
\newblock \emph{Neural computation}, 9(8):1735--1780.

\bibitem[{Hou and Zhao(2026)}]{hou2026langgap}
Yuchen Hou and Lin Zhao. 2026.
\newblock Langgap: Diagnosing and closing the language gap in
  vision-language-action models.
\newblock \emph{arXiv preprint arXiv:2603.00592}.

\bibitem[{Karamcheti et~al.(2024)Karamcheti, Nair, Balakrishna, Liang, Kollar,
  and Sadigh}]{karamcheti2024prismatic}
Siddharth Karamcheti, Suraj Nair, Ashwin Balakrishna, Percy Liang, Thomas
  Kollar, and Dorsa Sadigh. 2024.
\newblock Prismatic vlms: Investigating the design space of
  visually-conditioned language models.
\newblock In \emph{International Conference on Machine Learning (ICML)}.

\bibitem[{Khazatsky et~al.(2024)Khazatsky, Pertsch, Nair, Balakrishna, Dasari,
  Karamcheti, Nasiriany, Srirama, Chen, Ellis et~al.}]{droid}
Alexander Khazatsky, Karl Pertsch, Suraj Nair, Ashwin Balakrishna, Sudeep
  Dasari, Siddharth Karamcheti, Soroush Nasiriany, Mohan~Kumar Srirama,
  Lawrence~Yunliang Chen, Kirsty Ellis, and 1 others. 2024.
\newblock Droid: A large-scale in-the-wild robot manipulation dataset.
\newblock \emph{arXiv preprint arXiv:2403.12945}.

\bibitem[{Kim et~al.(2025)Kim, Finn, and Liang}]{openvla-oft}
Moo~Jin Kim, Chelsea Finn, and Percy Liang. 2025.
\newblock \href {https://doi.org/10.48550/arXiv.2502.19645} {Fine-tuning
  vision-language-action models: Optimizing speed and success}.
\newblock \emph{arXiv preprint arXiv:2502.19645}.

\bibitem[{Kim et~al.(2024)Kim, Pertsch, Karamcheti, Xiao, Balakrishna, Nair,
  Rafailov, Foster, Lam, Sanketi, Vuong, Kollar, Burchfiel, Tedrake, Sadigh,
  Levine, Liang, and Finn}]{openvla}
Moo~Jin Kim, Karl Pertsch, Siddharth Karamcheti, Ted Xiao, Ashwin Balakrishna,
  Suraj Nair, Rafael Rafailov, Ethan Foster, Grace Lam, Pannag Sanketi, Quan
  Vuong, Thomas Kollar, Benjamin Burchfiel, Russ Tedrake, Dorsa Sadigh, Sergey
  Levine, Percy Liang, and Chelsea Finn. 2024.
\newblock \href {https://doi.org/10.48550/arXiv.2406.09246} {Openvla: An
  open-source vision-language-action model}.
\newblock \emph{arXiv preprint arXiv:2406.09246}.

\bibitem[{Kovatchev et~al.(2018)Kovatchev, Mart{\'i}, and
  Salam{\'o}}]{kovatchev-etal-2018-etpc}
Venelin Kovatchev, M.~Ant{\`o}nia Mart{\'i}, and Maria Salam{\'o}. 2018.
\newblock \href {https://aclanthology.org/L18-1221/} {{ETPC} - a paraphrase
  identification corpus annotated with extended paraphrase typology and
  negation}.
\newblock In \emph{Proceedings of the Eleventh International Conference on
  Language Resources and Evaluation ({LREC} 2018)}, Miyazaki, Japan. European
  Language Resources Association (ELRA).

\bibitem[{Li et~al.(2024)Li, Liu, Zhang, Yu, Xu, Wu, Cheang, Jing, Zhang, Liu,
  Li, and Kong}]{roboflamingo}
Xinghang Li, Minghuan Liu, Hanbo Zhang, Cunjun Yu, Jie Xu, Hongtao Wu, Chilam
  Cheang, Ya~Jing, Weinan Zhang, Huaping Liu, Hang Li, and Tao Kong. 2024.
\newblock Vision-language foundation models as effective robot imitators.
\newblock In \emph{International Conference on Learning Representations
  (ICLR)}.

\bibitem[{Liu et~al.(2023)Liu, Zhu, Gao, Feng, Liu, Zhu, and Stone}]{libero}
Bo~Liu, Yifeng Zhu, Chongkai Gao, Yihao Feng, Qiang Liu, Yuke Zhu, and Peter
  Stone. 2023.
\newblock \href {https://arxiv.org/abs/2306.03310} {Libero: Benchmarking
  knowledge transfer for lifelong robot learning}.
\newblock In \emph{Advances in Neural Information Processing Systems (NeurIPS),
  Datasets and Benchmarks Track}, volume~36, pages 44776--44791.

\bibitem[{Liu et~al.(2022)Liu, Gong, and Liu}]{rectifiedflow}
Xingchao Liu, Chengyue Gong, and Qiang Liu. 2022.
\newblock Flow straight and fast: Learning to generate and transfer data with
  rectified flow.
\newblock \emph{arXiv preprint arXiv:2209.03003}.

\bibitem[{Loshchilov and Hutter(2017)}]{adamw}
Ilya Loshchilov and Frank Hutter. 2017.
\newblock Decoupled weight decay regularization.
\newblock \emph{arXiv preprint arXiv:1711.05101}.

\bibitem[{Mees et~al.(2021)Mees, Hermann, Rosete-Beas, and Burgard}]{calvin}
Oier Mees, Lukas Hermann, Erick Rosete-Beas, and Wolfram Burgard. 2021.
\newblock \href {https://arxiv.org/abs/2112.03227} {{CALVIN}: A benchmark for
  language-conditioned policy learning for long-horizon robot manipulation
  tasks}.
\newblock \emph{arXiv preprint arXiv:2112.03227}.

\bibitem[{Miller(1995)}]{miller1995wordnet}
George~A Miller. 1995.
\newblock Wordnet: a lexical database for english.
\newblock \emph{Communications of the ACM}, 38(11):39--41.

\bibitem[{OpenAI(2023)}]{openai2023gpt4}
OpenAI. 2023.
\newblock Gpt-4 technical report.
\newblock \emph{arXiv preprint arXiv:2303.08774}.

\bibitem[{O’Neill et~al.(2024)O’Neill, Rehman, Maddukuri, Gupta, Padalkar,
  Lee, Pooley, Gupta, Mandlekar, Jain et~al.}]{oxe}
Abby O’Neill, Abdul Rehman, Abhiram Maddukuri, Abhishek Gupta, Abhishek
  Padalkar, Abraham Lee, Acorn Pooley, Agrim Gupta, Ajay Mandlekar, Ajinkya
  Jain, and 1 others. 2024.
\newblock Open x-embodiment: Robotic learning datasets and rt-x models: Open
  x-embodiment collaboration 0.
\newblock In \emph{2024 IEEE International Conference on Robotics and
  Automation (ICRA)}, pages 6892--6903. IEEE.

\bibitem[{Papineni et~al.(2002)Papineni, Roukos, Ward, and Zhu}]{bleu}
Kishore Papineni, Salim Roukos, Todd Ward, and Wei-Jing Zhu. 2002.
\newblock Bleu: a method for automatic evaluation of machine translation.
\newblock In \emph{Proceedings of the 40th annual meeting of the Association
  for Computational Linguistics}, pages 311--318.

\bibitem[{Peebles and Xie(2023)}]{diffusiontransformer}
William Peebles and Saining Xie. 2023.
\newblock Scalable diffusion models with transformers.
\newblock In \emph{2023 IEEE/CVF International Conference on Computer Vision
  (ICCV)}, pages 4172--4182. IEEE.

\bibitem[{{Physical Intelligence} et~al.(2025){Physical Intelligence}, Black,
  Brown, Darpinian, Dhabalia, Driess, Esmail, Equi, Finn, Fusai, Galliker,
  Ghosh, Groom, Hausman, Ichter, Jakubczak, Jones, Ke, LeBlanc, Levine,
  Li-Bell, Mothukuri, Nair, Pertsch, Ren, Shi, Smith, Springenberg, Stachowicz,
  Tanner, Vuong, Walke, Walling, Wang, Yu, Zhilinsky et~al.}]{pi05}
{Physical Intelligence}, Kevin Black, Noah Brown, James Darpinian, Karan
  Dhabalia, Danny Driess, Adnan Esmail, Michael Equi, Chelsea Finn, Niccolo
  Fusai, Manuel~Y. Galliker, Dibya Ghosh, Lachy Groom, Karol Hausman, Brian
  Ichter, Szymon Jakubczak, Tim Jones, Liyiming Ke, Devin LeBlanc, and 18
  others. 2025.
\newblock \href {https://arxiv.org/abs/2504.16054} {{$\pi_{0.5}$}: a
  vision-language-action model with open-world generalization}.
\newblock \emph{arXiv preprint arXiv:2504.16054}.

\bibitem[{Reimers and Gurevych(2019)}]{sentence}
Nils Reimers and Iryna Gurevych. 2019.
\newblock Sentence-bert: Sentence embeddings using siamese bert-networks.
\newblock In \emph{Proceedings of the 2019 Conference on Empirical Methods in
  Natural Language Processing and the 9th International Joint Conference on
  Natural Language Processing (EMNLP-IJCNLP)}, pages 3982--3992.

\bibitem[{Reuss et~al.(2025)Reuss, Zhou, R{\"u}hle, Ya{\u{g}}murlu, Otto, and
  Lioutikov}]{flower}
Moritz Reuss, Hongyi Zhou, Marcel R{\"u}hle, {\"O}mer~Erdin{\c{c}}
  Ya{\u{g}}murlu, Fabian Otto, and Rudolf Lioutikov. 2025.
\newblock Flower: Democratizing generalist robot policies with efficient
  vision-language-action flow policies.
\newblock \emph{arXiv preprint arXiv:2509.04996}.

\bibitem[{Sakoe and Chiba(1978)}]{sakoe1978dynamic}
Hiroaki Sakoe and Seibi Chiba. 1978.
\newblock \href {https://doi.org/10.1109/TASSP.1978.1163055} {Dynamic
  programming algorithm optimization for spoken word recognition}.
\newblock \emph{IEEE Transactions on Acoustics, Speech, and Signal Processing},
  26(1):43--49.

\bibitem[{Salvador and Chan(2007)}]{fastdtw}
Stan Salvador and Philip Chan. 2007.
\newblock Toward accurate dynamic time warping in linear time and space.
\newblock \emph{Intelligent data analysis}, 11(5):561--580.

\bibitem[{Steiner et~al.(2024)Steiner, Pinto, Tschannen, Keysers, Wang, Bitton,
  Gritsenko, Minderer, Sherbondy, Long, Qin, Ingle, Bugliarello, Kazemzadeh,
  Mesnard, Alabdulmohsin, Beyer, and Zhai}]{paligemma2}
Andreas Steiner, Andr{\'e}~Susano Pinto, Michael Tschannen, Daniel Keysers,
  Xiao Wang, Yonatan Bitton, Alexey Gritsenko, Matthias Minderer, Anthony
  Sherbondy, Shangbang Long, Siyang Qin, Reeve Ingle, Emanuele Bugliarello,
  Sahar Kazemzadeh, Thomas Mesnard, Ibrahim Alabdulmohsin, Lucas Beyer, and
  Xiaohua Zhai. 2024.
\newblock \href {https://arxiv.org/abs/2412.03555} {Paligemma 2: A family of
  versatile vlms for transfer}.
\newblock \emph{arXiv preprint arXiv:2412.03555}.

\bibitem[{Team(2024)}]{qwen25}
Qwen Team. 2024.
\newblock \href {https://arxiv.org/abs/2412.15115} {Qwen2.5 technical report}.
\newblock \emph{arXiv preprint arXiv:2412.15115}.

\bibitem[{Touvron et~al.(2023)Touvron, Martin, Stone, Albert, Almahairi,
  Babaei, Bashlykov, Batra, Bhargava, Bhosale et~al.}]{touvron2023llama2}
Hugo Touvron, Louis Martin, Kevin Stone, Peter Albert, Amjad Almahairi, Yasmine
  Babaei, Nikolay Bashlykov, Soumya Batra, Prajjwal Bhargava, Shruti Bhosale,
  and 1 others. 2023.
\newblock \href {https://arxiv.org/abs/2307.09288} {Llama 2: Open foundation
  and fine-tuned chat models}.
\newblock \emph{arXiv preprint arXiv:2307.09288}.

\bibitem[{Wang et~al.(2026)Wang, Zhang, Liu, Zhang, Cai, Liu, and
  Liu}]{wang2026libero-x}
Guodong Wang, Chenkai Zhang, Qingjie Liu, Jinjin Zhang, Jiancheng Cai, Junjie
  Liu, and Xinmin Liu. 2026.
\newblock Libero-x: Robustness litmus for vision-language-action models.
\newblock \emph{arXiv preprint arXiv:2602.06556}.

\bibitem[{Wang et~al.(2024{\natexlab{a}})Wang, Yang, Huang, Yang, Majumder, and
  Wei}]{e5}
Liang Wang, Nan Yang, Xiaolong Huang, Linjun Yang, Rangan Majumder, and Furu
  Wei. 2024{\natexlab{a}}.
\newblock Multilingual e5 text embeddings: A technical report.
\newblock \emph{arXiv preprint arXiv:2402.05672}.

\bibitem[{Wang et~al.(2025)Wang, Ding, Li, Cui, Ge, Tong, Song, Zhao, Zhao, Hou
  et~al.}]{wang2025vlaadapter}
Yihao Wang, Pengxiang Ding, Lingxiao Li, Can Cui, Zirui Ge, Xinyang Tong,
  Wenxuan Song, Han Zhao, Wei Zhao, Pengxu Hou, and 1 others. 2025.
\newblock Vla-adapter: An effective paradigm for tiny-scale
  vision-language-action model.
\newblock \emph{arXiv preprint arXiv:2509.09372}.

\bibitem[{Wang et~al.(2024{\natexlab{b}})Wang, Zhou, Song, Huang, Shu, and
  Ma}]{ladev}
Zhijie Wang, Zhehua Zhou, Jiayang Song, Yuheng Huang, Zhan Shu, and Lei Ma.
  2024{\natexlab{b}}.
\newblock \href {https://arxiv.org/abs/2410.05191} {{LADEV}: A language-driven
  testing and evaluation platform for vision-language-action models in robotic
  manipulation}.
\newblock \emph{arXiv preprint arXiv:2410.05191}.

\bibitem[{Wu et~al.(2023)Wu, Antonova, Kan, Lepert, Zeng, Song, Bohg,
  Rusinkiewicz, and Funkhouser}]{tidybot}
Jimmy Wu, Rika Antonova, Adam Kan, Marion Lepert, Andy Zeng, Shuran Song,
  Jeannette Bohg, Szymon Rusinkiewicz, and Thomas Funkhouser. 2023.
\newblock \href {https://arxiv.org/abs/2305.05658} {Tidybot: Personalized robot
  assistance with large language models}.
\newblock \emph{arXiv preprint arXiv:2305.05658}.

\bibitem[{Wu et~al.(2024)Wu, Hou, Liu, Che, Ju, Yang, Li, Zhao, Xu, Yang
  et~al.}]{robomind}
Kun Wu, Chengkai Hou, Jiaming Liu, Zhengping Che, Xiaozhu Ju, Zhuqin Yang, Meng
  Li, Yinuo Zhao, Zhiyuan Xu, Guang Yang, and 1 others. 2024.
\newblock Robomind: Benchmark on multi-embodiment intelligence normative data
  for robot manipulation.
\newblock \emph{arXiv preprint arXiv:2412.13877}.

\bibitem[{Xiao et~al.(2024)Xiao, Wu, Xu, Dai, Hu, Lu, Zeng, Liu, and
  Yuan}]{xiao2024florence2}
Bin Xiao, Haiping Wu, Weijian Xu, Xiyang Dai, Houdong Hu, Yumao Lu, Michael
  Zeng, Ce~Liu, and Lu~Yuan. 2024.
\newblock Florence-2: Advancing a unified representation for a variety of
  vision tasks.
\newblock In \emph{Proceedings of the IEEE/CVF Conference on Computer Vision
  and Pattern Recognition (CVPR)}, pages 4818--4829.

\bibitem[{Yadav et~al.(2025)Yadav, Zhou, Wagenmaker, Pertsch, and
  Levine}]{retain}
Yajat Yadav, Zhiyuan Zhou, Andrew Wagenmaker, Karl Pertsch, and Sergey Levine.
  2025.
\newblock \href {https://arxiv.org/abs/2512.08333} {Robust finetuning of
  vision-language-action robot policies via parameter merging}.
\newblock \emph{arXiv preprint arXiv:2512.08333}.

\bibitem[{Zhang et~al.(2019)Zhang, Kishore, Wu, Weinberger, and
  Artzi}]{bertscore}
Tianyi Zhang, Varsha Kishore, Felix Wu, Kilian~Q Weinberger, and Yoav Artzi.
  2019.
\newblock Bertscore: Evaluating text generation with bert.
\newblock \emph{arXiv preprint arXiv:1904.09675}.

\bibitem[{Zheng et~al.(2025)Zheng, Li, Wang, Liu, Kang, Feng, Zheng, Zou, Chen,
  Zeng et~al.}]{zheng2025xvla}
Jinliang Zheng, Jianxiong Li, Zhihao Wang, Dongxiu Liu, Xirui Kang, Yuchun
  Feng, Yinan Zheng, Jiayin Zou, Yilun Chen, Jia Zeng, and 1 others. 2025.
\newblock X-vla: Soft-prompted transformer as scalable cross-embodiment
  vision-language-action model.
\newblock \emph{arXiv preprint arXiv:2510.10274}.

\bibitem[{Zhou et~al.(2025)Zhou, Xu, Tie, Chen, Zhang, Chu, Zhou, and
  Sun}]{libero_pro}
Xueyang Zhou, Yangming Xu, Guiyao Tie, Yongchao Chen, Guowen Zhang, Duanfeng
  Chu, Pan Zhou, and Lichao Sun. 2025.
\newblock \href {https://arxiv.org/abs/2510.03827} {{LIBERO-PRO}: Towards
  robust and fair evaluation of vision-language-action models beyond
  memorization}.
\newblock \emph{arXiv preprint arXiv:2510.03827}.

\bibitem[{Zitkovich et~al.(2023)}]{rt2}
Brianna Zitkovich and 1 others. 2023.
\newblock \href {https://proceedings.mlr.press/v229/zitkovich23a.html} {{RT-2}:
  Vision-language-action models transfer web knowledge to robotic control}.
\newblock In \emph{Proceedings of The 7th Conference on Robot Learning (CoRL)},
  volume 229 of \emph{Proceedings of Machine Learning Research}, pages
  2165--2183.

\end{thebibliography}



\clearpage

\appendix

\onecolumn
\begin{leftline}
	{
		\LARGE{\textsc{Appendix}}
	}
\end{leftline}

	\etocdepthtag.toc{mtappendix}
    \etocsettagdepth{mtchapter}{none}
    \etocsettagdepth{mtappendix}{subsection}
    
    {
        \hypersetup{linkcolor=black}
    	\large\tableofcontents
    }
\clearpage

\twocolumn
\section{LIBERO-Para: A Controlled VLA Benchmark for Paraphrase Robustness}
\label{app:taxonomy}

\begin{figure*}[t]
    \centering
    \includegraphics[width=1\linewidth]{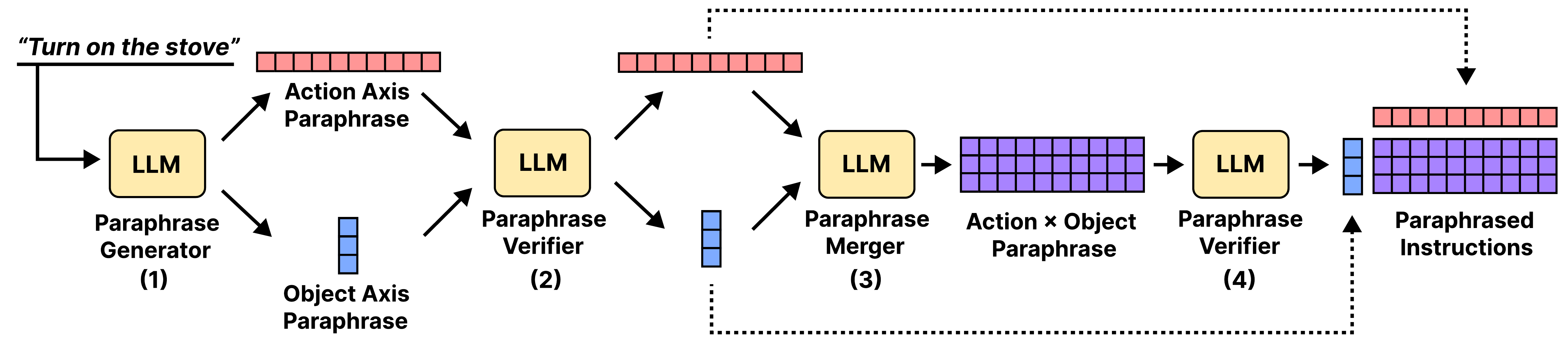}
    \caption{Overview of the LIBERO-Para dataset generation workflow. The process consists of four stages: (1) axis-wise paraphrase generation, (2) verification, (3) merging, and (4) final verification.}
    \label{fig:work-flow}
\end{figure*}

This appendix details the paraphrase taxonomy adopted in LIBERO-Para and explains the rationale for excluding certain types from the source taxonomies (EPT and Directive Types).
Our taxonomy is grounded in the Extended Paraphrase Typology (EPT)~\citep{kovatchev-etal-2018-etpc} and the Directive types proposed by \citet{ervin-tripp1976}.
From the 26 atomic paraphrase types in EPT and the six different types in Directive Types, we select 13 types that satisfy the following criteria:
(i) applicability to robotic manipulation instructions (i.e., direct imperatives),
(ii) preservation of the original meaning,
(iii) compliance with visual and spatial constraints, and
(iv) grammatical naturalness.
All paraphrases are generated under these constraints and are used exclusively for evaluation.

\begin{table}[t]
\centering
\footnotesize
\begin{tabular}{lll}
\hline
\textbf{Category} & \textbf{Type} & \textbf{Source (Year)} \\
\hline
\multirow{3}{*}{Obj-Lexical} 
  & same polarity habitual    & \multirow{3}{*}{EPT (2018)} \\
  & Same polarity contextual  &  \\
  & Addition                  &  \\
\hline
\multirow{3}{*}{Act-Lexical} 
  & Same polarity habitual    & \multirow{3}{*}{EPT (2018)} \\
  & Same polarity contextual  &  \\
  & Addition                  &  \\
\hline
\multirow{2}{*}{Act-Structural} 
  & Coordination              & \multirow{2}{*}{EPT (2018)} \\
  & Subordination             &  \\
\hline
\multirow{5}{*}{Act-Pragmatic} 
  & Personal need             & \multirow{5}{*}{\shortstack[l]{Directive\\ Types\\(1976)}} \\
  & Question directive        &  \\
  & Embedded imperative       &  \\
  & Permission                &  \\
  & Hint                      &  \\
\hline
\end{tabular}
\caption{Selected paraphrase types in LIBERO-Para. Types are derived from Extended Paraphrase Typology (EPT) ~\citep{kovatchev-etal-2018-etpc} 
and Directive Types~\citep{ervin-tripp1976}.}
\label{tab:selected_types}
\end{table}

\begin{table}[t]
\centering
\small
\begin{tabular}{l|l}
\hline
& \\[-1.5ex]
\textbf{\shortstack[l]{Categories}} & \multicolumn{1}{l}{\textbf{\shortstack[l]{Atomic Type}}} \\
\hline
\multirow{3}{*}{Morphology} 
  & Inflectional changes \\
  & Modal verb changes \\
  & Derivational changes \\
\hline
\multirow{5}{*}{Lexicon} 
  & Spelling changes \\
  & Same polarity substitution (habitual) \\
  & Same polarity substitution (contextual) \\
  & Same polarity substitution (named entities) \\
  & Change of format \\
\hline
\multirow{4}{*}{\shortstack[l]{Lexical-\\syntactic}} 
  & Opposite polarity substitution (habitual) \\
  & Opposite polarity substitution (contextual) \\
  & Synthetic/analytic substitution \\
  & Converse substitution \\
\hline
\multirow{5}{*}{Syntax} 
  & Diathesis alternation \\
  & Negation switching \\
  & Ellipsis \\
  & Coordination changes \\
  & Subordination and nesting changes \\
\hline
\multirow{4}{*}{Discourse} 
  & Punctuation changes \\
  & Direct/indirect style alternations \\
  & Sentence modality changes \\
  & Syntax/discourse structure changes \\
\hline
\multirow{3}{*}{Other} 
  & Addition/Deletion \\
  & Change of order \\
  & Semantic based \\
\hline
\multirow{3}{*}{Extremes} 
  & Identity \\
  & Non-paraphrase \\
  & Entailment \\
\hline
\end{tabular}
\caption{Extended Paraphrase Typology (EPT) categories and atomic types~\citep{kovatchev-etal-2018-etpc}.}
\label{tab:ept}
\end{table}

\subsection{Excluded Types from Extended Paraphrase Typology}
\label{app:ept_exclusion}
While EPT provides a broad inventory of paraphrase operations, many types are unsuitable for robotic manipulation instructions under our design constraints. Tab.~\ref{tab:selected_types} lists the types selected for LIBERO-Para, and Tab.~\ref{tab:ept} presents the full Extended Paraphrase Typology (EPT) with its high-level categories and corresponding atomic types. Below we describe, category by category, which types we exclude and why.

\textbf{Morphology.} We exclude the entire Morphology category.
Inflectional changes may alter object cardinality (e.g., pluralization) or modify temporal interpretation when applied to actions.
Modal verb changes can shift intent, introducing semantic drift.
Derivational changes alter part-of-speech (e.g., ``\textit{pick}'' $\rightarrow$ ``\textit{picker}''), which violates imperative structure or disrupts the intended reference.


\textbf{Lexicon.}
We retain same polarity substitution (habitual and contextual) and exclude the remaining types.
Spelling changes are insufficient to constitute meaningful variation.
Same polarity substitution involving named entities is rarely applicable, as robotic instructions predominantly use common nouns and generic verbs.
Change of format is often either trivial or difficult to apply while preserving meaning.

\textbf{Lexical-syntactic.}
We exclude this category in its entirety.
Opposite polarity substitution is unnatural for object nouns and leads to awkward or unintended speech acts when applied to actions.
Synthetic/analytic substitutions (e.g., ``\textit{bowl}'' $\leftrightarrow$ ``\textit{round container}'') are unnatural in concise imperatives.
Converse substitutions introduce role-swapping constructions that are rarely natural in commands.

\textbf{Syntax.}
We retain coordination and subordination changes and exclude the remaining types.
Diathesis alternation yields passive-like commands, which are unnatural in robotic instructions.
Negation switching overlaps with opposite polarity substitutions.
Ellipsis introduces ambiguity in short imperatives and overlaps with addition/deletion.

\textbf{Discourse.}
We exclude this category.
Robotic commands are treated as direct imperatives; alternations in style or sentence modality may alter the intended directive force.
Syntax/discourse structure changes are overly high-level relative to atomic instructions and hinder controlled evaluation.

\textbf{Other.}
We retain only addition.
Given the brevity of imperative commands, deletion frequently removes essential components or produces ungrammatical outputs.
Change of order is often unnatural in short imperatives.
Semantic-based types lack a precise definition and are unsuitable for controlled evaluation.

\textbf{Extremes.} We exclude this category entirely.
Identity involves no transformation.
Non-paraphrase violates meaning preservation.
Entailment represents an inferential relation rather than a meaning-preserving transformation.
\begin{table}[t]
\centering
\footnotesize
\begin{tabular}{l|l}
\hline
\textbf{Directive Type} & \textbf{Example} \\
\hline
Need statements & \textit{``I need a match''} \\
\hline
Imperatives & \textit{``Gimme a match'', ``a match'}' \\
\hline
Embedded imperatives & \textit{``Could you gimme a match?''} \\
\hline
Permission directives & \textit{``May I have a match?''} \\
\hline
Question directives & \textit{``Gotta match?''} \\
\hline
Hints & \textit{``The matches are all gone'}' \\
\hline
\end{tabular}
\caption{Six directive types from Directive Types~\citep{ervin-tripp1976}.}
\label{tab:ervin-tripp}
\end{table}

\subsection{Excluded Type from Directive Types}
\label{app:Benchmark_excluded}

Ervin-Tripp proposed a directive taxonomy that categorizes Directive Types into six types~\citep{ervin-tripp1976}, as summarized in Tab.~\ref{tab:ervin-tripp}. Among these, we select five types for the Action-Pragmatic axis: need statements, embedded imperatives, permission directives, question directives, and hints.

\textbf{Imperatives.} We exclude the imperative type from our paraphrase taxonomy. Since imperatives (e.g.,\textit{ ``Pick up the bowl''}) represent the canonical form of robotic manipulation instructions, they serve as the original instruction rather than a paraphrase variant. In our benchmark design, this type corresponds to the baseline condition (Action axis: None) against which other pragmatic variations are compared.

\begin{table*}[t]
\centering
\resizebox{2\columnwidth}{!}{%
\begin{tabular}{l|c|ccc|cc|ccccc|c}
\hline
& & \multicolumn{3}{c|}{\textbf{Act-Lexical}} & \multicolumn{2}{c|}{\textbf{Act-Structural}} & \multicolumn{5}{c|}{\textbf{Act-Pragmatic}} & \\
\textbf{Object} & \textbf{None} & add & ctx & hab & coord & subord & need & embed & perm & quest & hint & \textbf{Total} \\
\hline
None & -- & 100 & 79 & 74 & 98 & 75 & 93 & 93 & 83 & 87 & 88 & 870 \\
Addition & 98 & 100 & 100 & 100 & 100 & 100 & 100 & 99 & 99 & 99 & 100 & 1,095 \\
Contextual & 87 & 100 & 100 & 100 & 100 & 99 & 100 & 100 & 100 & 94 & 96 & 1,076 \\
Habitual & 74 & 100 & 98 & 100 & 97 & 94 & 100 & 95 & 100 & 95 & 98 & 1,051 \\
\hline
\textbf{Total} & 259 & 400 & 377 & 374 & 395 & 368 & 393 & 387 & 382 & 375 & 382 & \textbf{4,092} \\
\hline
\end{tabular}
}

\vspace{0.5em}
\footnotesize
\textbf{Abbreviations:} add = addition, ctx = same\_polarity\_contextual, hab = same\_polarity\_habitual, coord = coordination, subord = subordination, need = need\_statement, embed = embedded\_imperative, perm = permission\_directive, quest = question\_directive.

\caption{LIBERO-Para dataset statistics. Each cell shows the number of paraphrased instructions for the corresponding Object (row) and Action (column) variation type combination. ``None'' indicates no variation on that axis.}
\label{tab:dataset_statistics}
\end{table*}

\begin{table}[t]
\centering
\footnotesize
\resizebox{\columnwidth}{!}{%
\begin{tabular}{l|c}
\hline
\textbf{Original Instruction} & \textbf{Count} \\
\hline
\textit{Put the wine bottle on top of the cabinet} & 423 \\
\textit{Open the middle drawer of the cabinet} & 416 \\
\textit{Turn on the stove} & 414 \\
\textit{Put the wine bottle on the rack} & 413 \\
\textit{Put the cream cheese in the bowl} & 411 \\
\textit{Open the top drawer and put the bowl inside} & 410 \\
\textit{Put the bowl on top of the cabinet} & 410 \\
\textit{Push the plate to the front of the stove} & 406 \\
\textit{Put the bowl on the stove} & 403 \\
\textit{Put the bowl on the plate} & 386 \\
\hline
\textbf{Total} & \textbf{4,092} \\
\hline
\end{tabular}}
\caption{Number of paraphrased instructions per original instruction in LIBERO-Para.}
\label{tab:instruction_statistics}
\end{table}

\subsection{Paraphrase Dataset Generation}
\label{app:Benchmark_datasetgeneration}



This section describes the paraphrase dataset generation process using LLMs.
As illustrated in Fig.~\ref{fig:work-flow}, our workflow consists of four stages:
(1) axis-wise paraphrase generation,
(2) paraphrase verification,
(3) axis merging, and
(4) final verification. \rev{Each stage is executed by
a role-assigned LLM (generator, merger, or verifier); the full prompts,
including the persona, task specification, and output format for every role,
are provided in
\cref{fig:appen_prompt1,fig:appen_prompt2,fig:appen_prompt3,fig:appen_prompt4,fig:appen_prompt5,fig:appen_prompt6,fig:appen_prompt7,fig:appen_prompt8,fig:appen_prompt9}.}

\textbf{Axis-wise Paraphrase Generation.}
Given an original instruction, a paraphrase generator (LLM) independently produces paraphrases along the action axis (10 types) and the object axis (3 types).
Each generated paraphrase is filtered by a paraphrase verifier (LLM) to ensure meaning preservation and grammatical naturalness.

\textbf{Paraphrase Merging.}
Verified action-axis and object-axis paraphrases modify independent components of the instruction and can therefore be combined.
If $n$ action paraphrases and $m$ object paraphrases pass verification, up to $n \times m$ merged paraphrases are possible.
Merged paraphrases are further validated by the verifier (LLM) before inclusion in the dataset.

\textbf{Design Principles.}
Rather than prompting a single LLM to generate all paraphrase types jointly, we adopt an axis-wise generation and merging strategy.
This modular design assigns a single role to each LLM (generator, merger, and verifier), reducing task complexity and improving generation reliability.
All LLM calls use Gemini 2.5 Pro.
Detailed prompts used at each stage are provided at the end of the paper for readability, and are illustrated in 
\cref{fig:appen_prompt1,fig:appen_prompt2,fig:appen_prompt3,fig:appen_prompt4,fig:appen_prompt5,fig:appen_prompt6,fig:appen_prompt7,fig:appen_prompt8,fig:appen_prompt9}.

\subsection{Statistics of LIBERO-Para}
\label{sec:appendix_statistics}

\begin{figure}[t]
    \centering
    \includegraphics[width=1\linewidth]{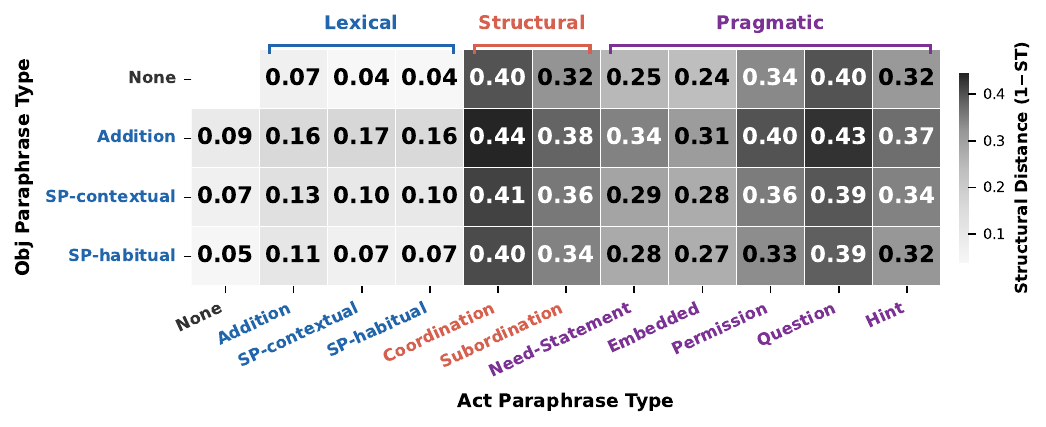}
    \caption{Average Structural Distance ($1 - S_T$) per Object × Action cell (SK weight = 0.0, ST weight = 1.0 in PRIDE). Structural distance is dominated by action paraphrase type rather than object substitution: Coordination and Subordination columns score above 0.28 across all rows, while lexical action types remain below 0.17, confirming that structural rewriting originates from act-level transformations. \rev{Near-zero values are clamped to 0.00 for display.}}
    \label{fig:LIBERO-Para_structural_distance_stat}
\end{figure}
\begin{figure}[t]
    \centering
    \includegraphics[width=1\linewidth]{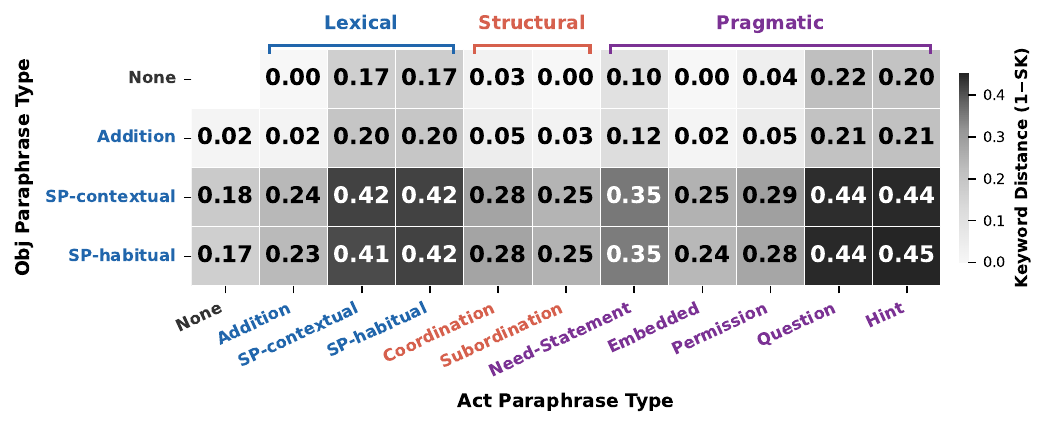}
    \caption{Average Keyword Distance (1 -- $S_K$) per Object × Action cell (SK weight = 1.0, ST weight = 0.0 in PRIDE). Scores are driven primarily by object paraphrasing: SP-contextual and SP-habitual rows score higher regardless of action type, with SP-habitual × Hint reaching 0.45. \rev{Near-zero values are clamped to 0.00 for display.}}  
    \label{fig:LIBERO-Para_keyword_distance_stat}
\end{figure}






LIBERO-Para consists of 4,092 paraphrases generated from 10 original LIBERO-Goal instructions, selected from the four LIBERO task types (Spatial, Object, Goal, and Long) where linguistic understanding is essential for successful execution.

The dataset is organized along two axes:
an Object axis with three lexical types, and an Action axis with ten types (three lexical, two structural, and five pragmatic).
This two-axis design yields 43 distinct paraphrase type combinations: three Object-only, ten Action-only, and thirty compositional types ($3 \times 10$).

Tab.~\ref{tab:dataset_statistics} reports the number of paraphrases for each Object $\times$ Action combination.
The dataset includes 259 Object-only paraphrases (Action = None), 870 Action-only paraphrases (Object = None), and 2,963 compositional paraphrases.
To facilitate diverse analyses, samples are distributed approximately uniformly across cells, with around 100 paraphrases per cell.

To examine how each component of PRIDE contributes to paraphrase difficulty, Figs.~\ref{fig:LIBERO-Para_structural_distance_stat} and~\ref{fig:LIBERO-Para_keyword_distance_stat} present the average structural distance ($1 - S_T$) and keyword distance ($1 - S_K$), respectively, which correspond to isolating each term in the PD formulation (Eq.~\ref{eq:3}).
Structural distance is dominated by the Action axis: coordination and subordination columns consistently score above 0.28 regardless of object type, whereas lexical action types stay below 0.17.
Keyword distance, in contrast, is primarily driven by the Object axis: contextual and habitual substitutions yield high distances (0.41--0.45) due to synonym replacement, while rows without object paraphrasing remain near zero.
This decomposition confirms that the two PRIDE components capture complementary sources of difficulty---syntactic divergence from action paraphrasing and lexical divergence from object paraphrasing.

Finally, Tab.~\ref{tab:instruction_statistics} reports the number of paraphrases per original instruction.
Each instruction yields 386--423 paraphrases, indicating a balanced distribution.

\subsection{Human Evaluation}
\label{appen:Benchmark_HumanEvaluation}

To verify the semantic validity of LIBERO-Para, we conducted a human evaluation on a randomly sampled 5\% subset (205 samples) of the full benchmark.
Fifteen annotators independently judged whether each original--paraphrase pair would elicit the same successful behavior in the given scene, using a binary Yes/No decision.
 
\paragraph{Inter-Annotator Agreement.}
We report Gwet's AC1~\citep{gwet2008computing} as the inter-annotator agreement (IAA) metric.
We chose AC1 over Cohen's or Fleiss' $\kappa$ because our labels are heavily skewed toward the positive class, a setting in which $\kappa$ is known to be substantially deflated despite high observed agreement~\citep{feinstein1990high}.
On our 15-annotator evaluation, Gwet's AC1 is \textbf{0.854}, indicating strong agreement.
 
\paragraph{Consensus Statistics.}
Under a majority-vote criterion ($\geq$8/15 annotators marking Yes), 204 out of 205 samples (99.51\%) were judged as meaning-preserving.
Under a stricter threshold requiring $\geq$12/15 agreement (80\%), 183 out of 205 samples (89.27\%) passed.
Across all 205 samples, annotators selected Yes at an average rate of 14.13/15 (94.18\%), further supporting high item-level consensus.
 
\paragraph{Error Analysis.}
We examined the 22 samples that failed the stricter criterion and found that disagreement was concentrated in paraphrases where the original imperative form was transformed into suggestive, declarative, or indirect speech-act forms.
This indicates that disagreement primarily arose from differences in annotator interpretation of speech-act form rather than semantic distortion of the paraphrase itself. At the same time, such cases confirm that the benchmark includes pragmatically challenging linguistic reformulations that go beyond simple lexical substitution.
\paragraph{Annotation Protocol.}
\begin{figure}[t]
    \centering
    \includegraphics[width=1\linewidth]{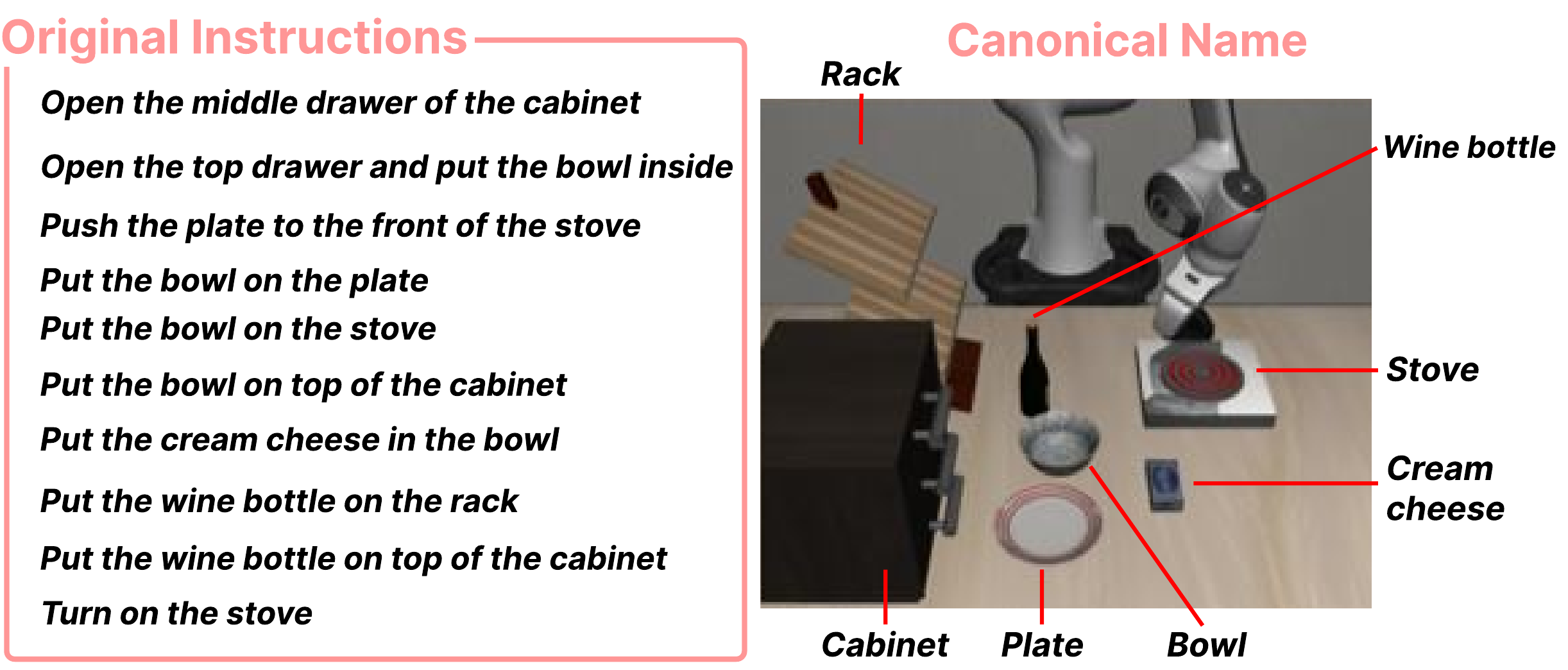}
    \caption{LIBERO-Goal task instructions (left) and corresponding scene with canonical object names (right). Each object is referred to by a single unique keyword throughout all instructions (e.g., \textit{stove}, \textit{bowl}, \textit{rack}), with no lexical variation across tasks.}
    \label{fig:canonical_instructions}
\end{figure}
\begin{table}[!h]
\centering
\small
\resizebox{\columnwidth}{!}{%
\begin{tabular}{lcccc}
\hline
TED Norm. & Pearson & Spearman & Range & $[0,1]$ Bounded \\
\hline
\textbf{Sum (Current)} & \textbf{-0.802} & \textbf{-0.831} &
\textbf{$[0,1]$} & \textbf{Inherently} \\
Max (raw) & -0.715 & -0.747 & $[-1,1]$ & No \\
Max (clipped) & -0.715 & -0.747 & $[0,1]$ & Post-hoc \\
Average & -0.650 & -0.676 & $[-1,1]$ & No \\
\hline
\end{tabular}%
}
\caption{\rev{Ablation of TED normalization choices for $S_T$. Pearson and
Spearman correlations are computed between PRIDE and success rate across
43 paraphrase cells and seven VLA configurations. Max (raw) produces
9 negative values among 4,092 paraphrases, while Max (clipped) applies
post-hoc clipping to $[0,1]$.}}
\label{tab:ted_norm}
\end{table}
Each annotator received an Excel spreadsheet containing 205 randomly sampled original--paraphrase pairs.
They were instructed to mark \texttt{O} if the paraphrased instruction would elicit the same successful behavior as the original instruction in the given VLA scene (LIBERO-Goal initial scene), and \texttt{X} otherwise.
The 15 annotators included participants with varying levels of 
familiarity with robotic manipulation tasks, ranging from 
domain-familiar researchers to non-expert volunteers.
All annotators were informed that their responses would be used for research purposes.

\section{PRIDE: Paraphrase Robustness Index in Robotic Instructional DEviation}
\label{sec:appen_metric}

\begin{table*}[t]
\centering
\small
\resizebox{\textwidth}{!}{
\begin{tabular}{llllccccccc}
\toprule
Original & Paraphrase & Type & SR(\%) & PRIDE & $1-S_K$ & $1-S_T$ & 1-BERT & 1-BLEU & 1-METEOR \\
\midrule
\multirow{3}{*}{\shortstack[l]{Put the cream\\cheese in the bowl}}
 & ``carefully put the cream cheese in the bowl'' & addition(act) & 90.8 & 0.03 & 0.00 & 0.07 & 0.14 & 0.16 & 0.02 \\
 & ``put the cheese spread in the vessel'' & SP-contextual(obj) & 70.3 & 0.35 & 0.27 & 0.43 & 0.12 & 0.80 & 0.36 \\
 & ``Is the spread supposed to go in the container?'' & SP-contextual(obj) \& question & 31.5 & 0.56 & 0.60 & 0.53 & 0.28 & 0.91 & 0.65 \\
\midrule
\multirow{3}{*}{\shortstack[l]{Turn on the stove}}
 & ``carefully turn on the stove'' & addition(act) & 90.8 & 0.06 & 0.00 & 0.11 & 0.15 & 0.33 & 0.03 \\
 & ``turn on the range'' & SP-habitual(obj) & 71.8 & 0.36 & 0.34 & 0.38 & 0.19 & 0.41 & 0.26 \\
 & ``Is the range hot yet?'' & SP-habitual(obj) \& hint & 33.2 & 0.65 & 0.70 & 0.60 & 0.40 & 0.92 & 0.88 \\
\midrule
\multirow{3}{*}{\shortstack[l]{Open the middle\\drawer of the cabinet}}
 & ``carefully open the middle drawer of the cabinet'' & addition(act) & 90.8 & 0.03 & 0.00 & 0.05 & 0.10 & 0.12 & 0.01 \\
 & ``Find the cabinet, then proceed to open the middle drawer'' & coordination & 80.2 & 0.25 & 0.00 & 0.50 & 0.10 & 0.12 & 0.01 \\
 & ``The storage unit's middle compartment is currently shut'' & SP-habitual(obj) \& hint & 31.6 & 0.46 & 0.41 & 0.50 & 0.19 & 0.72 & 0.49 \\
\bottomrule
\end{tabular}
}
\caption{Qualitative comparison of PRIDE, a task-grounded paraphrase distance metric, with general-purpose NLP distance metrics on selected LIBERO-Para examples. Each task group presents three paraphrases of increasing linguistic distance: a minor addition, a lexical substitution, and a compound paraphrase combining object substitution with an indirect speech act. PRIDE increases monotonically as success rate (SR) degrades, reflecting its decomposition into keyword similarity ($S_K$) and structural similarity ($S_T$). In contrast, 1$-$BERT lacks discriminative range, 1$-$BLEU fluctuates inconsistently, and 1$-$METEOR fails to capture structurally induced difficulty when keywords are preserved (e.g., coordination in the third group scores 0.01 despite a 10.6 pp SR drop).}
\label{tab:pride_comparison}
\end{table*}
\paragraph{Motivation.}
General-purpose NLP distance metrics such as BERTScore~\citep{bertscore}, 
BLEU~\citep{bleu}, and METEOR~\citep{meteor} are designed to measure 
surface-level or semantic similarity between text pairs, without 
considering how linguistic changes affect downstream task execution.
In grounded robotic instruction following, however, not all lexical 
changes are equally disruptive: replacing a task-critical object noun 
(e.g., ``stove'' $\to$ ``range'') directly impacts visual grounding and 
action selection, whereas syntactic additions (e.g., prepending 
``carefully'') leave the core command intact.
PRIDE is designed to reflect this asymmetry by decomposing paraphrase 
distance into two robot-relevant axes: keyword divergence ($S_K$), which 
captures whether task-critical referents are preserved, and structural 
divergence ($S_T$), which measures how far the utterance departs from the 
imperative form that VLA models are predominantly trained on.
\paragraph{Dependency-Head-Aware Extension of $S_K$.}

The default $S_K$ measures action/object-centered lexical preservation using
content-word matching. While this formulation is suitable for the short and
structured instructions considered in our benchmark, it does not explicitly
distinguish task-critical action/object heads from peripheral modifiers.

As an extension, dependency parsing can be used to isolate action roots,
direct objects, and nominal modifiers, enabling head-only or head-weighted
variants of $S_K$. For example, given \textit{``Put the bowl on the stove''}
and \textit{``Place the container on the cooktop,''} a head-aware variant
can explicitly compare \textit{Put} $\leftrightarrow$ \textit{Place} as
action heads and \textit{bowl} $\leftrightarrow$ \textit{container} as
object heads, while distinguishing such task-critical components from
peripheral modifiers. This provides a more targeted view of lexical
preservation when dependency heads can be reliably identified. We retain the content-word-based $S_K$ in Eq.~\eqref{eq:1} as the default formulation;
the dependency-head-aware formulation is provided as an extension and is not
used to recompute the results reported in this work.

\paragraph{\rev{Parser Stability.}}
\rev{We use the spaCy \texttt{en\_\allowbreak core\_\allowbreak web\_\allowbreak sm}
parser to obtain the dependency trees used in $S_T$. To assess parser stability,
we manually verified the dependency assignments for a random sample of 150
paraphrases from LIBERO-Para. Of these, 146 out of 150 parses were judged
correct, corresponding to an accuracy of 97.3\%. This validation supports the
stability of the dependency parsing used to compute $S_T$ across the
paraphrases in our benchmark.}

\paragraph{\rev{TED Normalization Ablation.}}
\rev{The normalization in Eq.~\eqref{eq:2} is designed to keep $S_T$
within the $[0,1]$ range. For two dependency trees $T_O$ and $T_P$,
TED is upper-bounded by $|T_O|+|T_P|$; therefore, the sum normalization
used in Eq.~\eqref{eq:2} inherently guarantees bounded structural
similarity. We additionally compare this choice against max, clipped-max,
and average tree-size normalization, as shown in Table~\ref{tab:ted_norm}.}

\rev{The sum normalization yields the strongest PRIDE--SR correlation
while inherently preserving the $[0,1]$ range. Max and average
normalizations divide the same TED value by smaller denominators, which
increases the structural penalty and can over-amplify minor syntactic
variation. Accordingly, we retain the sum normalization in
Eq.~\eqref{eq:2}.}

\paragraph{Qualitative Comparison with NLP Metrics.}
Tab.~\ref{tab:pride_comparison} illustrates how PRIDE captures task-relevant 
linguistic variation compared to general-purpose NLP metrics.
For each task group, we present three paraphrases of increasing difficulty: 
a minor addition (e.g., prepending ``carefully''), a lexical substitution 
of the object or action, and a compound paraphrase combining object 
substitution with an indirect speech act.
PRIDE increases monotonically as success rate degrades---for instance, 
in the first group, PRIDE rises from 0.03 to 0.35 to 0.56 as SR drops 
from 90.8\% to 70.3\% to 31.5\%.
This graduated behavior stems from the complementary design of its two 
components: $S_K$ remains near zero for syntactic-only changes 
(e.g., addition) but sharply increases when task-critical keywords are 
replaced, while $S_T$ captures structural divergence from the imperative 
form even when keywords are preserved.

In contrast, conventional NLP metrics each exhibit notable limitations.
1$-$BERTScore~\citep{bertscore} remains in a narrow range (0.10--0.28) 
across all paraphrase types, failing to distinguish between benign 
additions and highly disruptive compound paraphrases.
1$-$BLEU~\citep{bleu} behaves erratically: in the first group, it assigns 
a higher distance to a simple object substitution (0.80) than the gap 
between that substitution and a far more disruptive compound form 
(0.80 $\to$ 0.91), compressing meaningful difficulty differences.
1$-$METEOR~\citep{meteor} tracks the overall degradation trend more 
faithfully than the other two metrics, owing to its synonym and stem 
matching via WordNet~\citep{miller1995wordnet}.
However, it still fails to capture structurally induced difficulty: 
in the third group, coordination (``Find the cabinet, then proceed to 
open the middle drawer'') receives the same score as a trivial addition 
(both 0.01), despite a 10.6 pp SR gap (90.8\% $\to$ 80.2\%), because 
the original keywords are largely preserved.
More fundamentally, METEOR provides only a single scalar distance 
and cannot decompose \textit{why} a paraphrase is distant---whether 
due to keyword replacement or structural transformation---limiting 
its diagnostic utility.
PRIDE addresses this through its explicit $S_K$/$S_T$ decomposition, 
enabling researchers to attribute performance degradation to specific 
linguistic dimensions.

\begin{figure*}[t]
    \centering
    \includegraphics[width=1\linewidth]{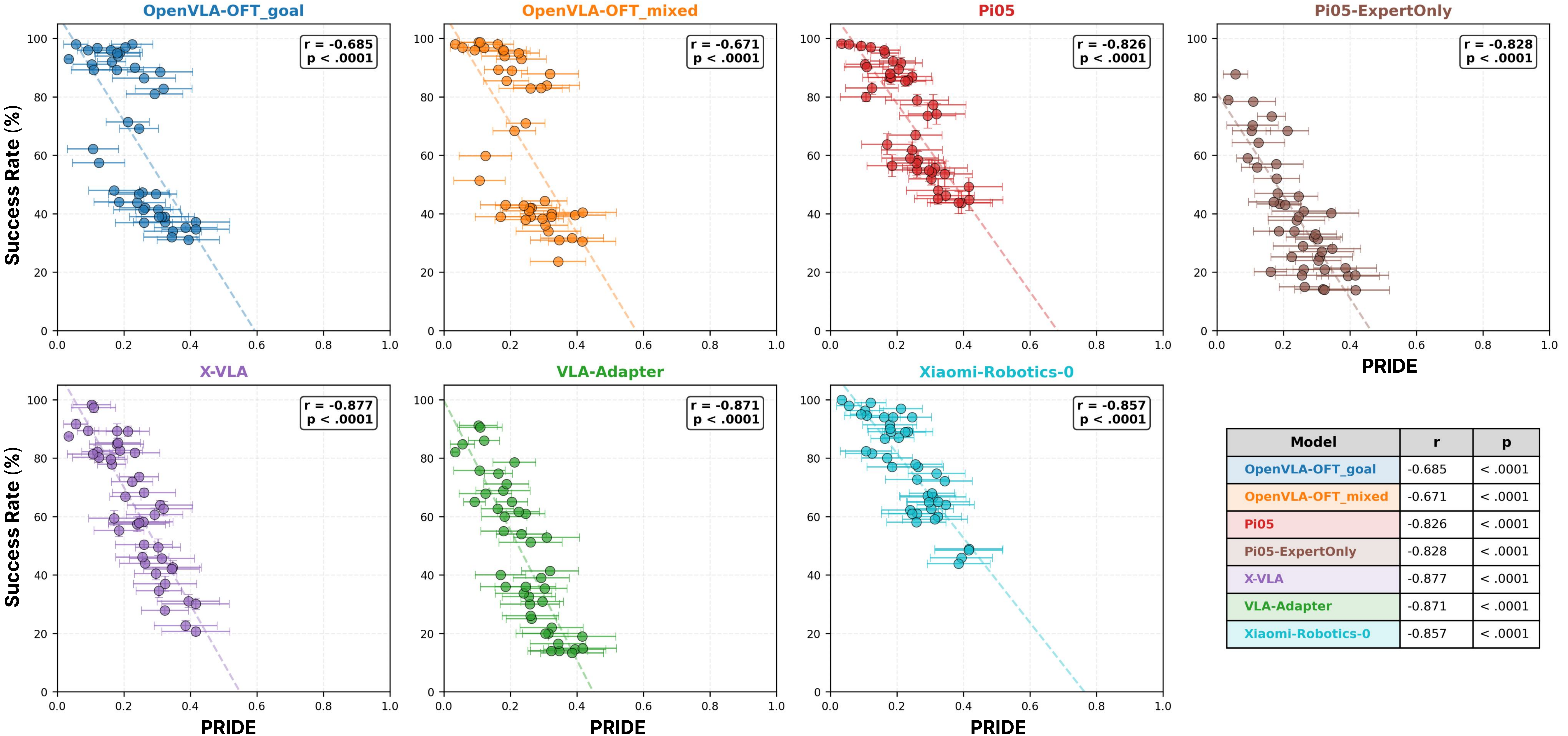}
    \caption{Correlation between PRIDE score (PD) and success rate (SR) for each VLA model on LIBERO-Para. Each point represents the mean SR of a paraphrase cell, with error bars indicating standard deviation. Colors are unified per model for visual clarity. All models exhibit statistically significant negative correlations ($p < .0001$), with Pearson $r$ values ranging from $-0.671$ to $-0.877$, validating that higher paraphrase distance consistently leads to lower task success. The summary table (bottom right) reports $r$ and $p$ for all models.}
    \label{fig:good_correlation_my_bro}
\end{figure*}

\begin{table}[t]
\centering
\rev{
\small
\begin{tabular*}{\columnwidth}{@{\extracolsep{\fill}}lccc@{}}
\toprule
Metric & $r$ & $\rho$ & Ranked 1st (of 7) \\
\midrule
\textbf{PRIDE (ours)} & \textbf{-0.802} & \textbf{-0.831} & \textbf{6} \\
1\,--\,METEOR    & -0.777 & -0.785 & 1\rlap{$^{\dagger}$} \\
1\,--\,BERTScore & -0.669 & -0.689 & 0 \\
1\,--\,BLEU      & -0.642 & -0.716 & 0 \\
\bottomrule
\end{tabular*}
}
\caption{\rev{Correlation with success rate over 43 cells $\times$ 7 VLAs
(5 seeds per VLA). $r$ denotes Pearson's and $\rho$ denotes Spearman's
correlation. Ranked 1st: number of VLAs on which the metric achieves the
strongest correlation. All values are significant ($p<0.0001$).
$^{\dagger}$1\,--\,METEOR ranks first only on
$\pi$\textsubscript{0.5} (expert-only), the sole model with base-task SR
below 90\%, where base-task noise dominates.}}
\label{tab:pride_vs_other}
\end{table}
\begin{table}[t]
\centering
\rev{
\small
\begin{tabular*}{\columnwidth}{@{\extracolsep{\fill}}lcccc@{}}
\toprule
\multirow{2}{*}{Cell group ($n$)} & \multicolumn{2}{c}{PRIDE} & \multicolumn{2}{c}{1\,--\,METEOR} \\
\cmidrule(lr){2-3} \cmidrule(l){4-5}
 & $r$ & $\rho$ & $r$ & $\rho$ \\
\midrule
Action-only (10)   & \textbf{-0.747} & \textbf{-0.733} & -0.719 & -0.711 \\
Object-only (3)$^{\ddagger}$ & \textbf{-0.969} & -0.857 & -0.920 & -0.857 \\
Compositional (30) & \textbf{-0.756} & \textbf{-0.799} & -0.678 & -0.676 \\
\bottomrule
\end{tabular*}
}
\caption{\rev{Per-cell-group correlation with success rate (mean across 7
VLAs, 5 seeds per VLA), following the cell partition of
Tab.~\ref{tab:super_additive}. All values are significant ($p<0.0001$).
$^{\ddagger}$For Object-only ($n=3$), the small sample limits Spearman
resolution: both metrics tie at -0.857, while Pearson remains
discriminative.}}
\label{tab:pride_vs_meteor}
\end{table}
\begin{table}[t]
\centering
\rev{
\resizebox{\columnwidth}{!}{
\begin{tabular}{lccc}
\toprule
OOD measure (category) & PD & $S_K$ & $S_T$ \\
\midrule
E5 cosine distance (semantic)          & 0.844 & 0.784 & 0.378\rlap{$^{*}$} \\
Content-lemma novelty (lexical)        & 0.789 & 0.882 & 0.189\rlap{$^{\wedge}$} \\
Dependency-triple Jaccard (lexical)    & 0.886 & 0.841 & 0.377\rlap{$^{*}$} \\
POS-bigram cosine distance (syntactic) & 0.628 & 0.001\rlap{$^{\wedge}$} & 0.921 \\
\bottomrule
\end{tabular}
}
}
\caption{\rev{Absolute Pearson correlation $|r|$ between PRIDE components and
four independent OOD-ness measures at the 43-cell level (same setting as
Fig.~\ref{fig:good_correlation_my_bro}). Unmarked: $p<0.0001$; $^{*}$: $0.0001 \le p < 0.05$;
$^{\wedge}$: not significant ($p \ge 0.05$). $S_K$ aligns with
lexical measures but not the syntactic one; $S_T$ shows the
opposite pattern.}}
\label{tab:double_dissociation}
\end{table}
\paragraph{Quantitative Validation.}
Beyond qualitative examples, we verify that PRIDE 
correlates with actual task performance. 
Fig.~\ref{fig:good_correlation_my_bro} plots the 
mean success rate of each paraphrase cell against 
its PRIDE score for all seven models. All models 
exhibit statistically significant negative correlations 
(Pearson $r$ ranging from $-0.671$ to $-0.877$, 
$p < .0001$), confirming that higher paraphrase 
distance consistently leads to lower task success. 
This validates PRIDE as a meaningful difficulty 
metric for paraphrase robustness evaluation.

\rev{Moreover, PRIDE outperforms general-purpose NLP metrics in predicting
success rate. As shown in Tab.~\ref{tab:pride_vs_other}, PRIDE achieves the
strongest correlation ($r=-0.802$, $\rho=-0.831$), outperforming
1\,--\,METEOR, 1\,--\,BERTScore, and 1\,--\,BLEU, and ranks first on 6 of 7
VLAs. The sole exception is $\pi$\textsubscript{0.5} (expert-only), the only
model with base-task SR below 90\% (78.6\% vs.\ 96--99\% for the others),
a degenerate case where base-task noise dominates the signal. Per-seed
analysis confirms that this advantage is deterministic at the model level:
PRIDE ranks first in all 30 conditions (6 VLAs $\times$ 5 seeds, excluding
the degenerate case). Tab.~\ref{tab:pride_vs_meteor} further decomposes this
advantage by cell group against the strongest baseline, 1\,--\,METEOR. PRIDE
leads in every group, and its margin is largest in the Compositional setting:
simultaneous action and object substitution saturates METEOR's unigram
matching, while the Sentence-BERT-based $S_K$ and dependency-based $S_T$
continue to resolve both lexical and structural variation.}

\paragraph{\rev{External Validation of $S_K$ and $S_T$.}}
\rev{To verify that the two components capture disentangled linguistic
dimensions, we correlate PD, $S_K$, and $S_T$ with four independent OOD-ness
measures computed by formulations distinct from PRIDE: E5 cosine distance
(semantic)~\citep{e5}, content-lemma novelty and dependency-triple Jaccard
(lexical), and POS-bigram cosine distance (syntactic). As shown in
Tab.~\ref{tab:double_dissociation}, $S_K$ aligns strongly with lexical
measures ($|r|$ up to 0.882) while remaining uncorrelated with the syntactic
one ($|r|=0.189$, n.s.), and $S_T$ shows the exact opposite pattern
($|r|=0.921$ syntactic vs.\ $|r|=0.001$ lexical, n.s.). This double
dissociation demonstrates that $S_K$ and $S_T$ capture complementary,
non-overlapping dimensions. Notably, $S_K$'s higher correlation with
content-lemma novelty than PD is consistent with its design intent:
Sentence-BERT embeddings are sensitive to lexical substitutions, aligning
$S_K$ with the lexical-content axis. This disentanglement enables the
axis-specific diagnosis of VLA failures that single-score metrics cannot
provide.}









\section{Experiment}
\subsection{Setup}
\label{appen:Experiment_Setup}

\begin{figure*}[t]
    \centering
    \includegraphics[width=1\linewidth]{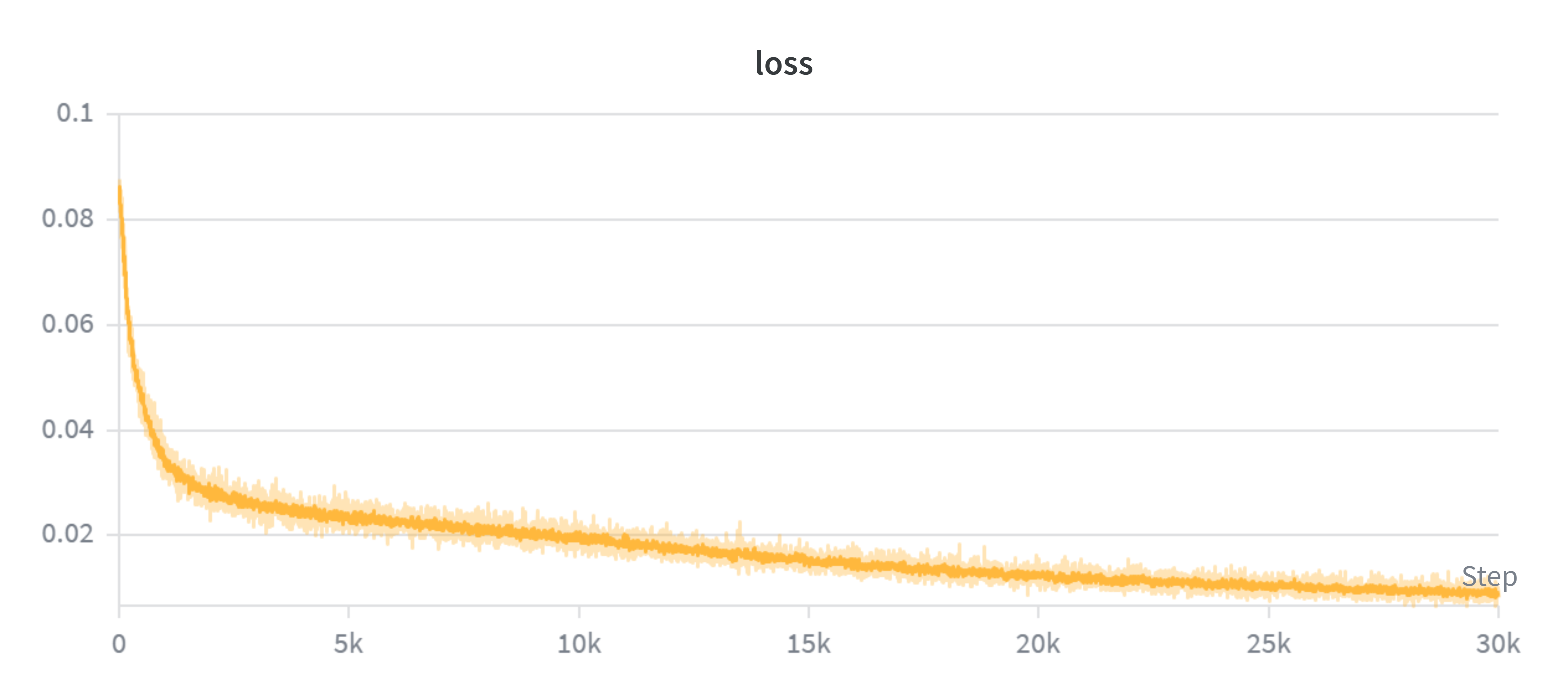}
    \caption{Training loss curve of $\pi_{0.5}$ (expert-only) fine-tuned on LIBERO. The model is trained for 30K steps, matching the original training configuration. The loss converges around 15K steps, indicating stable training completion.}
    \label{fig:pi05_nocap}
\end{figure*}

\paragraph{Computing Infrastructure.}
All experiments were conducted on NVIDIA RTX A6000 and NVIDIA L40S GPUs.
Specifically, OpenVLA-OFT variants were evaluated on RTX A6000 GPUs, while all other models (X-VLA, VLA-Adapter, $\pi_{0.5}$, $\pi_{0.5}$ (expert-only), and Xiaomi-Robotics-0) were evaluated on L40S GPUs.
The total evaluation cost across all seven model configurations amounts to approximately \textbf{194 GPU hours} (Tab.~\ref{tab:gpu_hours}).
For $\pi_{0.5}$ (expert-only), the training was also performed on L40S GPUs following the original $\pi_{0.5}$ fine-tuning protocol (see Tab.~\ref{tab:appen_pi05_expert_only_finetune} for training hyperparameters and Fig.~\ref{fig:pi05_nocap} for the training loss curve).

\begin{table}[h]
\centering
\small
\resizebox{\columnwidth}{!}{%
\begin{tabular}{lccc}
\toprule
Model & GPU & VRAM (GB) & Eval Hours \\
\midrule
OpenVLA-OFT\textsubscript{goal} & A6000 & $\sim$16 & $\sim$12 \\
OpenVLA-OFT\textsubscript{mixed} & A6000 & $\sim$16 & $\sim$12 \\
X-VLA & L40S & $\sim$6.5 & $\sim$11 \\
VLA-Adapter & L40S & $\sim$3 & $\sim$11 \\
$\pi$\textsubscript{0.5} & L40S & $\sim$38 & $\sim$70 \\
$\pi$\textsubscript{0.5} (expert-only) & L40S & $\sim$38 & $\sim$70 \\
Xiaomi-Robotics-0 & L40S & $\sim$14 & $\sim$8 \\
\midrule
\textbf{Total} & & & $\sim$\textbf{194} \\
\bottomrule
\end{tabular}%
}
\caption{Evaluation GPU hours and peak VRAM usage per model configuration.}
\label{tab:gpu_hours}
\end{table}

\begin{figure}[t]
\centering
\includegraphics[width=\columnwidth]{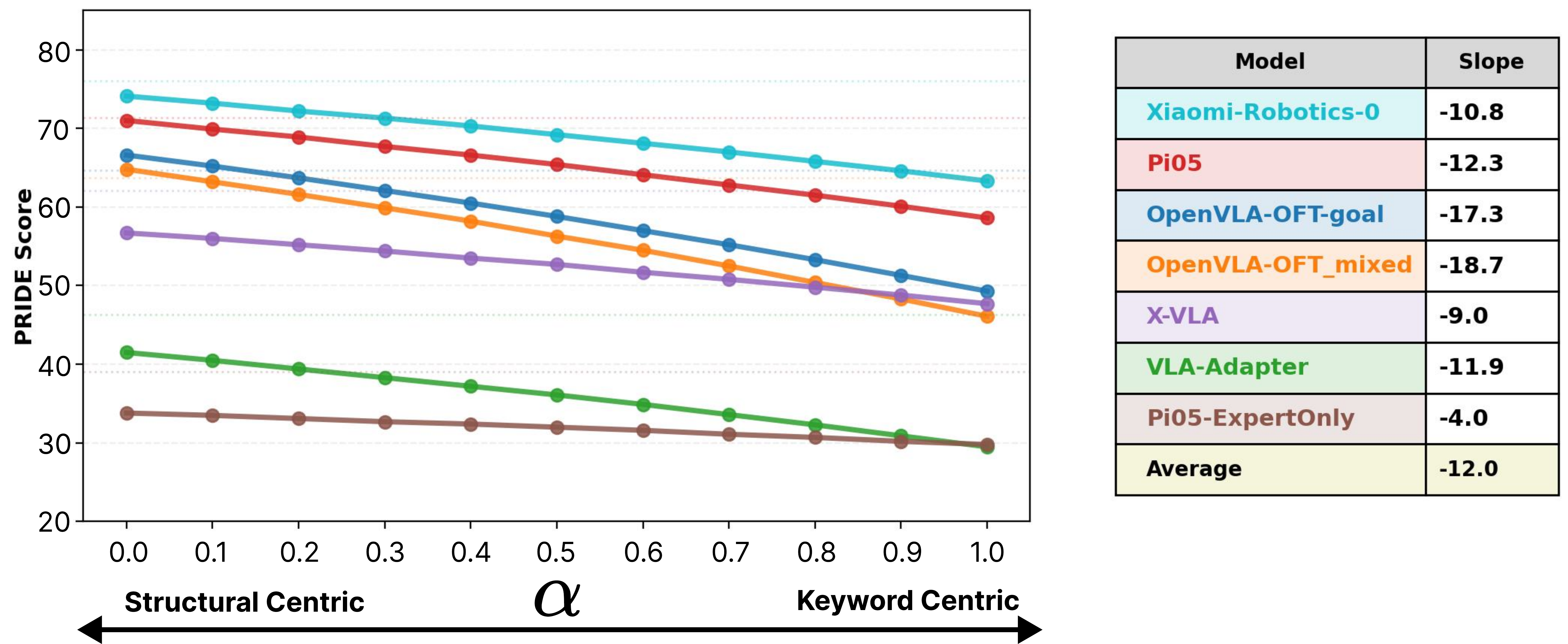}
\caption{Effect of the weighting parameter $\alpha$ on PRIDE scores across all models. 
Left: as $\alpha$ increases from 0 (structure-centric) to 1 (keyword-centric), 
PRIDE scores decrease consistently for all models, indicating that keyword-based 
evaluation assigns higher credit to samples that models already solve easily. 
Right: per-model linear slope of the PRIDE--$\alpha$ curve. 
Steeper negative slopes indicate stronger dependence on keyword similarity 
over structural similarity.}
\label{fig:alpha_sweep}
\end{figure}

\begin{figure}[t]
\centering
\includegraphics[width=\columnwidth]{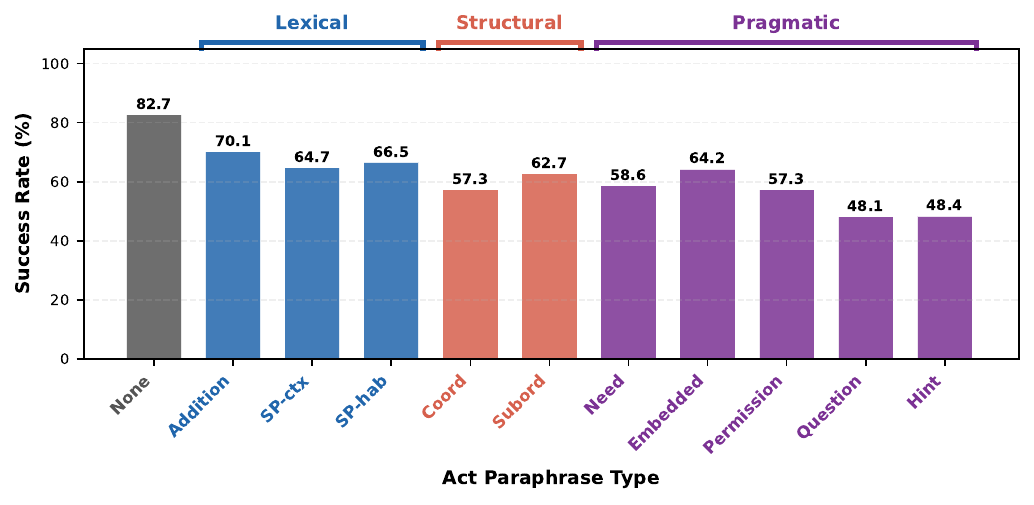}
\caption{Success rate breakdown by action paraphrase type, averaged across 
all 7 model configurations. Paraphrase types are grouped into three 
linguistic categories: \textcolor[HTML]{4A72B8}{Lexical} (surface-level 
word changes), \textcolor[HTML]{C0504D}{Structural} (syntactic 
reorganization), and \textcolor[HTML]{7030A0}{Pragmatic} (indirect 
speech acts). Performance degrades progressively from the original 
instruction (82.7\%) through lexical variants (66--70\%) and structural 
variants (57--63\%) to the most indirect pragmatic forms such as 
Question (48.1\%) and Hint (48.4\%).}
\label{fig:action_axis}
\end{figure}
\begin{table*}[t]
\centering
\small
\resizebox{\textwidth}{!}{%
\begin{tabular}{lll}
\toprule
Model & Code & Weights \\
\midrule
OpenVLA-OFT\textsubscript{goal} & \url{https://github.com/moojink/openvla-oft} & \url{https://huggingface.co/moojink/openvla-7b-oft-finetuned-libero-goal} \\
OpenVLA-OFT\textsubscript{mixed} & (same as above) & \url{https://huggingface.co/moojink/openvla-7b-oft-finetuned-libero-spatial-object-goal-10} \\
$\pi$\textsubscript{0.5} & \url{https://github.com/Physical-Intelligence/openpi} (JAX) & \texttt{gs://openpi-assets/checkpoints/pi05\_libero} \\
$\pi$\textsubscript{0.5} (expert-only) & (same as $\pi_{0.5}$) & Fine-tuned from \texttt{gs://openpi-assets/checkpoints/pi05\_base} \\
X-VLA & \url{https://github.com/huggingface/lerobot} & \url{https://huggingface.co/lerobot/xvla-libero} \\
VLA-Adapter & \url{https://github.com/OpenHelix-Team/VLA-Adapter} & \url{https://huggingface.co/VLA-Adapter/LIBERO-Goal-Pro} \\
Xiaomi-Robotics-0 & \url{https://github.com/XiaomiRobotics/Xiaomi-Robotics-0} & \url{https://huggingface.co/XiaomiRobotics/Xiaomi-Robotics-0-LIBERO} \\
\bottomrule
\end{tabular}%
}
\caption{Code repositories and pretrained weight sources for all evaluated models.}
\label{tab:model_links}
\end{table*}
\begin{table*}[t]
\centering
\resizebox{\textwidth}{!}{
\begin{tabular}{llllclcc}
\toprule
Model & Release & Arch.\ Type & VLM Backbone & VLM Params & Action Module & Action Params & Total Params \\
\midrule
OpenVLA-OFT & 2025.03 & Parallel Decoding & Prismatic (Llama 2) & 7B & L1 MLP & {<}1M & 7.5B \\
$\pi$\textsubscript{0.5} & 2025.09 & VLM + Action Expert & PaliGemma 2 & 3B & Flow matching expert & 0.3B & 3.3B \\
VLA-Adapter & 2025.09 & Bridge-based & Prismatic (Qwen2.5-0.5B) & 0.5B & Bridge Attention Policy & 97M & 0.6B \\
X-VLA & 2026.01 & Soft-prompted & Florence-2 & 0.5B & Flow matching transformer & {$\sim$}0.4B & 0.9B \\
Xiaomi-Robotics-0 & 2026.02 & VLM + Action Expert & Qwen3-VL-4B & 4B & Flow matching DiT & {$\sim$}0.7B & 4.7B \\
\bottomrule
\end{tabular}
}
\caption{Architecture-level specifications of the evaluated VLA models. Release denotes the public code release date (YYYY.MM). Models span a range of architectural paradigms---from parallel decoding to bridge-based adapters to flow matching action experts---with total parameter counts ranging from 0.6B to 7.5B. OpenVLA-OFT variants (goal/mixed) share the same architecture and are listed as a single entry.}
\label{tab:appen_model_spec}
\end{table*}
\paragraph{Model Weights and Code.}
The code repositories and pretrained weights are listed in Tab.~\ref{tab:model_links}.
All evaluated models use publicly released checkpoints and official codebases, except for $\pi_{0.5}$ (expert-only), which we fine-tuned from the base $\pi_{0.5}$ checkpoint by freezing the VLM and updating only the action expert.
Tab.~\ref{tab:appen_model_spec} summarizes architecture-level specifications, and Tab.~\ref{tab:model_LIBEROfinetune_spec} details the fine-tuning configurations.
\paragraph{Backbone and Data References.}
The VLM backbones used across evaluated models include 
Prismatic~\citep{karamcheti2024prismatic} with Llama~2~\citep{touvron2023llama2}, 
PaliGemma~2~\citep{paligemma2}, 
Florence-2~\citep{xiao2024florence2}, 
Qwen2.5~\citep{qwen25}, 
and Qwen3-VL~\citep{qwen3vl}. 
For pre-training data, OpenVLA-OFT uses the Open X-Embodiment (OXE) dataset~\citep{oxe}, 
and X-VLA is pre-trained on Droid~\citep{droid}, RoboMind~\citep{robomind}, 
and Agibot~\citep{agibot}. 
All $\pi_{0.5}$ variants use AdamW~\citep{adamw} as the optimizer (Tab.~\ref{tab:appen_pi05_expert_only_finetune}).

\begin{table*}[t]
\centering
\small
\resizebox{\textwidth}{!}{%
\begin{tabular}{lllllc}
\toprule
\multirow{2}{*}{Model} & Pre-train & \multirow{2}{*}{LIBERO Data} & \multicolumn{2}{c}{Fine-tune (Domain Adaptation)} & \multirow{2}{*}{Weights} \\
\cmidrule(l){4-5}
 & Pre-train Data & & FT Method & FT Scope & \\
\midrule
OpenVLA-OFT\textsubscript{goal}  & OXE (970k traj) & Goal only & LoRA (r=32) & All modules & Released \\
OpenVLA-OFT\textsubscript{mixed} & OXE (970k traj) & All 4 suites & LoRA (r=32) & All modules & Released \\
$\pi$\textsubscript{0.5}         & Proprietary + open (10K+ hrs) & All 4 suites & Full & All modules & Released \\
$\pi$\textsubscript{0.5} (expert-only) & Proprietary + open (10K+ hrs) & All 4 suites & Full & Action Expert only & Ours \\
VLA-Adapter                       & No robotic pretrain & Goal only & LoRA (r=64) & All modules & Released \\
X-VLA                             & Droid + Robomind + Agibot (290k) & All 4 suites & LoRA (r=64) & All modules & Released \\
Xiaomi-Robotics-0                 & Open + in-house ($\sim$200M steps) & All 4 suites & Full & All modules & Released \\
\bottomrule
\end{tabular}%
}
\caption{LIBERO fine-tuning configurations of the evaluated VLA models. All models are fine-tuned on LIBERO and evaluated on LIBERO-Para. ``Released'' denotes publicly available checkpoints; ``Ours'' denotes checkpoints fine-tuned by us following the original training protocol (see Fig.~\ref{fig:pi05_nocap} for the training loss curve). Models trained on ``All 4 suites'' use the mixed configuration of LIBERO-Goal, LIBERO-Spatial, LIBERO-Object, and LIBERO-Long. Note that $\pi_{0.5}$ (expert-only) freezes the vision-language backbone and updates only the action expert.}
\label{tab:model_LIBEROfinetune_spec}
\end{table*}
\begin{table*}[t]
\centering
\small
\begin{tabular}{lcc}
\toprule
 & $\pi$\textsubscript{0.5} & $\pi$\textsubscript{0.5} (expert-only) \\
\midrule
VLM (img + llm) & Fine-tuned & Frozen \\
Action Expert & Fine-tuned & Fine-tuned \\
Trainable Params & $\sim$3.3B & $\sim$300M \\
Batch Size & 256 & 256 \\
Peak LR & 5e-5 & 5e-5 \\
Optimizer & AdamW (grad clip 1.0) & AdamW (grad clip 1.0) \\
EMA Decay & 0.999 & 0.999 \\
Warmup Steps & 10k & 10k \\
Training Steps & 30k & 30k \\
Action Horizon & 10 & 10 \\
\bottomrule
\end{tabular}
\caption{Training configurations for $\pi$\textsubscript{0.5} variants. The expert-only variant freezes the VLM and fine-tunes only the Action Expert.}
\label{tab:appen_pi05_expert_only_finetune}
\end{table*}
\paragraph{Evaluation Protocol.}
Each model is evaluated across 5 different random seeds (7, 8, 9, 10, 11) per task--paraphrase configuration.
All reported success rates represent the \textbf{mean over 5 seeds}; standard deviations are not reported, as our analysis focuses on aggregate robustness trends across paraphrase types rather than per-configuration variance.
We use the LIBERO simulation environment with its default evaluation settings (i.e., maximum episode length and success criteria) as defined in the original LIBERO benchmark.

\subsection{Result}
\label{appen:Experiment_Result}

\paragraph{Reporting Protocol.}
All success rate values reported in this paper are the mean of 5 independent evaluation runs with different random seeds.
We do not perform hyperparameter search for evaluation; all models are evaluated using their officially released or documented inference configurations.

\begin{figure*}[t]
    \centering
    \includegraphics[width=1\linewidth]{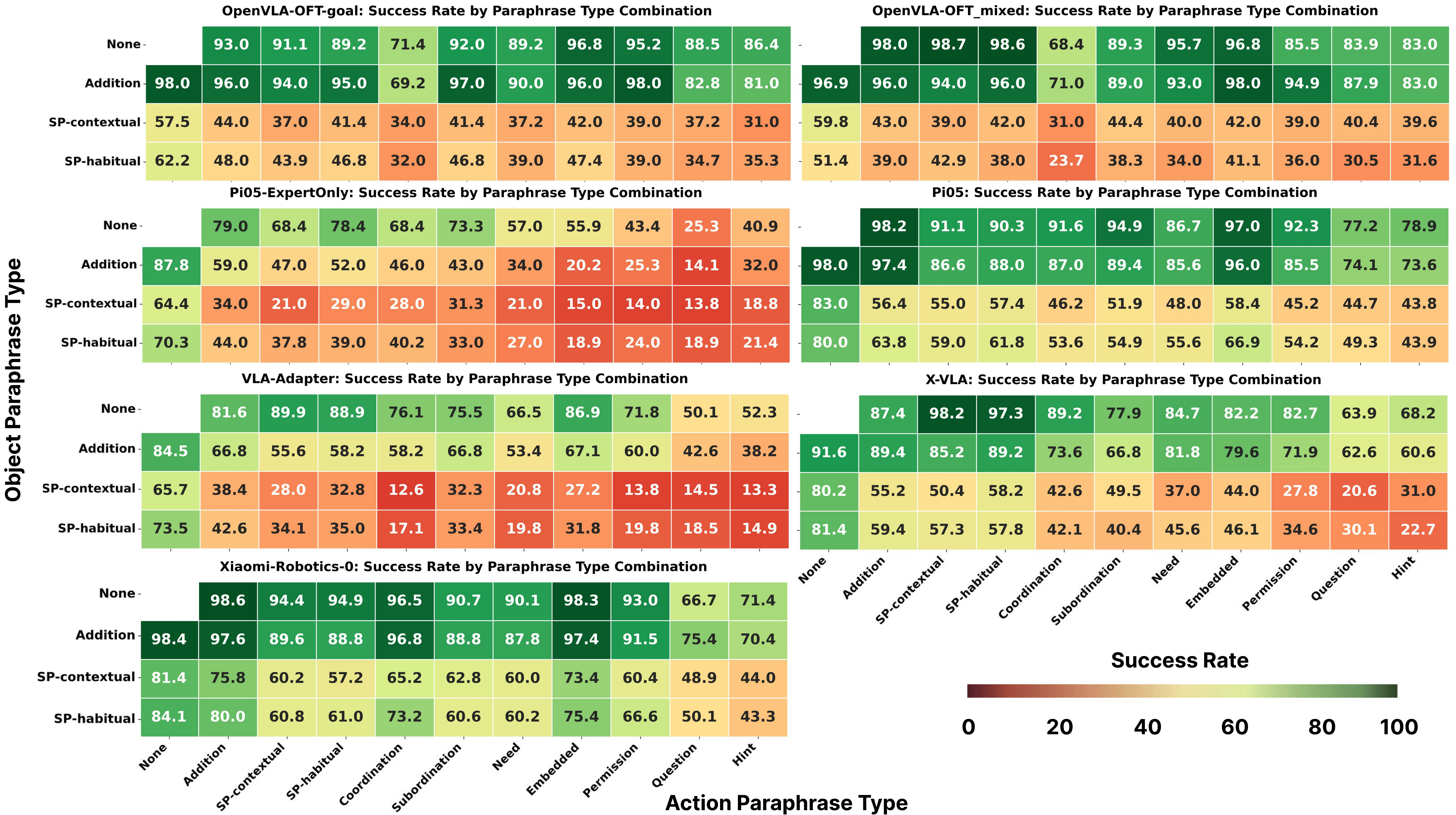}
    \caption{Per-model success rate heatmaps across all Object $\times$ Action 
paraphrase type combinations on LIBERO-Para. Rows represent object 
paraphrase types and columns represent action paraphrase types. 
Each cell reports the mean success rate over 5 seeds. 
The \texttt{None} row/column indicates the original (unparaphrased) 
instruction. All models show consistent degradation as paraphrase 
distance increases from the top-left (original) to the bottom-right 
(most distant) cells.}
    \label{fig:model_all_result}
\end{figure*}

\section{Analysis}
\label{appen:Analysis}

\subsection{Finding 1: Paraphrase Fragility Persists Across Architectures, Data Scales, and Fine-tuning Strategies}
\label{appen:Analysis_finding1}

Fig.~\ref{fig:model_all_result} presents per-model 
success rate heatmaps across all Object $\times$ Action 
paraphrase type combinations.
While all seven models degrade under object paraphrasing, 
the degradation manifests in two distinct patterns.
OpenVLA-OFT variants and VLA-Adapter exhibit a sharp 
cliff between object-preserved rows (None, Addition) 
and object-paraphrased rows (SP-contextual, SP-habitual)
---the top two rows remain nearly uniformly green 
while the bottom two rows shift abruptly to red.
This two-band pattern directly reflects the large 
preserved-vs-paraphrased gaps reported in 
Fig.~\ref{fig:object_preserved_vs_paraphrased}: 
OpenVLA-OFT\textsubscript{goal} (48.3\,pp), 
OpenVLA-OFT\textsubscript{mixed} (51.0\,pp), 
and VLA-Adapter (37.1\,pp) all show a clear visual 
boundary at the object paraphrasing threshold.
In contrast, $\pi_{0.5}$, X-VLA, and 
Xiaomi-Robotics-0 display a more gradual 
degradation across both axes without a single 
sharp boundary, consistent with their comparatively 
smaller preserved-vs-paraphrased gaps 
(19.8--35.7\,pp).

Despite these differences in degradation profile, 
the conclusion is shared: every model falls below 
50\% in the most challenging compound cells, 
confirming that paraphrase fragility is universal 
regardless of architecture.

\subsection{Finding 2: Object Grounding Is the Primary Bottleneck}
\label{appen:Analysis_finding2}
\paragraph{Alpha Sensitivity Analysis.}
Fig.~\ref{fig:alpha_sweep} examines how the balance between the two PRIDE 
components---keyword similarity ($S_K$) and structural similarity ($S_T$)---affects 
the overall robustness score.
As $\alpha$ shifts toward 1.0 (keyword-centric), PRIDE scores decrease 
across all models, revealing that models generally succeed on samples 
where keywords are preserved and fail when keywords are paraphrased.
Conversely, as $\alpha$ approaches 0.0 (structure-centric), scores rise 
uniformly, suggesting that structural variation alone is less disruptive 
than keyword replacement.
This confirms that object-level keyword changes, rather than 
syntactic reformulations, are the dominant factor driving 
success rate degradation across current VLA architectures.

The per-model slopes in the right panel of Fig.~\ref{fig:alpha_sweep} 
further reveal architecture-specific sensitivities.
OpenVLA-OFT\textsubscript{goal} and OpenVLA-OFT\textsubscript{mixed} 
exhibit the steepest slopes ($-17.3$ and $-18.7$, respectively), 
consistent with the large success rate gaps between object-preserved 
and object-paraphrased conditions reported in 
Fig.~\ref{fig:object_preserved_vs_paraphrased} 
(48.3pp and 51.0pp, respectively).
Their high keyword dependence indicates that these models rely 
heavily on exact object noun matching for task execution.

Interestingly, at $\alpha \geq 0.9$, X-VLA overtakes 
OpenVLA-OFT\textsubscript{mixed} in PRIDE score, indicating that 
despite lower structural robustness overall, X-VLA is more resilient 
to keyword-level variation.
Similarly, $\pi_{0.5}$ (expert-only) closes the gap with VLA-Adapter 
at $\alpha = 1.0$, suggesting relatively stronger keyword robustness 
despite its lower absolute performance.
These crossover patterns demonstrate the diagnostic utility of 
$\alpha$-\rev{sweep}: by adjusting the weighting, practitioners can 
identify which robustness dimension---keyword preservation or 
structural flexibility---a given model excels at, informing 
model selection for deployment environments where one linguistic 
dimension may be more prevalent than the other.

\paragraph{Action Indirectness.}
Fig.~\ref{fig:action_axis} breaks down success rate by 
action paraphrase type across all models.
Lexical-level changes (Addition, SP-contextual, SP-habitual) cause 
moderate degradation (66--70\%), while structural reorganizations 
(Coordination, Subordination) reduce success further to around 57--63\%.
The sharpest drops occur in the pragmatic category, where Question 
and Hint---forms that require pragmatic inference to recover the 
underlying imperative---bring success down to $\sim$48\%.

Notably, the overall action-axis degradation is milder than the 
object-axis degradation reported in Sec.~\ref{finding:finding2}.
We attribute this to the constrained nature of tabletop manipulation: 
the action space is limited to a small set of motor primitives 
(pick, place, push, open, etc.), and each object typically affords 
only a narrow range of feasible actions 
(e.g., \textit{stove} $\to$ \textit{turn on}).
This low action ambiguity allows models to converge on the correct 
primitive even under moderate linguistic variation.
However, when the directive intent itself becomes opaque---as in 
questions or hints---models can no longer reliably extract the 
intended action, leading to the steep drop in the pragmatic category.

\paragraph{LIBERO-Goal Instructions.} LIBERO-Goal instructions refer to each object by a single fixed reference throughout all tasks. As shown in Fig.~\ref{fig:canonical_instructions}, objects such as \textit{stove}, \textit{bowl}, and \textit{rack} appear consistently under the same surface form, with no synonym or alternative reference used in any instruction. Because models are fine-tuned exclusively on these fixed references, they are never exposed to lexical variation in object references during training. This single-reference convention likely reinforces surface-level keyword matching and contributes to the sharp performance drops observed when object nouns are replaced with semantically equivalent alternatives in LIBERO-Para.

\subsection{Finding 3: Failures Are Predominantly Planning-Level, Not Execution-Level}
\label{appen:Analysis_finding3}
\begin{figure*}[t]
    \centering
    \includegraphics[width=1\linewidth]{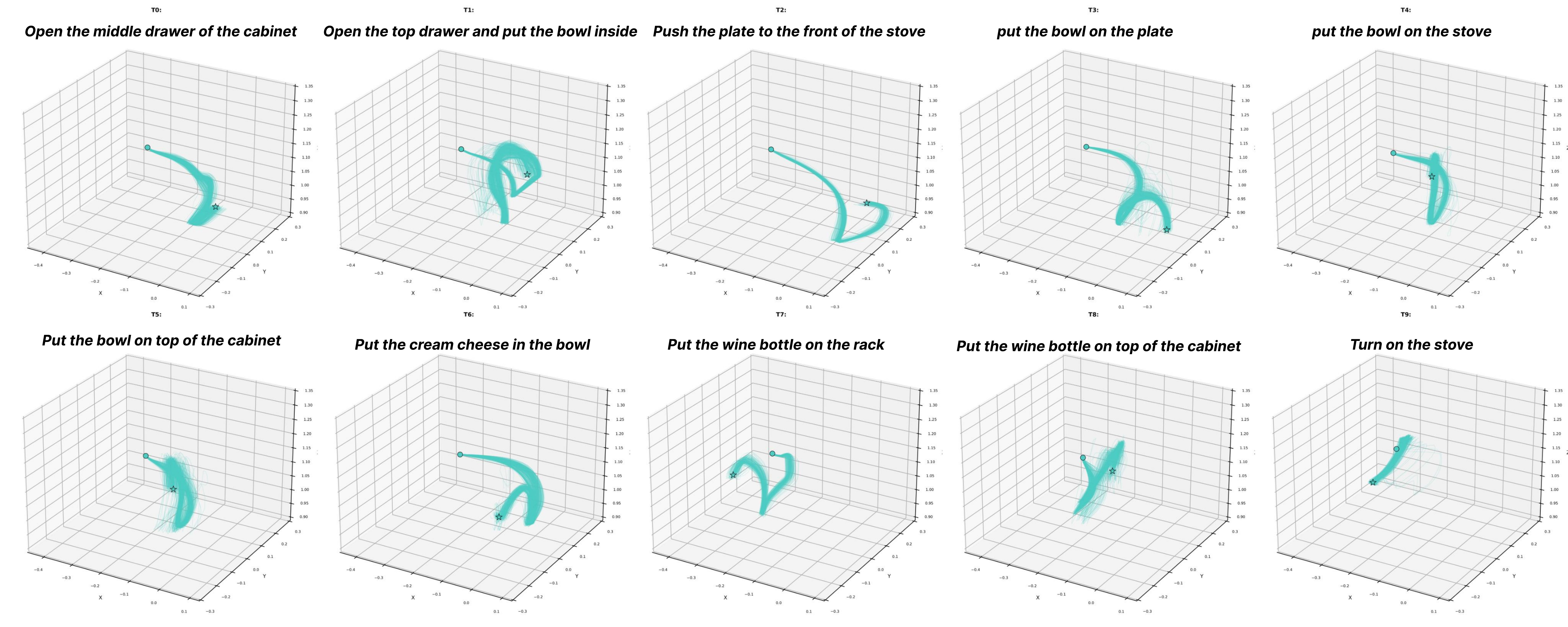}
    \caption{Successful EEF trajectories of Xiaomi-Robotics-0 on LIBERO-Para, grouped by LIBERO-Goal task index (T0--T9). Within each task, successful trajectories converge to a narrow corridor with low spatial variance, indicating that manipulation strategies are largely invariant to paraphrase variation. We observe consistent patterns across all evaluated models; a single model is shown for visual clarity. This consistency motivates the use of the mean successful trajectory as a pseudo ground-truth (GT) in Algorithm ~\ref{alg:near-far-gt}.}
    \label{fig:success_trajectory_pretty}
\end{figure*}

\begin{table*}[t]
\centering
\small
\begin{tabular}{lcccc c lcccc}
\toprule
\multicolumn{5}{c}{Near-GT \% of Total} & & \multicolumn{5}{c}{Far-GT \% of Total} \\
\cmidrule{1-5} \cmidrule{7-11}
Model & max & p99 & p95 & p90 & & Model & max & p99 & p95 & p90 \\
\midrule
OpenVLA-OFT\textsubscript{goal}  & 1.6  & 1.4 & 0.4 & 0.3 & & OpenVLA-OFT\textsubscript{goal}  & 33.7 & 33.9 & 34.9 & 35.0 \\
OpenVLA-OFT\textsubscript{mixed} & 3.3  & 1.0 & 0.1 & 0.0 & & OpenVLA-OFT\textsubscript{mixed} & 33.0 & 35.3 & 36.2 & 36.3 \\
$\pi$\textsubscript{0.5}         & 2.4  & 1.0 & 0.6 & 0.3 & & $\pi$\textsubscript{0.5}         & 26.2 & 27.6 & 28.0 & 28.3 \\
$\pi$\textsubscript{0.5} (expert-only) & 12.5 & 6.2 & 2.1 & 0.9 & & $\pi$\textsubscript{0.5} (expert-only) & 48.4 & 54.7 & 58.8 & 60.0 \\
VLA-Adapter                       & 4.2  & 2.2 & 0.5 & 0.0 & & VLA-Adapter                       & 49.5 & 51.5 & 53.2 & 53.7 \\
X-VLA                             & 5.2  & 2.9 & 0.8 & 0.4 & & X-VLA                             & 32.7 & 35.0 & 37.1 & 37.5 \\
Xiaomi-Robotics-0                 & 1.8  & 0.3 & 0.1 & 0.0 & & Xiaomi-Robotics-0                 & 22.2 & 23.7 & 23.9 & 24.0 \\
\bottomrule
\end{tabular}
\caption{$\tau$ threshold ablation for trajectory-based failure classification. ``max'' denotes the most lenient threshold (widest Near-GT boundary); p99, p95, and p90 progressively tighten the criterion. Across all thresholds, Far-GT (planning-level) failures consistently dominate, confirming that the finding is robust to threshold selection.}
\label{tab:tau_ablation}
\end{table*}

\paragraph{DTW-Based Trajectory Classification.}
We classify each failed episode as Near-GT (execution-level) 
or Far-GT (planning-level) based on its Dynamic Time Warping 
(DTW) distance to a pseudo ground-truth (GT) trajectory, 
as formalized in Algorithm~\ref{alg:near-far-gt}.

\paragraph{Why DTW.}
Trajectory lengths vary across episodes---successful episodes 
may terminate early while failed episodes often run to the 
maximum step limit. Euclidean distance requires fixed-length 
inputs and cannot account for temporal misalignment between 
trajectories that follow similar spatial paths at different 
speeds. DTW handles both variable-length sequences and 
temporal warping, making it suitable for comparing 
manipulation trajectories. We use \texttt{fastdtw}~\citep{fastdtw} with 
Euclidean distance as the local cost function, and normalize 
the resulting distance by sequence length to ensure 
comparability across episodes.

\paragraph{Resampling.}
To standardize input length for DTW, all trajectories are 
resampled to $K{=}50$ points via linear interpolation. 
This value was chosen as a practical trade-off between 
spatial resolution and computational cost across 
${\sim}$143K total episodes (4,092 paraphrases $\times$ 
5 seeds $\times$ 7 models).

\paragraph{EEF Position Only.}
From the 7-dimensional proprioceptive state 
$(x, y, z, r_x, r_y, r_z, g)$, we use only 
the first three dimensions corresponding to the 
end-effector (EEF) absolute position $(x, y, z)$. 
The remaining dimensions (orientation, gripper state) are 
excluded because spatial trajectory divergence is the most 
direct indicator of whether the model planned toward the 
correct target object---the core diagnostic question of 
this analysis.

\paragraph{Threshold Robustness.}
The threshold $\tau_t$ is set per-task as the maximum DTW 
distance among successful episodes 
(Algorithm~\ref{alg:near-far-gt}, line 12), representing 
the most lenient Near-GT boundary. To verify that our 
findings are not sensitive to this choice, we repeat the 
classification with progressively stricter thresholds 
(p99, p95, p90 of successful DTW distances). As shown in 
Tab.~\ref{tab:tau_ablation}, Far-GT failures remain 
dominant across all thresholds---tightening $\tau$ shifts 
some Near-GT episodes to Far-GT but does not alter the 
overall conclusion. For example, even under the strictest 
criterion (p90), $\pi_{0.5}$ (expert-only) retains 
the highest Near-GT ratio among all models, consistent 
with the frozen-VLM interpretation discussed in 
Sec.~\ref{finding:finding3}.

\paragraph{GT Trajectory Consistency.}
Fig.~\ref{fig:success_trajectory_pretty} visualizes 
successful EEF trajectories for each LIBERO-Goal task. 
Within each task, successful trajectories converge to a 
narrow spatial corridor with low variance, validating the 
use of their mean as a pseudo GT. This consistency arises 
from the LIBERO-Goal training data, which contains a single 
fixed demonstration path per task with no route diversity.

\begin{figure*}[t]
    \centering
    \includegraphics[width=0.95\linewidth]{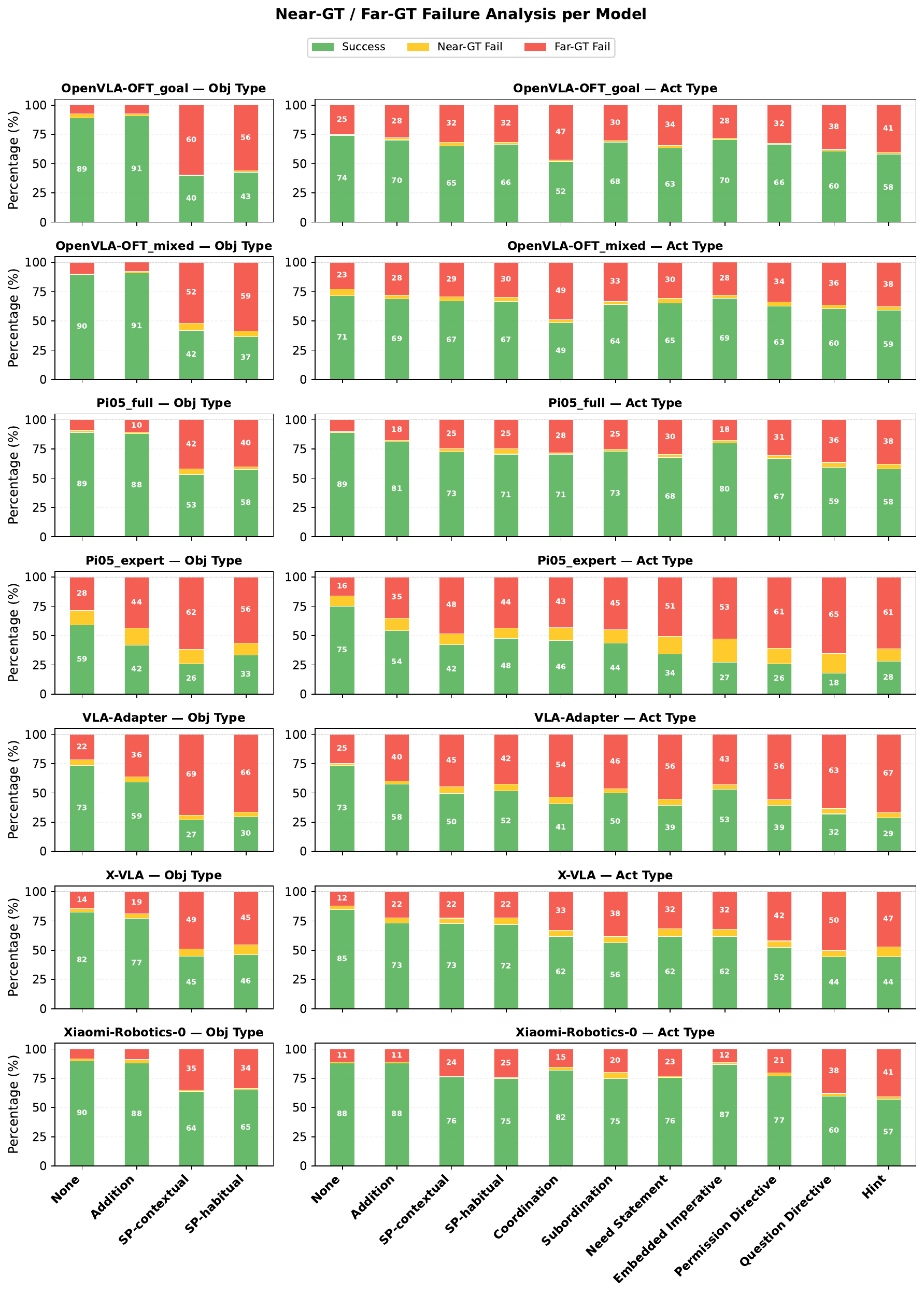}
    \caption{Near-GT / Far-GT failure breakdown per model, 
decomposed by Object axis (left) and Action axis (right). 
Each bar shows the proportion of Success (green), 
Near-GT failure (yellow, execution-level), and Far-GT 
failure (red, planning-level) episodes. The threshold 
$\tau$ is set to the maximum DTW distance among 
successful episodes per task. Across all models and 
paraphrase types, Far-GT failures consistently dominate, 
with no concentration of Near-GT failures along any 
specific axis. The exception is $\pi_{0.5}$ (expert-only), 
which exhibits a higher Near-GT ratio due to its frozen 
VLM preserving partial task identification while the 
unadapted action expert fails at execution.}
    \label{fig:nearGT_farGT_deepview_verybig}
\end{figure*}

\paragraph{Per-Model Failure Decomposition.}
Fig.~\ref{fig:nearGT_farGT_deepview_verybig} provides a 
fine-grained view of the failure classification 
from Tab.~\ref{tab:failure_classification}, 
decomposed along the Object and Action axes for 
each model individually.
Two observations are consistent across all models.
First, Near-GT (execution-level) failures account 
for a small fraction in every category, confirming 
that the dominance of Far-GT failures reported in 
Sec.~\ref{finding:finding3} is not an artifact 
of aggregation but holds at the per-type level.
Second, Near-GT failures do not concentrate along 
any particular paraphrase axis or type---they are 
distributed roughly uniformly, suggesting that 
execution-level errors are not systematically 
triggered by specific linguistic properties.

The sole exception is $\pi_{0.5}$ (expert-only), which 
shows elevated Near-GT ratios across most categories. 
As discussed in Sec.~\ref{finding:finding1}, this 
model freezes the VLM during fine-tuning, preserving 
pretrained language understanding that enables partial 
task identification. However, the unadapted action 
expert lacks the precision to convert correct plans 
into successful executions, resulting in trajectories 
that track the GT path but ultimately fail.

These patterns reinforce the conclusion that 
paraphrase robustness improvements should target 
the instruction-to-task identification stage---where 
Far-GT failures originate---rather than low-level 
motor control refinement.

\section{\rev{CALVIN-Para: Cross-Benchmark Validation}}
\label{sec:calvinpara}
\begin{table*}[t]
\centering
\rev{
\small
\begin{tabular}{lll}
\toprule
 & RoboFlamingo & FLOWER \\
\midrule
Variant & MPT-IFT-3B & -- \\
Split & ABCD$\to$D & ABCD$\to$D \\
Code & \href{https://github.com/RoboFlamingo/RoboFlamingo}{github.com/RoboFlamingo/RoboFlamingo} & \href{https://github.com/intuitive-robots/flower_vla_calvin}{github.com/intuitive-robots/flower\_vla\_calvin} \\
Weights & \href{https://huggingface.co/robovlms/RoboFlamingo}{hf.co/robovlms/RoboFlamingo} & \href{https://huggingface.co/mbreuss/flower_calvin_abcd}{hf.co/mbreuss/flower\_calvin\_abcd} \\
\bottomrule
\end{tabular}
}
\caption{\rev{Official pretrained checkpoints used for CALVIN-Para evaluation.}}
\label{tab:calvin_models}
\end{table*}
\subsection{\rev{Setup}}
\label{sec:calvinpara-setup}
\paragraph{\rev{CALVIN Benchmark.}}
\rev{CALVIN~\citep{calvin} is an open-source simulated benchmark for
long-horizon, language-conditioned manipulation, in which a 7-DoF Franka Panda
arm operates on a tabletop containing a sliding door, a drawer, colored blocks,
an LED, and a light bulb. It comprises four environments (A--D) that differ in
desk texture and object layout, and defines 34 manipulation tasks with
natural-language instructions. In the standard protocol, policies are evaluated
on 1,000 chains of five consecutive language-conditioned sub-tasks, where the
initial state is instantiated for the first sub-task of each chain and success
on later sub-tasks depends on earlier ones.}

\begin{figure}[t]
    \centering
    \includegraphics[width=1\linewidth]{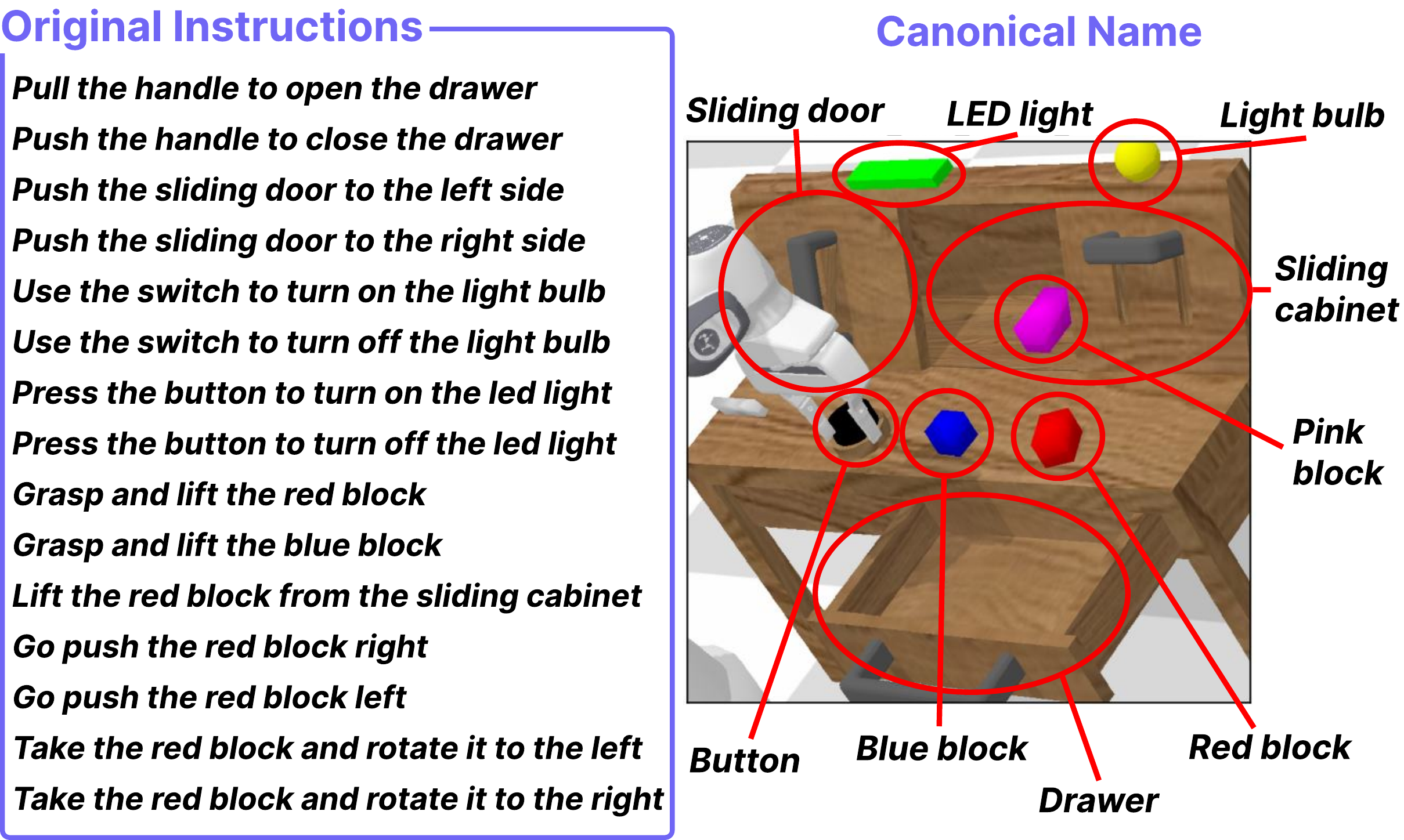}
    \caption{\rev{Base instructions used in CALVIN-Para (left) and corresponding scene with canonical object names (right). As in LIBERO-Goal (Fig.~\ref{fig:canonical_instructions}), each object is referred to by a single unique keyword throughout all instructions (e.g., \textit{drawer}, \textit{sliding door}, \textit{red block}), with no lexical variation across tasks.}}
    \label{fig:calvinpara_visual}
\end{figure}
\paragraph{\rev{Benchmark Construction.}}
\rev{We construct CALVIN-Para by applying the two-axis paraphrase design of
LIBERO-Para (Sec.~\ref{sec:liberopara}) to CALVIN~\citep{calvin} under the standard ABCD$\to$D split.
From the 34 original instructions, we select 15 base tasks in two stages.
First, we exclude 14 tasks distinguished solely by object color (e.g., lifting
the red, blue, or pink block), since color terms admit few natural synonyms and
thus cannot support meaningful object-axis paraphrasing; tasks in which a
colored object is the sole referent are retained. Second, we exclude 5 tasks
(e.g., \textit{place\_in\_drawer}, \textit{stack\_block}) that never appear in
the first slot of CALVIN's 5-task evaluation chains: their initial states
require preceding manipulation and cannot be instantiated without altering
CALVIN's original initial-state distribution, which would break the canonical
success-rate comparison. Each of the 43 variation cells contains 3 paraphrases
per base task (45 instructions per cell), yielding 1,935 paraphrased
instructions in total. Fig.~\ref{fig:calvinpara_visual} shows the 15 base
instructions and the corresponding scene.}

\paragraph{\rev{Evaluation Protocol.}}
\label{sec:calvinpara-evaluation-protocol}
\rev{To isolate paraphrase effects, we convert CALVIN's 5-step task chains into
a single-task setting. Unlike LIBERO-Goal, initial scenes vary across episodes;
the language instruction nevertheless serves as the task cue, so the effect of
paraphrasing is measured cleanly. We adopt the ABCD$\to$D split so that the
evaluation environment lies within the training distribution: tasks are seen
under canonical instructions while only the paraphrases are unseen, matching
the seen-task, unseen-instruction setting of LIBERO-Para.}
\begin{table}[t]
\centering
\small
\begin{tabular*}{\columnwidth}{@{\extracolsep{\fill}}lcc@{}}
\toprule
\multirow{2}{*}{Method} & CALVIN & CALVIN-Para \\
 & SR & SR (Drop) \\
\midrule
RoboFlamingo & 92.3 & 53.1 (\textcolor{red}{-39.2}) \\
FLOWER       & 99.7 & 58.3 (\textcolor{red}{-41.4}) \\
\bottomrule
\end{tabular*}
\caption{\rev{Success rate (SR) comparison between CALVIN and CALVIN-Para. Drop denotes the absolute decrease in success rate.}}
\label{tab:calvinpara_sr_drop}
\vspace{-0.10in}
\end{table}
\begin{table}[t]
\centering
\resizebox{\columnwidth}{!}{
\begin{tabular}{lccc}
\toprule
Method & SR & PRIDE & \textbf{Overestimation (\%)} \\
\midrule
RoboFlamingo & 53.1 & 43.5 & \textbf{18.1} \\
FLOWER       & 58.3 & 48.3 & \textbf{17.2} \\
\bottomrule
\end{tabular}
}
\caption{\rev{SR and PRIDE scores on CALVIN-Para, sorted by overestimation. Overestimation is computed as (SR\,--\,PRIDE)\,/\,SR, indicating how much uniform success rate overstates a model's paraphrase robustness. PRIDE is computed under the same formulation as Tab.~\ref{tab:pride_overestimation} (Eq.~\ref{eq:4}).}}
\label{tab:calvin_pride_overestimation}
\vspace{-0.10in}
\end{table}
\begin{figure*}[t]
    \centering
    \includegraphics[width=1\linewidth]{figures/calvinpara_heatmap_img.pdf}
    \caption{\rev{Per-model success rate heatmaps across all Object $\times$ Action
paraphrase type combinations on CALVIN-Para (left: RoboFlamingo, right:
FLOWER). Rows represent object paraphrase types and columns represent action
paraphrase types. Each cell reports the mean success rate over 5 seeds. The
\texttt{None} row/column indicates the original (unparaphrased) instruction.
As on LIBERO-Para (Fig.~\ref{fig:model_all_result}), both models show
consistent degradation as paraphrase distance increases from the top-left
(original) to the bottom-right (most distant) cells.}}
    \label{fig:calvinpara_heatmap}
\end{figure*}
\paragraph{\rev{Models.}}
\label{sec:calvinpara-models}
\rev{We evaluate two CALVIN-native VLAs not used in our LIBERO experiments,
RoboFlamingo~\citep{roboflamingo} and FLOWER~\citep{flower}, using official
ABCD$\to$D pretrained weights without modification
(Tab.~\ref{tab:calvin_models}). All inference is run on NVIDIA L40S GPUs, and
all results are averaged over 5 seeds.}

\begin{table}[t]
\centering
\small
\begin{tabular}{lcc}
\toprule
Method & Obj. preserved & Obj. paraphrased ($\Delta$) \\
\midrule
RoboFlamingo & 69.8 & 37.2 (\textcolor{red}{-32.6}) \\
FLOWER       & 77.6 & 39.9 (\textcolor{red}{-37.7}) \\
\bottomrule
\end{tabular}
\caption{\rev{Mean success rate for object-preserved (None, Addition) versus object-paraphrased (SP-contextual, SP-habitual) instructions on CALVIN-Para, following Fig.~\ref{fig:object_preserved_vs_paraphrased}. $\Delta$ denotes the absolute decrease in success rate.}}
\label{tab:calvinpara_obj_preserve_paraphrase}
\vspace{-0.10in}
\end{table}
\begin{table}[t]
\centering
\resizebox{\columnwidth}{!}{
\begin{tabular}{lcccc}
\toprule
\multirow{2}{*}{Model} & Success & \multicolumn{3}{c}{Failure} \\
\cmidrule(l){3-5}
 & Rate & Near-GT & Far-GT & \textbf{Far-GT(\%)} \\
\midrule
FLOWER       & 58.3 & 1.2 & 40.5 & \textbf{97.1} \\
RoboFlamingo & 53.1 & 3.1 & 43.8 & \textbf{93.4} \\
\bottomrule
\end{tabular}
}
\caption{\rev{Failure classification on CALVIN-Para (by Far-GT (\%)). Near-GT: execution-level failure near GT trajectory. Far-GT: planning-level failure from GT. As in LIBERO-Para (Tab.~\ref{tab:failure_classification}), the vast majority of failures are planning-level.}}
\label{tab:calvinpara_fargt}
\end{table}
\subsection{\rev{Results}}
\label{sec:calvinpara-results}
\paragraph{\rev{Success Rate Comparison.}}
\rev{As shown in Tab.~\ref{tab:calvinpara_sr_drop}, despite near-saturated
performance of 92.3\% and 99.7\% on canonical CALVIN instructions,
RoboFlamingo and FLOWER drop by 39.2 and 41.4\,pp under paraphrasing, within
the 22.8--51.9\,pp range observed on LIBERO-Para in Tab.~\ref{tab:sr_drop}.
At cell level, Fig.~\ref{fig:calvinpara_heatmap} shows the same overall
tendency as on LIBERO-Para: success rates degrade monotonically along both
axes, from the top-left (original) toward the bottom-right (most distant)
cells.}

\paragraph{\rev{PRIDE Reveals Hidden Severity.}}
\rev{Tab.~\ref{tab:calvin_pride_overestimation} shows that uniform SR
overstates robustness by 18.1\% and 17.2\% relative to PRIDE, comparable to
the most-overestimated models on LIBERO-Para in
Tab.~\ref{tab:pride_overestimation}. PRIDE thus captures the hidden severity
of paraphrase failures across benchmarks.}

\paragraph{\rev{Finding Reproduction.}}
\rev{All three findings from Sec.~\ref{sec:analysis} reproduce on CALVIN-Para.
First, the two models instantiate architecture designs absent from our LIBERO
experiments: RoboFlamingo couples a Flamingo-style VLM~\cite{flamingo}, in which language
tokens attend to visual features through gated cross-attention, with a
separate LSTM~\cite{lstm} policy head for temporal modeling, while FLOWER fuses a compact
Florence-2 encoder~\cite{xiao2024florence2} with a rectified-flow~\cite{rectifiedflow} DiT~\cite{diffusiontransformer} action expert at an intermediate
VLM layer. Both nonetheless exhibit the same fragility, extending Finding 1
in Sec.~\ref{finding:finding1}: paraphrase fragility persists regardless of
how vision, language, and action generation are coupled. Second, as shown in
Tab.~\ref{tab:calvinpara_obj_preserve_paraphrase}, replacing canonical object
names degrades SR by 32.6 and 37.7\,pp, mirroring the 19.8--51.0\,pp drops in
Fig.~\ref{fig:object_preserved_vs_paraphrased} and reproducing Finding 2 in
Sec.~\ref{finding:finding2}: models remain anchored to the single canonical
object word seen during training. Third, Tab.~\ref{tab:calvinpara_fargt}
shows that 93.4\% and 97.1\% of failures are Far-GT, consistent with the
79.5--95.5\% range in Tab.~\ref{tab:failure_classification} and reproducing
Finding 3 in Sec.~\ref{finding:finding3}: failures are predominantly
planning-level rather than execution-level.}

\begin{figure*}[t]
\begin{minipage}{\textwidth}
\begin{algorithm}[H]
\begin{algorithmic}[1]
\Require Set of episodes $\mathcal{E} = \{e_1, \ldots, e_N\}$ for original LIBERO-Goal task index $t \in \{0, 1, ..., 9\}$, each with trajectory $\boldsymbol{\tau}_i \in \mathbb{R}^{T_i \times 3}$ and outcome $s_i \in \{0,1\}$; resampling size $K{=}50$
\Ensure Classification of each failed episode as \textsc{Near-GT} or \textsc{Far-GT}
\Statex
\Statex \textbf{// Step 1: Partition episodes}
\State $\mathcal{S}_t \leftarrow \{e_i \in \mathcal{E} \mid s_i = 1\}$ \Comment{successes}
\State $\mathcal{F}_t \leftarrow \{e_i \in \mathcal{E} \mid s_i = 0\}$ \Comment{failures}
\Statex
\Statex \textbf{// Step 2: Construct pseudo-GT trajectory}
\For{each $e_i \in \mathcal{S}_t$}
    \State $\hat{\boldsymbol{\tau}}_i \leftarrow \textsc{Resample}(\boldsymbol{\tau}_i[\,:\,,\,:3],\; K)$
    \Comment{first 3 dims of proprio: EEF absolute position (x,y,z)}
\EndFor
\State $\boldsymbol{\tau}^{\text{GT}} \leftarrow \frac{1}{|\mathcal{S}_t|} \sum_{e_i \in \mathcal{S}_t} \hat{\boldsymbol{\tau}}_i$
\Statex
\Statex \textbf{// Step 3: Compute DTW distances}
\State $L_{\max} \leftarrow \max_{e_j \in \mathcal{S}_t} T_j$
\For{each $e_i \in \mathcal{E}$}
    \State $\boldsymbol{\tau}'_i \leftarrow \textsc{Resample}(\boldsymbol{\tau}_i[:L_{\max},\,:3],\; K)$
    \State $d_i \leftarrow \textsc{DTW}(\boldsymbol{\tau}'_i,\; \boldsymbol{\tau}^{\text{GT}})$
    \Comment{DTW: Dynamic Time Warping}
\EndFor
\Statex
\Statex \textbf{// Step 4: Threshold}
\State $\tau_t \leftarrow \max_{e_i \in \mathcal{S}_t} d_i$
\Statex
\Statex \textbf{// Step 5: Classify failures}
\For{each $e_i \in \mathcal{F}_t$}
    \If{$d_i \leq \tau_t$}
        \State $\textsc{Label}(e_i) \leftarrow \textsc{Near-GT}$ \Comment{execution-level}
    \Else
        \State $\textsc{Label}(e_i) \leftarrow \textsc{Far-GT}$ \Comment{planning-level}
    \EndIf
\EndFor
\State \Return $\{\textsc{Label}(e_i)\}_{e_i \in \mathcal{F}_t}$
\end{algorithmic}
\caption{Trajectory-based failure classification for Sec. ~\ref{finding:finding3} A pseudo ground-truth trajectory (GT) is constructed from successful episodes of each model on LIBERO-Para. Failed episodes are classified as Near-GT (execution-level) or Far-GT (planning-level) based on DTW distance, with the threshold $\tau_t$ set to the maximum distance among successes.}
\label{alg:near-far-gt}
\end{algorithm}
\end{minipage}
\end{figure*}

\section{AI Assistants}

During the course of this work, we used Google's Gemini 2.5 Pro (\url{https://gemini.google.com/})~\citep{gemini} for generating paraphrase candidates in the LIBERO-Para benchmark construction. All generated paraphrases were manually reviewed and filtered by the authors. Additionally, we used AI assistants including OpenAI's ChatGPT (\url{https://chatgpt.com/})~\citep{openai2023gpt4} and Anthropic's Claude (\url{https://claude.ai/})~\citep{anthropic2025claude4} to proofread and improve the clarity of our writing. We affirm that these tools served solely as assistive aids and did not contribute to core research ideas, experimental design, analysis, or interpretation of results. The final scientific content and all claims made in this paper are the sole responsibility of the authors.

\begin{figure*}[t]
\centering
\begin{tcolorbox}[
  colback=blue!5!white,
  colframe=blue!75!black,
  title=Generator Prompt,
  enhanced,
  boxsep=3pt,
  top=3pt,
  bottom=3pt,
  width=\textwidth
]
Paraphrase the given robot manipulation instruction while preserving semantic meaning.\par
\textbf{\\Rules:}
\begin{itemize}
  \setlength{\itemsep}{1pt}
  \setlength{\topsep}{1pt}
  \setlength{\parsep}{0pt}
  \item Preserve plurality (singular/plural must match original)
  \item Do not add visual attributes (color, size, shape, material)
  \item Do not add spatial attributes (position, location, direction)
  \item Modify only what is specified in the task guidelines below
\end{itemize}
\textbf{Scope:}
\begin{itemize}
  \setlength{\itemsep}{1pt}
  \setlength{\topsep}{1pt}
  \setlength{\parsep}{0pt}
  \item Object tasks: modify only object nouns, preserve verbs and structure
  \item Action tasks: modify only action elements, preserve object nouns
\end{itemize}
\textbf{Output:} One paraphrase per line. No explanations or alternatives.
\end{tcolorbox}
\caption{Common prompt template for the Paraphrase Generator (LLM), shared across all paraphrase types.}
\label{fig:appen_prompt1}
\end{figure*}
\begin{figure*}[t]
\centering
\begin{tcolorbox}[
  colback=blue!5!white,
  colframe=blue!75!black,
  title=Verifier Prompt,
  enhanced,
  boxsep=3pt,
  top=3pt,
  bottom=3pt,
  width=\textwidth
]
\textbf{\\Criteria:}
\begin{itemize}
  \setlength{\itemsep}{1pt}
  \setlength{\topsep}{1pt}
  \setlength{\parsep}{0pt}
  \item Task compliance: Required transformations applied, prohibited changes avoided.
  \begin{itemize}
    \setlength{\itemsep}{1pt}
    \setlength{\topsep}{1pt}
    \setlength{\parsep}{0pt}
    \item Object tasks (obj\_*): only object nouns changed, verbs/structure intact
    \item Action tasks (act\_*): only action elements changed, object nouns intact
  \end{itemize}
  \item Semantic preservation: Core action and target objects unchanged.
  \item Naturalness: Grammatically correct, natural phrasing.
  \item Format: Single instruction only, no meta-commentary or explanations.
  \item Lexical clarity: Avoid confusion with environment objects (stove, bowl, plate, rack, wine bottle, cream cheese, cabinet).
  \begin{itemize}
    \setlength{\itemsep}{1pt}
    \setlength{\topsep}{1pt}
    \setlength{\parsep}{0pt}
    \item Acceptable: ``cupboard'' (distinct from cabinet)
    \item Reject: ``dish'' (confusable with plate/bowl)
  \end{itemize}
\end{itemize}
\textbf{Output:} Accepted paraphrases only, one per line.
\end{tcolorbox}
\caption{Common prompt template for the Paraphrase Verifier (LLM), used to filter generated paraphrases.}
\label{fig:appen_prompt2}
\end{figure*}
\begin{figure*}[t]
\centering
\begin{tcolorbox}[
  colback=blue!5!white,
  colframe=blue!75!black,
  title=Merge Generator Prompt,
  enhanced,
  boxsep=3pt,
  top=3pt,
  bottom=3pt,
  width=\textwidth
]
Merge object paraphrases and action paraphrases into combined variants.

\textbf{Input:}
\begin{itemize}
  \setlength{\itemsep}{1pt}
  \setlength{\topsep}{1pt}
  \setlength{\parsep}{0pt}
  \item Original instruction
  \item Object paraphrase examples (nouns changed)
  \item Action paraphrase examples (verbs/structure changed)
\end{itemize}

\textbf{Task:} Create paraphrases with both object and action modifications applied.

\textbf{Process:}
\begin{enumerate}
  \setlength{\itemsep}{1pt}
  \setlength{\topsep}{1pt}
  \setlength{\parsep}{0pt}
  \item Identify object substitution patterns (e.g., ``drawer'' $\rightarrow$ ``compartment'')
  \item Identify action modification patterns (e.g., ``pick'' $\rightarrow$ ``grab'')
  \item Apply both transformations coherently
\end{enumerate}

\textbf{Example:}
\begin{itemize}
  \setlength{\itemsep}{1pt}
  \setlength{\topsep}{1pt}
  \setlength{\parsep}{0pt}
  \item Original: ``\textit{pick the bowl and place on the stove}''
  \item Object variant: ``\textit{pick the container and place on the cooktop}''
  \item Action variant: ``\textit{grab the bowl and put on the stove}''
  \item Merged: ``\textit{grab the container and put on the cooktop}''
\end{itemize}

\textbf{Output:} 5--10 merged paraphrases, one per line. No numbering or explanations.
\end{tcolorbox}
\caption{Prompt template for combining validated Object and Action paraphrases into merged variants.}
\label{fig:appen_prompt3}
\end{figure*}

\begin{figure*}[t]
\centering
\begin{tcolorbox}[
  colback=blue!5!white,
  colframe=blue!75!black,
  title=Merge Verifier Prompt,
  enhanced,
  boxsep=3pt,
  top=3pt,
  bottom=3pt,
  width=\textwidth
]
Evaluate merged paraphrases that combine object and action changes.

\textbf{Criteria:}
\begin{itemize}
  \setlength{\itemsep}{1pt}
  \setlength{\topsep}{1pt}
  \setlength{\parsep}{0pt}

  \item Completeness: Both object nouns AND action elements must differ from original.
  Reject if only one type changed.

  \item Pattern consistency: Changes follow the provided examples.

  \item Semantic preservation: Task intent and outcome unchanged.

  \item Naturalness: Grammatically correct, coherent combination.

  \item Format: Single instruction, no meta-commentary.

  \item Lexical clarity: No confusion with environment objects.
\end{itemize}

\textbf{Output:} Accepted paraphrases only, one per line.
\end{tcolorbox}
\caption{Prompt template for verifying merged paraphrases before final inclusion in the dataset.}
\label{fig:appen_prompt4}
\end{figure*}

\begin{figure*}[t]
\centering
\begin{tcolorbox}[
  width=\textwidth,
  colback=blue!5!white,
  colframe=blue!75!black,
  title=Object-Lexical Types Prompts,
  enhanced,
  boxsep=3pt,
  top=3pt,
  bottom=3pt
]

\textbf{obj\_lexical\_same\_polarity\_habitual} \\
Replace object nouns with general synonyms. \\
EPT: Same-polarity substitution (habitual)

\textbf{Guidelines:}
\begin{itemize}
  \setlength{\itemsep}{1pt}
  \setlength{\topsep}{1pt}
  \setlength{\parsep}{0pt}
  \item Use commonly accepted synonyms (bowl → cup, cabinet → cupboard)
  \item Preserve grammatical form
  \item Keep verbs and structure unchanged
\end{itemize}

\textbf{Examples:}
\begin{itemize}
  \setlength{\itemsep}{1pt}
  \setlength{\topsep}{1pt}
  \setlength{\parsep}{0pt}
  \item ``\textit{Pick the bowl and place on the stove}'' → ``\textit{Pick the cup and place on the stove.}''
  \item ``\textit{Open the middle layer of drawer}'' → ``\textit{Open the middle layer of compartment.}''
\end{itemize}

\vspace{6pt}
\hrule
\vspace{6pt}

\textbf{obj\_lexical\_same\_polarity\_contextual} \\
Replace object nouns with contextually appropriate alternatives. \\
EPT: Same-polarity substitution (contextual)

\textbf{Guidelines:}
\begin{itemize}
  \setlength{\itemsep}{1pt}
  \setlength{\topsep}{1pt}
  \setlength{\parsep}{0pt}
  \item Use contextually similar items (bowl → container, stove → cooking surface)
  \item Maintain semantic appropriateness for manipulation tasks
  \item Preserve structure
\end{itemize}

\textbf{Examples:}
\begin{itemize}
  \setlength{\itemsep}{1pt}
  \setlength{\topsep}{1pt}
  \setlength{\parsep}{0pt}
  \item ``\textit{Pick the bowl and place on the stove}'' → ``\textit{Pick the container and place on the cooking surface.}''
  \item ``\textit{Open the middle layer of drawer}'' → ``\textit{Open the middle tier of drawer.}''
\end{itemize}

\vspace{6pt}
\hrule
\vspace{6pt}

\textbf{obj\_lexical\_addition} \\
Add functional descriptors from object names. \\
EPT: Addition

\textbf{Guidelines:}
\begin{itemize}
  \setlength{\itemsep}{1pt}
  \setlength{\topsep}{1pt}
  \setlength{\parsep}{0pt}
  \item Add functional/categorical adjectives (soup bowl, kitchen cabinet)
  \item Exclude visual adjectives (color, size, material)
  \item Exclude spatial adjectives (top, left, big)
  \item Preserve plurality
\end{itemize}

\textbf{Examples:}
\begin{itemize}
  \setlength{\itemsep}{1pt}
  \setlength{\topsep}{1pt}
  \setlength{\parsep}{0pt}
  \item ``\textit{Pick the bowl and place on the stove}'' → ``\textit{Pick the mixing bowl and place on the kitchen stove.}''
  \item ``\textit{Open the middle layer of drawer}'' → ``\textit{Open the middle layer of storage drawer.}''
\end{itemize}
\end{tcolorbox}
\caption{Type-specific generation guidelines for Object-Lexical paraphrases. These guidelines are provided to the Generator for paraphrase generation and also appended to the Verifier to assess whether the generated output conforms to the intended variation type.}
\label{fig:appen_prompt5}
\end{figure*}

\begin{figure*}[t]
\centering
\begin{tcolorbox}[
  width=\textwidth,
  colback=blue!5!white,
  colframe=blue!75!black,
  title=Action-Lexical Types Prompts,
  enhanced,
  boxsep=3pt,
  top=3pt,
  bottom=3pt
]
\textbf{act\_lexical\_same\_polarity\_habitual} \\
Replace action verbs with general synonyms. \\
EPT: Same-polarity substitution (habitual)\par
\textbf{Guidelines:}
\begin{itemize}
  \setlength{\itemsep}{1pt}
  \setlength{\topsep}{1pt}
  \setlength{\parsep}{0pt}
  \item Use common verb synonyms (pick → grab, place → put)
  \item Preserve structure and arguments
  \item Keep object nouns unchanged
\end{itemize}
\textbf{Examples:}
\begin{itemize}
  \setlength{\itemsep}{1pt}
  \setlength{\topsep}{1pt}
  \setlength{\parsep}{0pt}
  \item ``\textit{Pick the bowl and place on the stove}'' → ``\textit{Grab the bowl and put on the stove.}''
  \item ``\textit{Open the middle layer of drawer}'' → ``\textit{Pull the middle layer of drawer.}''
\end{itemize}
\vspace{6pt}
\hrule
\vspace{6pt}
\textbf{act\_lexical\_same\_polarity\_contextual} \\
Replace action verbs with contextually appropriate alternatives. \\
EPT: Same-polarity substitution (contextual)\par
\textbf{Guidelines:}
\begin{itemize}
  \setlength{\itemsep}{1pt}
  \setlength{\topsep}{1pt}
  \setlength{\parsep}{0pt}
  \item Use context-appropriate alternatives (pick → grasp, place → position)
  \item Preserve core action meaning
  \item Keep object nouns unchanged
\end{itemize}
\textbf{Examples:}
\begin{itemize}
  \setlength{\itemsep}{1pt}
  \setlength{\topsep}{1pt}
  \setlength{\parsep}{0pt}
  \item ``\textit{Pick the bowl and place on the stove}'' → ``\textit{Grasp the bowl and position on the stove.}''
  \item ``\textit{Open the middle layer of drawer}'' → ``\textit{Access the middle layer of drawer.}''
\end{itemize}
\vspace{6pt}
\hrule
\vspace{6pt}
\textbf{act\_lexical\_addition\_deletion} \\
Add or remove manner adverbs. \\
EPT: Addition/Deletion\par
\textbf{Guidelines:}
\begin{itemize}
  \setlength{\itemsep}{1pt}
  \setlength{\topsep}{1pt}
  \setlength{\parsep}{0pt}
  \item Add single-word adverbs (carefully, gently, slowly)
  \item For phrasal additions, use act\_structural\_ellipsis instead
  \item Keep verb and structure unchanged
\end{itemize}
\textbf{Examples:}
\begin{itemize}
  \setlength{\itemsep}{1pt}
  \setlength{\topsep}{1pt}
  \setlength{\parsep}{0pt}
  \item ``\textit{Pick the bowl and place on the stove}'' → ``\textit{Carefully pick the bowl and place on the stove.}''
  \item ``\textit{Open the middle layer of drawer}'' → ``\textit{Gently open the middle layer of drawer.}''
\end{itemize}
\end{tcolorbox}
\caption{Type-specific generation guidelines for Action-Lexical paraphrases. These guidelines are provided to the Generator for paraphrase generation and also appended to the Verifier to assess whether the generated output conforms to the intended variation type.}
\label{fig:appen_prompt6}
\end{figure*}
\begin{figure*}[t]
\centering
\begin{tcolorbox}[
  width=\textwidth,
  colback=blue!5!white,
  colframe=blue!75!black,
  title=Action-Structural Types Prompts,
  enhanced,
  boxsep=3pt,
  top=3pt,
  bottom=3pt
]
\textbf{act\_structural\_coordination} \\
Modify coordination structure. \\
EPT: Coordination changes (syntax-based)\par
\textbf{Guidelines:}
\begin{itemize}
  \setlength{\itemsep}{1pt}
  \setlength{\topsep}{1pt}
  \setlength{\parsep}{0pt}
  \item Split into separate sentences
  \item Add explicit ordering (First... then...)
  \item Combine with coordination
  \item Preserve all information
\end{itemize}
\textbf{Examples:}
\begin{itemize}
  \setlength{\itemsep}{1pt}
  \setlength{\topsep}{1pt}
  \setlength{\parsep}{0pt}
  \item ``\textit{Pick the bowl and place on the stove}'' → ``\textit{Pick the bowl. Place it on the stove.}''
  \item ``\textit{Open the middle layer of drawer}'' → ``\textit{Locate the drawer, then open the middle layer.}''
\end{itemize}
\vspace{6pt}
\hrule
\vspace{6pt}
\textbf{act\_structural\_subordination} \\
Modify subordination structure. \\
EPT: Subordination changes (syntax-based)\par
\textbf{Guidelines:}
\begin{itemize}
  \setlength{\itemsep}{1pt}
  \setlength{\topsep}{1pt}
  \setlength{\parsep}{0pt}
  \item Convert coordination to subordination
  \item Use temporal subordinators (after, once, when)
  \item Use purpose subordinators (so that, in order to)
  \item Preserve semantic relations
\end{itemize}
\textbf{Examples:}
\begin{itemize}
  \setlength{\itemsep}{1pt}
  \setlength{\topsep}{1pt}
  \setlength{\parsep}{0pt}
  \item ``\textit{Pick the bowl and place on the stove}'' → ``\textit{After picking the bowl, place it on the stove.}''
  \item ``\textit{Open the middle layer of drawer}'' → ``\textit{Open the middle layer of drawer in order to access the contents.}''
\end{itemize}
\end{tcolorbox}
\caption{Type-specific generation guidelines for Action-Structural paraphrases. These guidelines are provided to the Generator for paraphrase generation and also appended to the Verifier to assess whether the generated output conforms to the intended variation type.}
\label{fig:appen_prompt7}
\end{figure*}
\begin{figure*}[t]
\centering
\begin{tcolorbox}[
  width=\textwidth,
  colback=blue!5!white,
  colframe=blue!75!black,
  title=Action-Pragmatic Types Prompts (1/2),
  enhanced,   
  boxsep=3pt,
  top=3pt,
  bottom=3pt
]

\textbf{act\_pragmatical\_need\_statement} \\
Express as speaker's need. \\
Ervin-Tripp Type 1: Need Statements\par
\textbf{Guidelines:}
\begin{itemize}
  \setlength{\itemsep}{1pt}
  \setlength{\topsep}{1pt}
  \setlength{\parsep}{0pt}
  \item Convert to first-person need (I need, I want, I require)
  \item Focus on desired outcome rather than process
  \item Preserve core action and objects
\end{itemize}
\textbf{Examples:}
\begin{itemize}
  \setlength{\itemsep}{1pt}
  \setlength{\topsep}{1pt}
  \setlength{\parsep}{0pt}
  \item ``\textit{Pick the bowl and place on the stove}'' → ``\textit{I need the bowl placed on the stove.}''
  \item ``\textit{Open the middle layer of drawer}'' → ``\textit{I want the middle layer of drawer open.}''
\end{itemize}
\vspace{6pt}
\hrule
\vspace{6pt}

\textbf{act\_pragmatical\_embedded\_imperative} \\
Embed in question frame with modals. \\
Ervin-Tripp Type 3: Embedded Imperatives\par
\textbf{Guidelines:}
\begin{itemize}
  \setlength{\itemsep}{1pt}
  \setlength{\topsep}{1pt}
  \setlength{\parsep}{0pt}
  \item Use modal questions (Could you, Would you, Can you)
  \item Keep agent and action explicit but softened
\end{itemize}
\textbf{Examples:}
\begin{itemize}
  \setlength{\itemsep}{1pt}
  \setlength{\topsep}{1pt}
  \setlength{\parsep}{0pt}
  \item ``\textit{Pick the bowl and place on the stove}'' → ``\textit{Could you pick the bowl and place it on the stove?}''
  \item ``\textit{Open the middle layer of drawer}'' → ``\textit{Can you open the middle layer of drawer?}''
\end{itemize}
\vspace{6pt}
\hrule
\vspace{6pt}

\textbf{act\_pragmatical\_permission\_directive} \\
Frame as permission request. \\
Ervin-Tripp Type 4: Permission Directives\par
\textbf{Guidelines:}
\begin{itemize}
  \setlength{\itemsep}{1pt}
  \setlength{\topsep}{1pt}
  \setlength{\parsep}{0pt}
  \item Request permission or access (May I have, Can I have access to)
  \item Focus on speaker's access requiring hearer's action
\end{itemize}
\textbf{Examples:}
\begin{itemize}
  \setlength{\itemsep}{1pt}
  \setlength{\topsep}{1pt}
  \setlength{\parsep}{0pt}
  \item ``\textit{Pick the bowl and place on the stove}'' → ``\textit{May I have the bowl placed on the stove?}''
  \item ``\textit{Open the middle layer of drawer}'' → ``\textit{Can I have access to the middle layer of drawer?}''
\end{itemize}
\end{tcolorbox}
\caption{Type-specific generation guidelines for Action-Pragmatic paraphrases (a). These guidelines are provided to the Generator for paraphrase generation and also appended to the Verifier to assess whether the generated output conforms to the intended variation type.}
\label{fig:appen_prompt8}
\end{figure*}
\begin{figure*}[t]
\centering
\begin{tcolorbox}[
  width=\textwidth,
  colback=blue!5!white,
  colframe=blue!75!black,
  title=Action-Pragmatic Types Prompts (2/2),
  enhanced,   
  boxsep=3pt,
  top=3pt,
  bottom=3pt
]

\textbf{act\_pragmatical\_question\_directive} \\
Pose information question implying action. \\
Ervin-Tripp Type 5: Question Directives\par
\textbf{Guidelines:}
\begin{itemize}
  \setlength{\itemsep}{1pt}
  \setlength{\topsep}{1pt}
  \setlength{\parsep}{0pt}
  \item Ask contextually relevant question
  \item Do not explicitly specify desired act
  \item Action inferred from context
\end{itemize}
\textbf{Examples:}
\begin{itemize}
  \setlength{\itemsep}{1pt}
  \setlength{\topsep}{1pt}
  \setlength{\parsep}{0pt}
  \item ``\textit{Pick the bowl and place on the stove}'' → ``\textit{Is the bowl still sitting on the counter?}''
  \item ``\textit{Open the middle layer of drawer}'' → ``\textit{Do we know the contents of the middle layer of drawer?}''
\end{itemize}
\vspace{6pt}
\hrule
\vspace{6pt}

\textbf{act\_pragmatical\_hint} \\
Make statement implying action through inference. \\
Ervin-Tripp Type 6: Hints\par
\textbf{Guidelines:}
\begin{itemize}
  \setlength{\itemsep}{1pt}
  \setlength{\topsep}{1pt}
  \setlength{\parsep}{0pt}
  \item Use general statement, not surface directive
  \item State condition or desired state
  \item Rely on situational inference
  \item Do not mention action directly
\end{itemize}
\textbf{Examples:}
\begin{itemize}
  \setlength{\itemsep}{1pt}
  \setlength{\topsep}{1pt}
  \setlength{\parsep}{0pt}
  \item ``\textit{Pick the bowl and place on the stove}'' → ``\textit{The stove surface is now clear for the bowl.}''
  \item ``\textit{Open the middle layer of drawer}'' → ``\textit{The middle layer of drawer is still closed.}''
\end{itemize}
\end{tcolorbox}
\caption{Type-specific generation guidelines for Action-Pragmatic paraphrases (b), continued. These guidelines are provided to the Generator for paraphrase generation and also appended to the Verifier to assess whether the generated output conforms to the intended variation type.}
\label{fig:appen_prompt9}
\end{figure*}


\end{document}